\documentclass{article} %
\usepackage{iclr2026_conference,times}

\usepackage{amsmath,amsfonts,bm}

\def\eqref#1{equation~\ref{#1}}

\def\1{\bm{1}}

\DeclareMathAlphabet{\mathsfit}{\encodingdefault}{\sfdefault}{m}{sl}
\SetMathAlphabet{\mathsfit}{bold}{\encodingdefault}{\sfdefault}{bx}{n}

\usepackage{hyperref}
\usepackage{url}
\usepackage{graphicx}
\usepackage[capitalise]{cleveref}
\usepackage{subcaption}
\usepackage{booktabs}
\usepackage{multirow}
\usepackage{placeins}

\title{Do Transformers Use their Depth Adaptively? Evidence from a Relational Reasoning Task}

\author{Alicia Curth, Rachel Lawrence, Sushrut Karmalkar \& Niranjani Prasad\\
Microsoft Research Cambridge\\
\texttt{\{aliciacurth,ralawrence,skarmalkar,niprasa\}@microsoft.com} }

\iclrfinalcopy %
\begin{document}

\maketitle

\begin{abstract}
We investigate whether transformers use their depth adaptively across tasks of increasing difficulty. Using a controlled multi-hop relational reasoning task based on family stories, where difficulty is determined by the number of relationship hops that must be composed, we monitor (i) how predictions evolve across layers via early readouts (the logit lens) and (ii) how task-relevant information is integrated across tokens via causal patching. For pretrained models, we find some limited evidence for adaptive depth use: some larger models need fewer layers to arrive at plausible answers for easier tasks, and models generally use more layers to integrate information across tokens as chain length increases. For models finetuned on the task, we find clearer and more consistent evidence of adaptive depth use, with the effect being stronger for less constrained finetuning regimes that do not preserve general language modeling abilities.

\end{abstract}

\section{Introduction}\label{sec:intro}
Large language models (LLMs) have made rapid advances over the last five years, with increasing model scale apparently driving the emergence of a priori unexpected capabilities \citep{brown2020language, wei2022emergent, ganguli2022predictability}. As models grow by stacking more layers, they gain additional computational \textit{depth}, yet it has recently been called into question whether this depth is actually used efficiently:  \cite{csordas2025language} find that the second half of a model's layers makes a much smaller contribution to the residual stream than earlier layers, and conclude through a series of empirical investigations that later layers are not performing fundamentally new computations.

These empirical findings stand in contrast with recent theoretical work emphasising depth as a critical bottleneck for transformers to solve algorithmic and reasoning tasks of increasing difficulty \citep{merrill2022saturated,sanford2024understanding, merrill2025little}. The theoretical importance of depth has further motivated the insight that chain-of-thought approaches work because they equip models with additional effective computational depth \citep{merrill2023expressive}, and has spurred investigations into recurrent transformer architectures that allow models to adaptively determine their own depth \citep{dehghani2018universal, saunshi2025reasoning, fan2024looped}.

The tension between these perspectives raises a natural question: even if depth use appears inefficient on average, might models have learned to implicitly ``reserve capacity'' for harder tasks? Here, we thus define adaptive depth use as the property that the effective depth a model relies on to produce an output scales with task difficulty.  While \cite{csordas2025language} and \cite{hu2025affects} present some empirical evidence that this is not the case, we revisit this question with an empirical setup that affords us more precise control over how task difficulty is defined and how depth use is measured. 

\textbf{Outlook.} We study a relational reasoning task, the family story relationship composition benchmark of \cite{sinha2019clutrr}, in which difficulty is controlled by varying the number of hops in the reasoning chain. We restrict answers to a single token, confining the computation of interest to a single forward pass. We apply logit lens \citep{nostalgebraist2020logit} and causal patching \citep{zhang2023towards} analyses to investigate how predictions and cross-token information integration, respectively, evolve differently across tasks of varying difficulty. Investigating the behavior of pretrained models, we consider five families of open-weights models ranging from 120M to 14B parameters (GPT-2, Pythia, Phi, Qwen2/2.5, and LLaMA), and find evidence that some larger models need fewer layers to arrive at plausible predictions for easier tasks, and that models rely on more layers for information integration across tokens as chain length increases. Investigating models finetuned on the task (GPT-2 and Pythia across a range of sizes), the picture depends on the training regime: models trained under a more constrained setup (preserving general language modeling ability) do not arrive at plausible predictions earlier for easier tasks, but do use more of their earlier layers for information integration -- effectively starting this process earlier in the network. Models trained under a less constrained regime show the strongest evidence of adaptive depth use: they use more layers overall for information integration, and arrive at plausible predictions earlier for easier tasks.

\section{Background and related work}\label{sec:background}
\textbf{Setup: Transformers and logit lens.} We consider only pre-Layer-Norm transformers with $L$ layers. Their hidden states $h_l \in \mathbb{R}^{d \times T}$, where $d$ is the hidden dimension and $T$ is the sequence length, evolve across depth via residual connections as
\begin{equation}\label{eq:residual}
  \textstyle  h_{l} = h_{l-1} + f_l(h_{l-1}) = h_k + \sum^{l-1}_{q=k}f_q(h_{q-1})
\end{equation}
where $f_l(h_{l-1})$ is a standard transformer block comprising attention, feed-forward, and layernorm operations. The predictive distribution over the next token at position $t$ is computed from the final hidden state $h_{L,t} \in \mathbb{R}^d$ via the language model head, parameterized by the unembedding matrix $W_U \in \mathbb{R}^{V \times d}$, a final layernorm $LN_f$, and a softmax:
\begin{equation}
  \textstyle   p_t = \text{softmax}(\text{logit}_t) \text{ with } \text{logit}_t = W_U LN_f(h_{L,t}) = W_U LN_f\!\left(h_{l, t} + \sum^L_{k=l} [f_k(h_{k-1})]_t\right)
\end{equation}

The logit lens \citep{nostalgebraist2020logit, geva2022transformer} is an interpretability technique that exploits the residual structure in \cref{eq:residual} to inspect intermediate hidden states by projecting them into vocabulary space using the language model head, effectively assuming that the remaining layer contributions $\sum^L_{k=l} [f_k(h_{k-1})]_t = 0$. The resulting early read-out distribution at layer $l$ is:
\begin{equation}\label{eq:logitlens}
    \tilde{p}_{l,T} = \text{softmax}(W_U LN_f(h_{l,T}))
\end{equation}

This method has been widely used to interpret language model hidden states in vocabulary space, e.g. \cite{halawi2023overthinking} use it to show that accuracy on correct answers peaks at intermediate layers before declining when models are provided with misleading in-context examples, \cite{merullo2024language} use it to reveal distinct stages of processing in factual recall tasks, and \cite{lepori2025racing} use it to show that final answers are decodable halfway through the network in tasks requiring significant contextualisation (e.g.\ disambiguating whether ``bank'' refers to a financial or geographical entity). A recent \textit{trained} alternative to the logit lens is the tuned lens \citep{belrose2023eliciting}, which fits an affine probe at each layer to predict the final model output, compensating for future layer transformations. We choose to rely on the logit lens here because we are interested in how close the residual stream is at each layer to a state from which the answer can be directly read out, rather than what it would become after further processing.

\textbf{Related work: How do transformers use their depth?}
Understanding what precisely happens at different depths in language models has recently received increased attention. Motivated in part by observations that some layers can be shuffled or removed without significantly degrading performance \citep{gromov2024unreasonable,sun2025transformer}, \cite{lad2024remarkable} conduct an empirical investigation across model families and identify four distinct phases of inference with increasing depth in LLMs : detokenization, feature engineering, prediction ensembling (as features are aggregated into candidate next-token predictions), and residual sharpening (as irrelevant features are suppressed to finalize the prediction). Similarly, \cite{queipo2025attention} identify three stages: an initial mixing stage that builds context, a subsequent compression stage, and a final refinement stage that constructs task-specific representations. \cite{gupta2025llms} posit that earlier layers act as statistical guessers that steer models toward high-probability tokens, while later layers refine predictions by incorporating context.

Most related to our work are two recent papers that study depth use directly. \cite{csordas2025language} question whether LLMs use their depth efficiently, concluding through a series of empirical investigations that deeper models are not using their additional layers for more involved computation. They show that layers in the second half of the model contribute substantially less to the residual stream than those in the first half and that skipping later layers has a smaller effect on predictions. They also find no evidence that models use later layers to compose subresults when multi-hop reasoning is required, using pre-defined difficulty levels (MATH dataset) and  different hop numbers (MQuAKE dataset) and averaging depth scores across all tokens in the generated answers. \cite{hu2025affects} build on these findings and study potential drivers of effective depth, finding no systematic difference in relative depth usage across model sizes or task difficulty, using metrics that track aggregate changes in the residual stream averaged across all answer tokens in HellaSwag (natural language understanding), GSM8K (grade school math), and AIME24 (high school math
contests). We consider a more controlled setup that allows us to vary difficulty more systematically, restricting evaluation to single-token answers, and find evidence that partially diverges from these conclusions.

\section{Empirical investigation: Do LLMs use their depth adaptively in a multi-hop relational reasoning task?}\label{sec:exp}
\begin{figure}[t]
\centering
\small
\begin{minipage}[t]{0.42\linewidth}
\textbf{2-hop example}\\[4pt]
\fbox{\parbox{0.95\linewidth}{%
Michael has a son called John.\\
John has a sister named Elizabeth.\\[2pt]
Therefore, Elizabeth is Michael's \textcolor{orange}{\underline{daughter}}
}}
\end{minipage}%
\hfill
\begin{minipage}[t]{0.54\linewidth}
\textbf{5-hop example}\\[4pt]
\fbox{\parbox{0.95\linewidth}{%
Stella is the wife of Kenneth.\\
Gregory is Stella's son.\\
Gregory has a son called Jim.\\
Adrian is a brother of Jim.\\
Juan is Adrian's brother.\\[2pt]
Therefore, Juan is Kenneth's \textcolor{orange}{\underline{grandson}}
}}
\end{minipage}
\caption{\textbf{Example CLUTRR stories} \small for 2-hop and 5-hop reasoning. Each sentence states a family relation between two people. The model must compose the chain of relations to infer the \underline{target relationship} between the query pair.}
\label{fig:story-examples}
\end{figure}

\subsection{Experimental setup: Task and analyses}\label{sec:exp-setup}
\textbf{The CLUTRR task.} We evaluate relational reasoning using the Compositional Language Understanding and Text-based Relational Reasoning (CLUTRR) data generator\footnote{Adapted from \url{https://github.com/facebookresearch/clutrr/}.} of \citet{sinha2019clutrr}, which generates $k$-hop family stories requiring the composition of a chain of $k$ family relations. We use the simplest version of the generator, which produces stories free of distracting facts or language, ensuring that task difficulty stems solely from resolving family relationships of varying chain length. We use the default generator settings, which create a three-level family tree (three generations) with at most 4 siblings, to generate 2 to 10-hop stories for the main analyses in \cref{sec:pretrained}; for the longer stories in \cref{sec:finetuned}, we need to expand the maximum number of children to 6. With this setup, the possible answers are the tokens $\mathcal{F} = $ \texttt{[mother, father, grandfather, grandmother, son, daughter, grandson, granddaughter, brother, sister, uncle, aunt, nephew, niece]}\footnote{To be able to focus on computation at a single tokens, we restrict our analysis to questions with single-token answers: this is the case for all required relations in $\mathcal{F}$ but excludes the in-law relations that were part of the original generator as they decode to multiple tokens in most models.}. We modify the generator so that all stories end with the sentence \textit{``Therefore, PersonA is PersonB's''}, framing the reasoning task as a next-token prediction problem and allowing us to focus on computation at a single token. Examples of generated stories are shown in \cref{fig:story-examples}. We generate 100 test examples per $k$-hop level.

\textbf{Logit lens analyses of $h_{l,T}$.} Part of our investigation centers on applying the logit lens to decode and analyse the model's hidden states. Specifically, we focus on the probability distribution $\tilde{p}_{l,T}$ obtained by projecting the hidden state at the final token position $T$ (the ``'s'' token) through the language modeling head, as defined in \cref{eq:logitlens}. We begin by verifying that analysing hidden states in this way yields meaningful signal, by monitoring the total probability mass assigned to the set of permitted family relation tokens at each layer: $p_l^{\text{fam}} = \sum_{w \in \mathcal{F}} \tilde{p}_{l,T}(w)$. We then monitor whether the correct answer is identified at each layer via two metrics: (i) \textit{correctness}, whether the token assigned the highest overall probability ($\arg\max_{w \in \mathcal{V}} \tilde{p}_{l,T}(w)$) is the correct answer; and (ii) \textit{constrained correctness}, whether the top-ranked token among the relation tokens ($\arg\max_{w \in \mathcal{F}} \tilde{p}_{l,T}(w)$) is the correct answer. In addition to monitoring $h_l$ through the logit lens, we replicate part of the $h_l$ analyses in \cite{csordas2025language,hu2025affects} by tracking the relative contribution of each layer $\Delta_l=h_l-h_{l-1}$ to the residual stream $\frac{||\Delta_l||_2}{||h_{l-1}||_2}$ and the similarity of the update to the residual stream $cossim(\Delta_l, h_{l-1})$ in Appendix \ref{app:repr-metrics}.

\textbf{Causal patching analyses.}
To better understand where and how information is processed across network depth, we additionally conduct \textit{causal patching experiments} \citep{meng2022locating} with counterfactual relationship replacements, inspired by the setup that \cite{wu2025transformers} use to study how transformers learn variable binding. Given a test story, we replace a single relationship token $t^r=a$ with a counterfactual relation of the same gender $t^r=b$, changing also the ground truth query relationship. We compute the full hidden states associated with original prompt $h^a_l$ and mutated prompt $h^{b}_l$. A causal patching experiment then proceeds by examining the effect of intervening on the hidden state $h^{b}_{l^*,i}$ of the modified story at a single layer $l^*$ and a single token index $i$ by replacing it with the corresponding original $h^a_{l,i}$ and then running the model forward from that point. That is,
\begin{equation*}
    h^{patch_{l^*, i}: b \rightarrow a}_{l,j} = \left\{ \begin{array}{l @{\;} l}
        h^{b}_{l,j} 
            & \text{if } l < l^* \text{ or } j < i \text{ or } (l = l^* \text{ and } j \neq i) \\[6pt]
        h^{a}_{l,i} 
            & \text{if } i = j \text{ and } l = l^* \\[6pt]
        \bigl[h^{patch_{l^*, i}: b \rightarrow a}_{l-1} + f_l\!\left(h^{patch_{l^*, i}: b \rightarrow a}_{l-1}\right)\bigr]_j 
            & \text{if } l > l^* \text{ and } j \geq i
    \end{array} \right.
\end{equation*}

We track whether this intervention recovers the original model prediction at the final token $T$ using a normalized logit difference of the original predicted token $o$ and the new prediction $c$ similar to \cite{zhang2023towards, wu2025transformers} as 
\begin{equation*}
    Rec^{patch_{l^*, i}: b \rightarrow a}=\max\{0, \frac{\text{ld}^{patch_{l^*, i}:  b \rightarrow a}_T(o, c) - \text{ld}^{b}_T(o, c) }{\text{ld}^{a}_T(o, c) - \text{ld}^{b}_T(o, c)}\} \text{ where } \text{ld}^{(\cdot)} = \text{logit}^{(\cdot)}_T[o]-\text{logit}^{(\cdot)}_T[c]. 
\end{equation*}

Intuitively, this allows us to track how information moves across sequence positions and depth. To see this, note that patching the embedding ($l=0$) at the modified token $t^r$ would restore the prediction perfectly, as would patching the final-layer hidden state at the final token $T$. Being able to recover the original prediction by patching at layer $l^*>0$ at the modified token ($i=t^r$) then means that the semantic information in this token has either not yet propagated to future tokens, or is retrieved by the final token $T$ directly from position $t^r$. Conversely, recovering the original prediction by patching at the final output token ($i=T$) before the last layer ($l^*<L$) implies that all information about the modified token has already been integrated at the final position, with no future layers integrating further conflicting information about $t^r$ from non-patched positions. Being able to recover the original prediction by patching hidden states at token $i$ between $t^r$ and $T$ and layer $l$  indicates that information about $t^r$ flows to $T$ via $i$ at layer $l$. 

Note that the recovery score is meaningful only when the model changes its prediction in response to the counterfactual (which is the correct behavior, since the true answer changes). We therefore sample only stories for which the model's top prediction changes under the counterfactual, regardless of whether either prediction is correct. To ensure that all counterfactual stories have valid single-token answers, we start with simple stories containing only brother/sister relationships (correct answer: brother/sister) and intervene by replacing them with father/mother relationships at intermediate tokens, or uncle/aunt relationships at the first and last tokens (correct answer in both cases: uncle/aunt). Due to the high computational cost of patching analyses -- which require $k \times T \times L$ forward passes per example -- we restrict the analysis to a subset of models, hop counts, and examples throughout, using $n=30$ examples per configuration when available\footnote{For some model--hop combinations, fewer than 30 examples flip their prediction upon the intervention.}. In Appendix \ref{app:morestories}, we repeat the analyses for the original stories with more involved relationship chains and instead replace existing relations with brother/sister regardless of whether this leads to valid target answers and find consistent trends.

\subsection{Do pretrained transformers use their depth adaptively?}\label{sec:pretrained}
This section analyses how pretrained models use their depth on the family relations task. We consider five open-weights model families\footnote{Note that we use the stories as prompts exactly as presented in \cref{fig:story-examples} in order to constrain computation to a single token; we are therefore \textit{not} using the chat template that some of these models are post-trained to expect.} (GPT-2, Pythia, Phi, Qwen2, Qwen2.5, and LLaMA-3), spanning a range of sizes up to 14B parameters. The upper end of this range is determined by the computational cost and memory footprint of the analyses; within this constraint, we select sizes that allow us to study trends within each family. This gives us a diverse set of architectures and training recipes while keeping the analyses tractable.

\begin{figure}[t]
\vspace{-3em}
    \centering
    \includegraphics[width=0.9\linewidth]{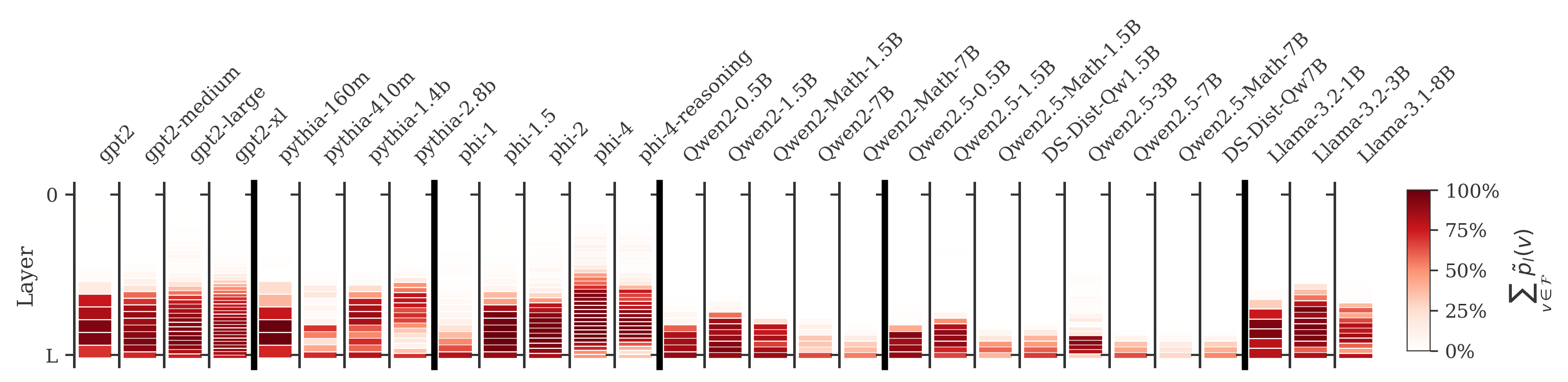}
    \caption{\textbf{Probability assigned to family relation tokens} \small when decoding the hidden states of the final token using the language modeling head directly, by layer across different pretrained model sizes and families, averaged across hops. Family relations are decodable at layer $l<L$ across all models.}
    \label{fig:family_rels}
\end{figure}

\textbullet{}{}\textbf{ Observation 1: \textit{Family relations are decodable from around two-thirds depth in all models; final layers appear to diffuse predictions.}} We begin by confirming whether answers can be decoded from early hidden states using the language model head directly, as shown in \cref{fig:family_rels}. Strikingly, in all models there is a layer around two-thirds of the way through the network from which the logit lens decodes into family relationships with very high probability. This aligns precisely with the observation in \cite{lad2024remarkable} that the early layers are responsible for detokenization and feature engineering: in our setting, after approximately two-thirds of the layers, the model has arrived at a region of the hidden space that is semantically meaningful for the task. The precise depth at which relations become decodable varies across model families but is consistent within them. Qwen2.5 models are decodable the latest, while Pythia models exhibit non-monotonic patterns.

Phi-4, by contrast, is decodable at the earliest point. We also observe that the probability mass assigned to family tokens dips in the final layers across almost all models. As shown in \cref{fig:entropy} in the appendix, the entropy of the induced distribution first decreases within the decodable layers, then rises again in the final layers\footnote{This is particularly striking for the Phi-4-Reasoning model, where we observe up to 50\% accuracy on the 10-hop task at intermediate layers, dropping back to 10\% at the final layer, where the top token is often ``\_\_\_'', particularly for longer stories (see e.g. the example in \cref{fig:patching-phi}). We believe this may be partly attributable to this model being post-trained to expect certain formats, which we are not using.}. This is interesting because it suggests that the final layers are not simply refining the predicted answer but also restructuring the distribution over answers, perhaps to achieve better calibration. A similar effect is observed by \cite{lv2024interpreting} in the context of factual recall, who conclude that the final layer has an ``anti-overconfidence" effect.

\begin{figure}[t]
\vspace{-2em}
  \centering
  \begin{subfigure}{0.8\textwidth}
    \centering
    \includegraphics[width=\linewidth]{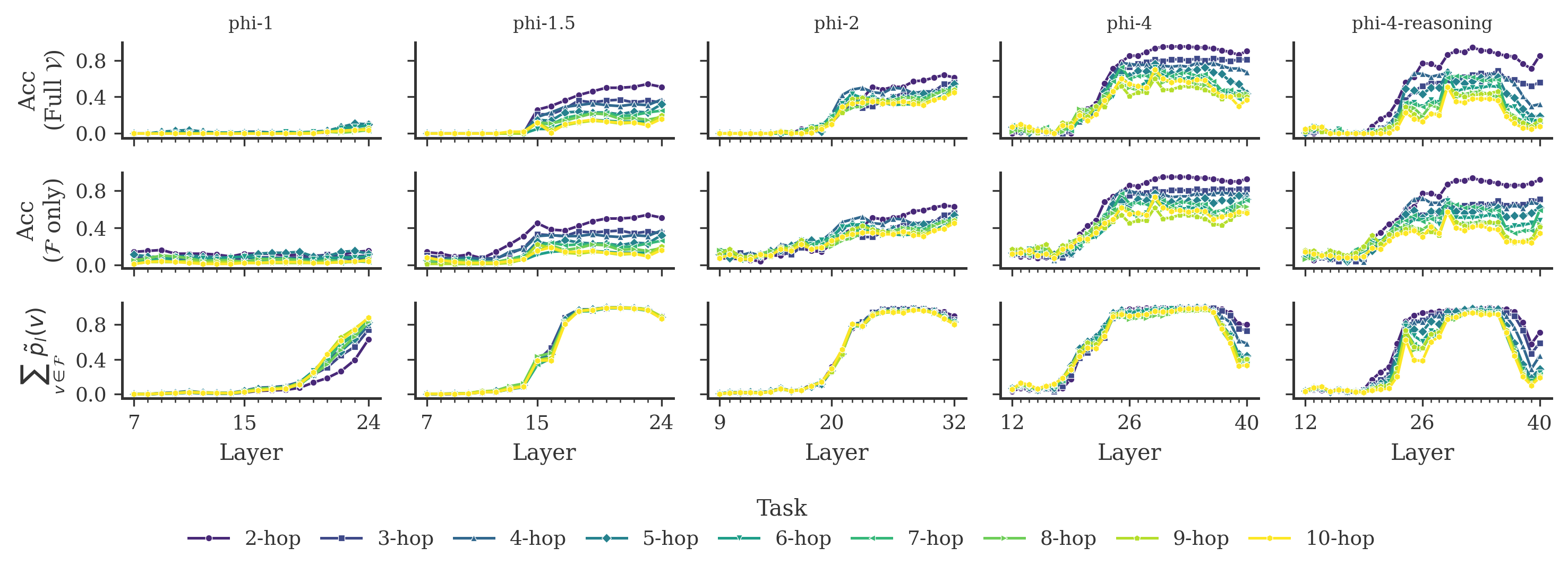}
    \vspace{-1.5em}
    \caption{Phi Family}
    \label{fig:a}
  \end{subfigure}%
  \hfill
  \begin{subfigure}{0.8\textwidth}
    \centering
    \includegraphics[width=\linewidth]{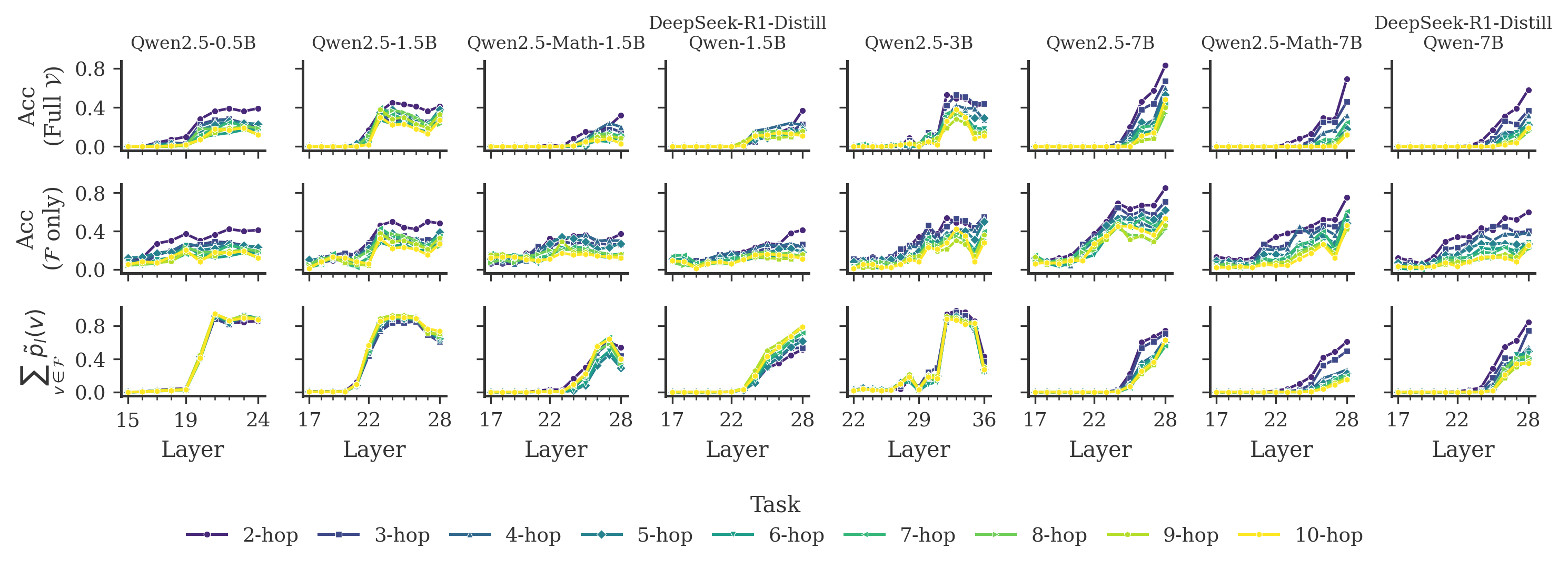}
      \vspace{-1.5em}
    \caption{Qwen2.5 Family}
    \label{fig:b}
  \end{subfigure}
  \vspace{-1em}
  \caption{\textbf{Logit lens results for Phi and Qwen2.5 families.} \small Correctness, constrained correctness and probability assigned to family relation tokens by logit lens predictions by layer, colored by hops. X-axes (layers) are left-truncated for better readability; previous accuracies \&  $p^{\text{fam}}_l$ are zero.}
  \label{fig:correctness}
\end{figure}

\begin{figure}[t]
  \centering
  \begin{subfigure}{0.99\textwidth}
    \centering
       \includegraphics[width=\linewidth]{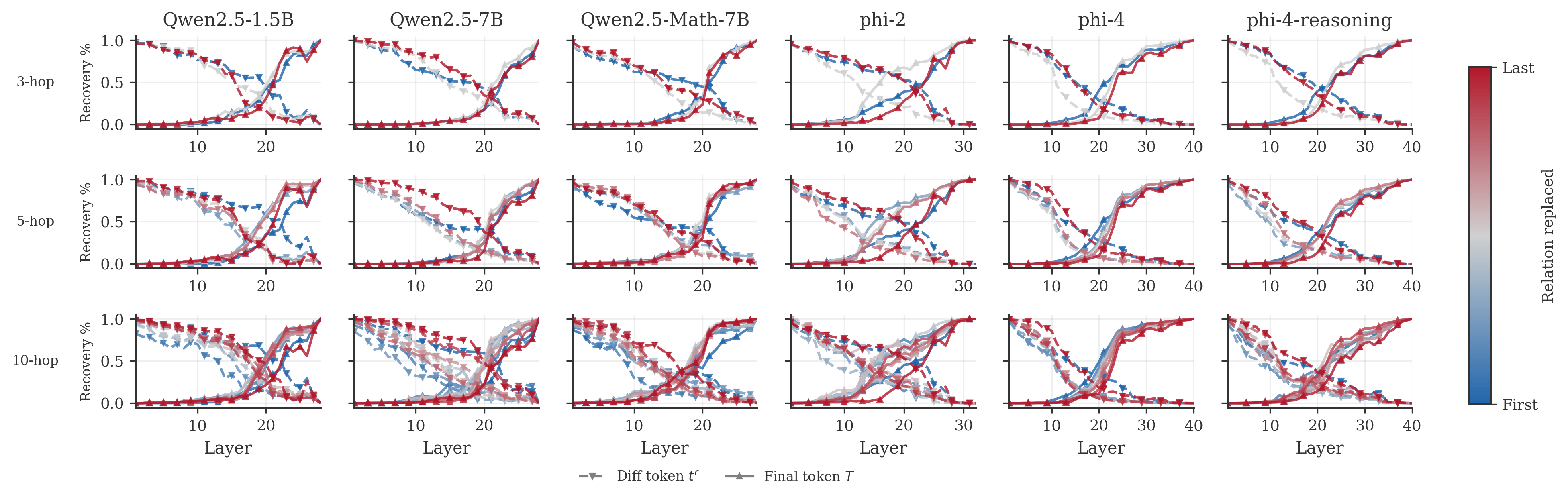}
       \vspace{-1.5em}
    \caption{Average recovery score by replaced relation across different hops (rows), colored by relation replaced.}
    \label{fig:a}
  \end{subfigure}%
  \hfill
  \begin{subfigure}{0.99\textwidth}
    \centering
     \includegraphics[width=\linewidth]{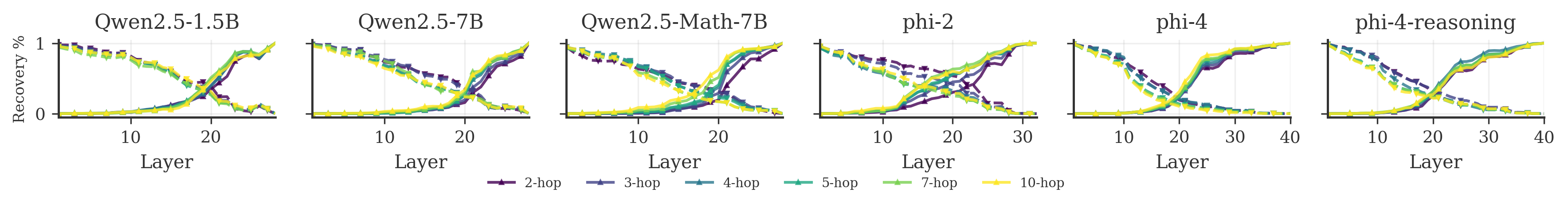}
            \vspace{-1.5em}
    \caption{Average recovery score across all relation replacement positions, colored by number of hops.}
    \label{fig:b}
  \end{subfigure}
    \vspace{-1em}
  \caption{\textbf{Causal patching results.} Average recovery score at the replaced token $t^r$ (dashed lines) and the final token $T$ (solid lines) by model depth.}
  \label{fig:cp-pretrained}
\end{figure}

\textbullet{}{}\textbf{ Observation 2: \textit{Larger models need less depth to arrive at plausible answers for easier tasks; smaller models show little such differentiation.}} We now turn to the core question of whether pretrained models use more layers to arrive at correct answers for harder tasks, by examining decoded answer accuracy and family token probability per layer and per task. We observe that the answer to this may depend on model size: In \cref{fig:correctness}, we consider models in the Phi and Qwen2.5 families (results for other model families are included in \cref{app:logitlens-pre}). For most smaller models, accuracy and family token probability begin to rise at the same layer across all tasks: while they reach different absolute levels, they do so at similar rates, and absolute performance is poor across the board. For the largest models, a somewhat different pattern emerges: in Phi-4-Reasoning, the Qwen2.5-7B and Qwen2-7B models and Llama-3.1-8B, top-token accuracy and family token probability for easier tasks begin to rise several layers before they do for harder tasks, suggesting that these models may indeed need more depth to process stories with longer relationship chains.

 \textbullet{}{} \textbf{ Observation 3: \textit{Longer relationship chains generally induce earlier cross-token information integration.}} We complement the logit lens analyses, which investigate when information becomes decodable across depth, with causal patching experiments, which provide insight into when information is propagated to and integrated at the final token. In \cref{fig:cp-pretrained}, we focus on patching at (i) the replacement token $t^r$ and (ii) final token $T$, to investigate whether contextual information about $t^r$ is (i) processed and (ii) integrated into the final answer at different depths for tasks of different difficulty. We present examples of full patching trajectories in \cref{app:cp-pre}.  Across the considered models, we find that information gets processed at earlier layers at the replacement token $t^r$ for longer stories (especially relationship tokens in the middle of longer stories in panel (a)) as indicated by recovery scores dropping at earlier layers, meaning that information is mixed into the residual streams of future tokens sooner for longer stories\footnote{An exception to this rule is the pretrained gpt2-large in \cref{fig:cp-gpt2} (b, left), which does not appear to integrate information earlier at $t^r$ for longer chains but instead integrates information later at $T$.}. Interestingly, recovery scores at the final token $T$ also begin to rise earlier on average for longer stories (panel (b)), suggesting that information is integrated at the final position earlier, potentially allowing more layers for subsequent answer refinement. While the former pattern (earlier mixing at $t^r$) appears also in the stories with more relationships in \cref{fig:cp-pretrained-allrels}, the latter (earlier integration at $T$) does not. Beyond these trends, there appear to be limited consistent patterns to the order in which information about individual relations is integrated across models, though most models integrate information about the last relationship last. Qwen2.5 models appear to fully integrate information about the first relationship at $T$ later than Phi models for longer stories.  %

Overall, the results in this section indicate that pretrained models do show some variation in depth use across task difficulty: larger models appear to need fewer layers to arrive at plausible answers for easier stories, and models generally use more layers to integrate information across tokens as chain length increases. This finding stands in partial contrast to \cite{hu2025affects}. We hypothesize that their setup was not sufficiently controlled to detect these differences: task difficulty was more loosely defined, and their metrics were averaged over many tokens. Moreover, the representation similarity metrics they employ (see Appendix \ref{app:repr-metrics}) are harder to interpret than the logit lens and causal patching analyses we conduct here, which are made possible precisely by the controlled nature of the task and the restriction to single-token answers.

\subsection{Do transformers finetuned to solve the task use their depth adaptively?}\label{sec:finetuned}
\begin{figure}[t]
 \vspace{-2em}
    \centering
    \includegraphics[width=0.990\linewidth]{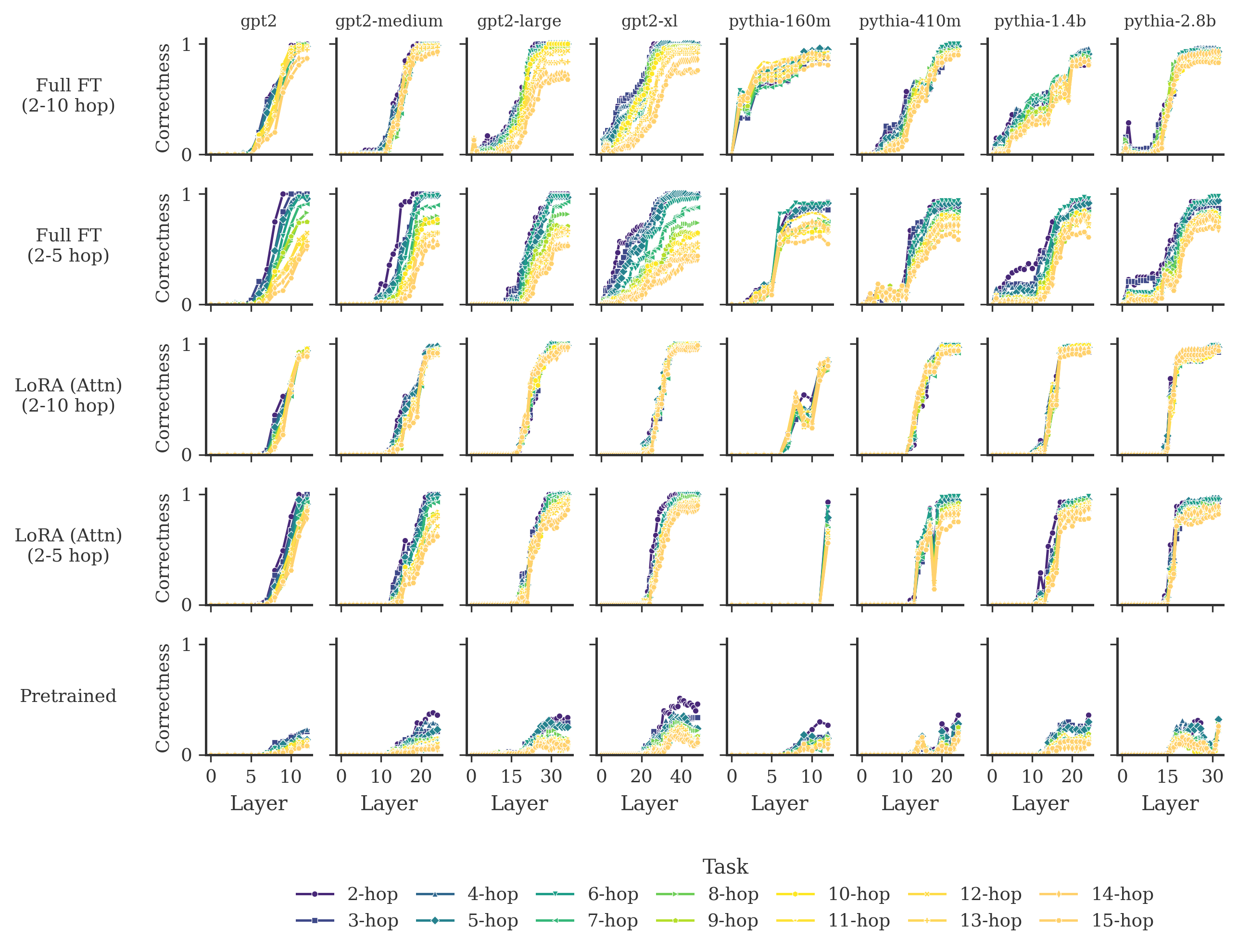}
    \vspace{-1em}
    \caption{\textbf{Accuracy of the top logit lens prediction} across models (columns) and training regimes (rows) by layer and colored by number of hops.}
    \label{fig:compare-correct}
\end{figure}
\textbf{Finetuning setup.}  Next, we finetune the GPT-2 and Pythia model families to solve the task, allowing us to investigate whether models specialised for the task \emph{learn} to use their depth adaptively. We consider two finetuning regimes: (i) LoRA finetuning \citep{hu2022lora} of only parameter matrices in the attention modules, and (ii) full finetuning of the whole model. In both cases, we supervise using the loss on the answer token only (completion-only supervision). We train on 1000 examples consisting of an equal mixture of (i) 2--5-hop and (ii) 2--10-hop stories for 20 epochs. Unless indicated otherwise, we present results for models finetuned on 2--10-hop stories by default.

Our test set evaluates generalisation in two ways. First, within-task generalisation is assessed using the train--test split functionality of the CLUTRR generator, which ensures that the specific combinations of relations in the test set were not seen during training. Second, to test length generalisation, we evaluate on 11--15-hop stories, for which we need to increase the maximum number of allowed siblings to 6. Note that, as the CLUTRR generator is restricted to three-generation family trees, longer stories may contain structural shortcuts and necessarily include a higher proportion of sibling relationships. Finally, since family trees are shared across examples in the CLUTRR generator (making it possible in principle to solve the original task by memorising relations by name rather than reasoning about them) we rename all individuals to Person1 through PersonK and randomise the order in which numerical indices appear. Further implementation details are provided in Appendix \ref{app:impl}.

\textbullet{}{} \textbf{Observation 4: \textit{All models learn the task, but only LoRA finetuned models length-generalise and preserve general language modeling ability; full finetuning causes earlier decodability but does not length generalise.}}  We begin by considering model performance and make a number of interesting observations. Examining correctness in \cref{fig:compare-correct}, all models regardless of size and finetuning regime successfully learn to solve stories of in-domain length not seen during training. Second, LoRA finetuning yields models that length-generalise to longer hop counts without performance loss, while full finetuning results in degraded performance on longer stories. Third, and perhaps unsurprisingly, restricting parameter updates to the attention heads via LoRA leaves the semantic structure of the residual stream largely unchanged: family relations remain decodable from the language modeling head at a similar depth as in the base models. Full finetuning, by contrast, causes family relations to become decodable much earlier, particularly in larger models (see also \cref{fig:compare-probs} in the Appendix). It is worth noting that aggressively finetuning on the final answer token only causes the fully finetuned models to lose their general language modeling ability entirely, while LoRA finetuned models retain it; see \cref{tab:perplexity} in Appendix \ref{app:perplexity}. This observation is in line with \cite{biderman2024lora} who find that LoRA learns \textit{and} forgets substantially less than full finetuning.

\begin{figure}[t]
\vspace{-2em}
  \centering
  \begin{subfigure}{0.48\textwidth}
    \centering
    \includegraphics[width=\linewidth]{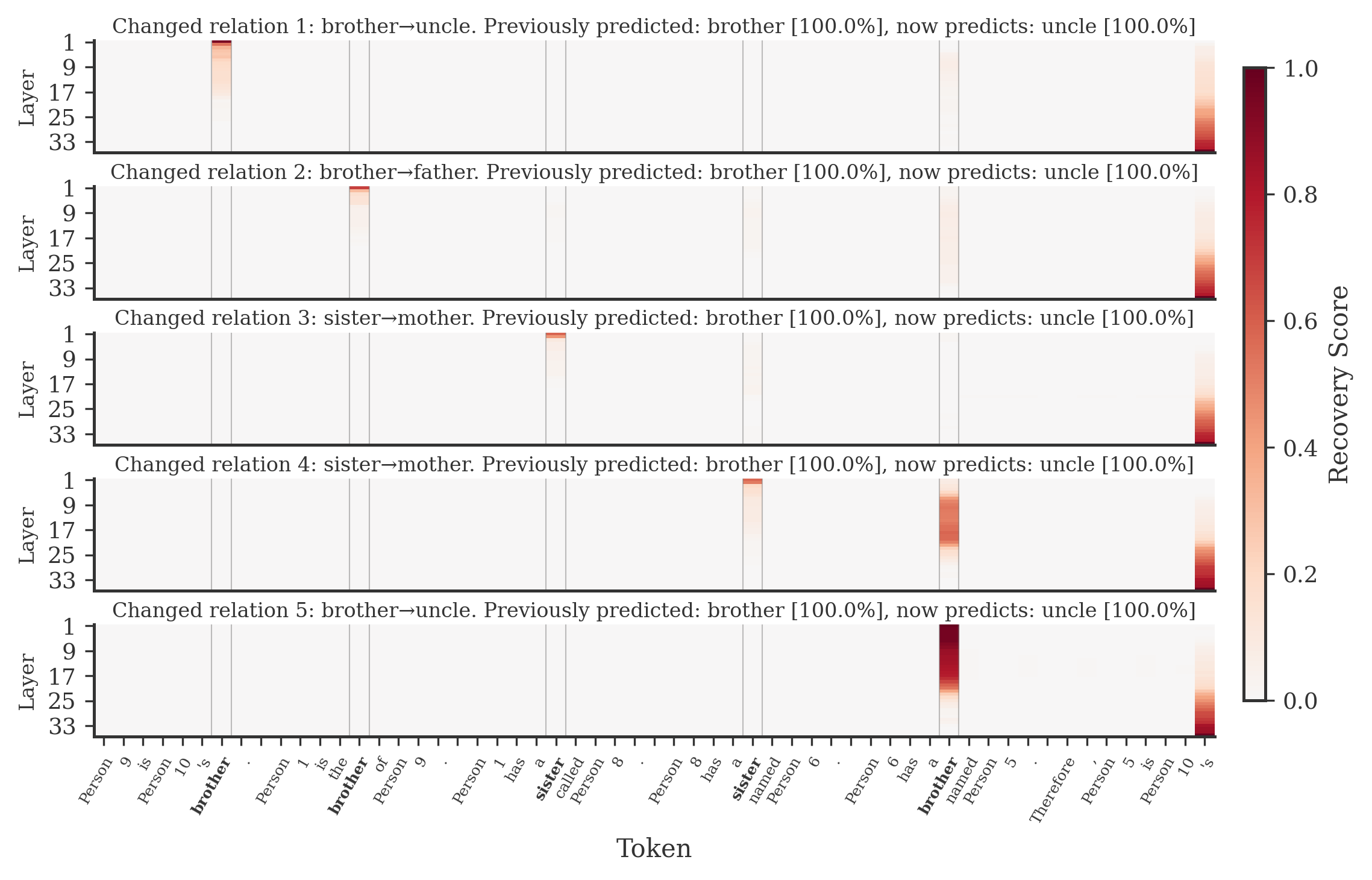}
     \vspace{-1.5em}
    \caption{Patch: $uncle \rightarrow brother$, full finetuning}
    \label{fig:a}
  \end{subfigure}%
  \hfill
  \begin{subfigure}{0.48\textwidth}
    \centering
    \includegraphics[width=\linewidth]{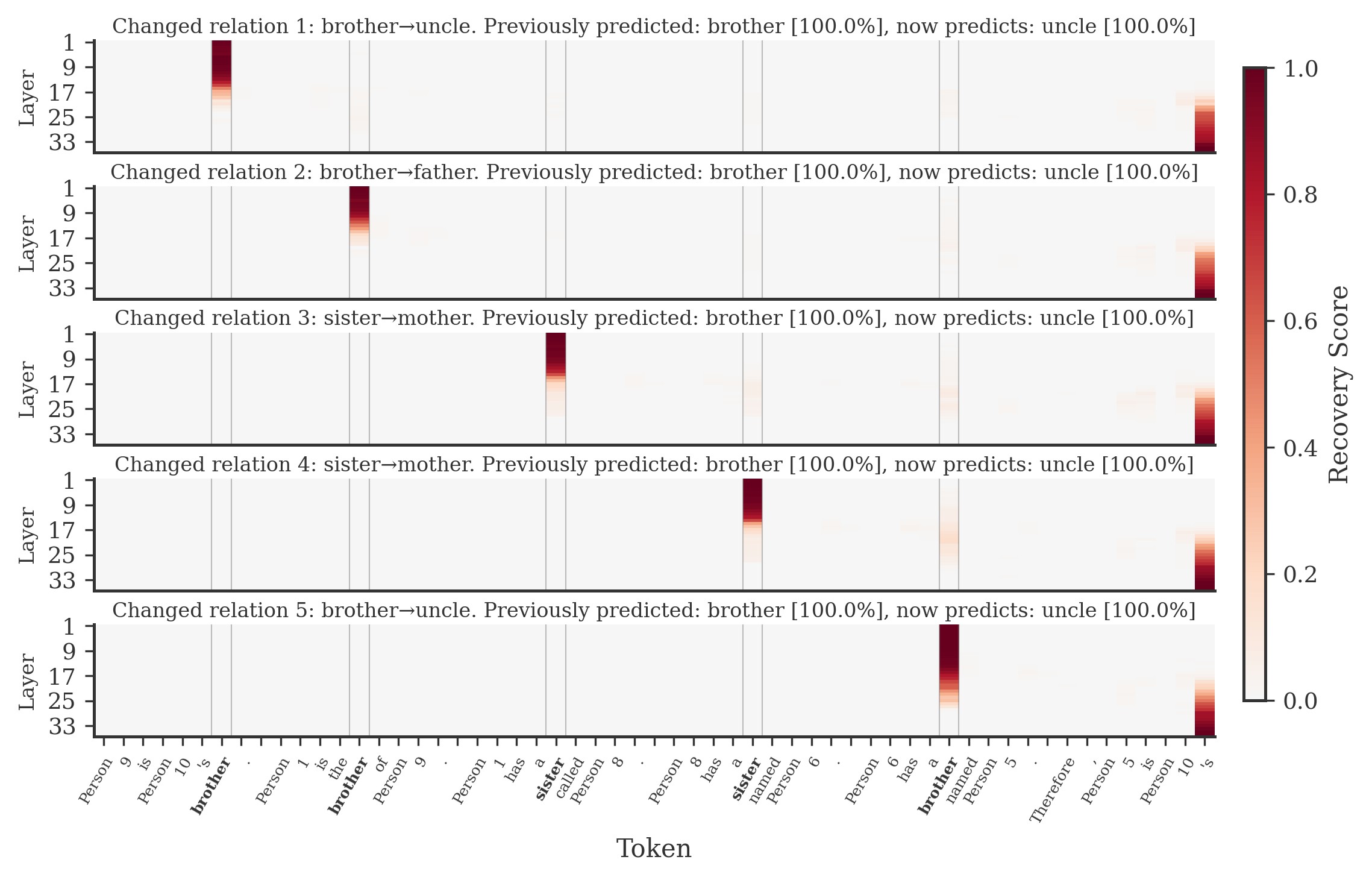}
        \vspace{-1.5em}
    \caption{Patch:  $uncle \rightarrow brother$, LoRA finetuning}
    \label{fig:b}
  \end{subfigure}
    \begin{subfigure}{0.48\textwidth}
    \centering
    \includegraphics[width=\linewidth]{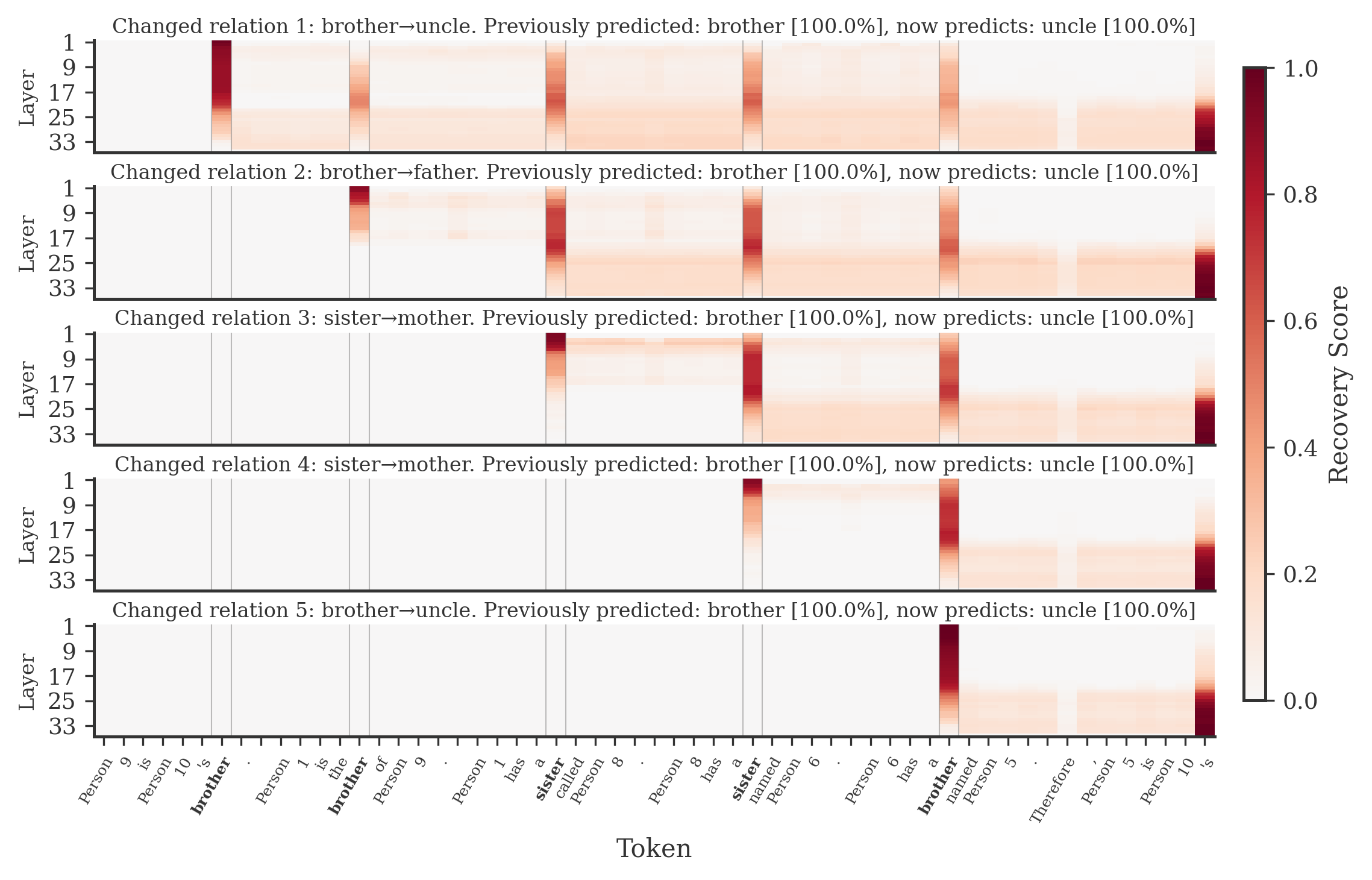}
    \vspace{-1.5em}
    \caption{Patch $brother \rightarrow uncle$, full finetuning}
    \label{fig:a}
  \end{subfigure}%
  \hfill
  \begin{subfigure}{0.48\textwidth}
    \centering
    \includegraphics[width=\linewidth]{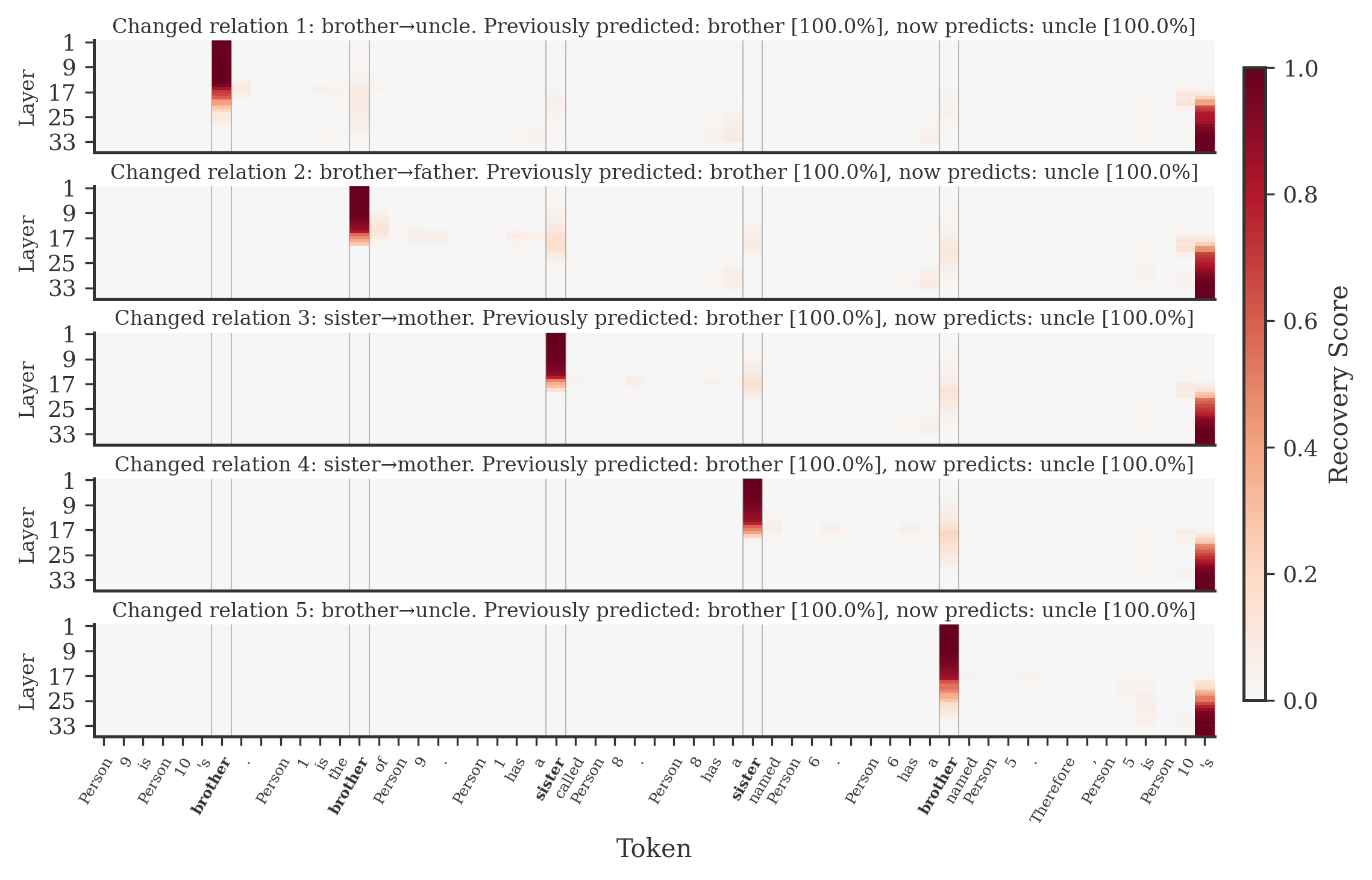}
        \vspace{-1.5em}
    \caption{Patch $brother \rightarrow uncle$, LoRA finetuning}
    \label{fig:b}
  \end{subfigure}
    \begin{subfigure}{0.48\textwidth}
    \centering
    \includegraphics[width=\linewidth]{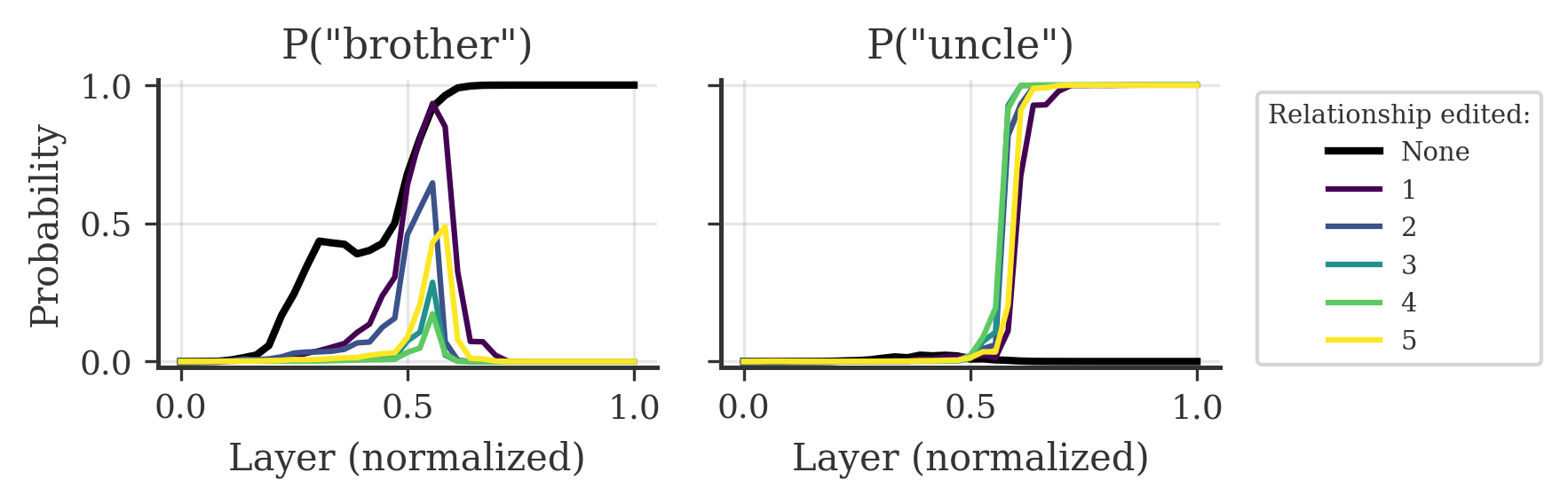}
        \vspace{-1.5em}
    \caption{Logit Lens: Full finetuning}
    \label{fig:a}
  \end{subfigure}%
  \hfill
  \begin{subfigure}{0.48\textwidth}
    \centering
    \includegraphics[width=\linewidth]{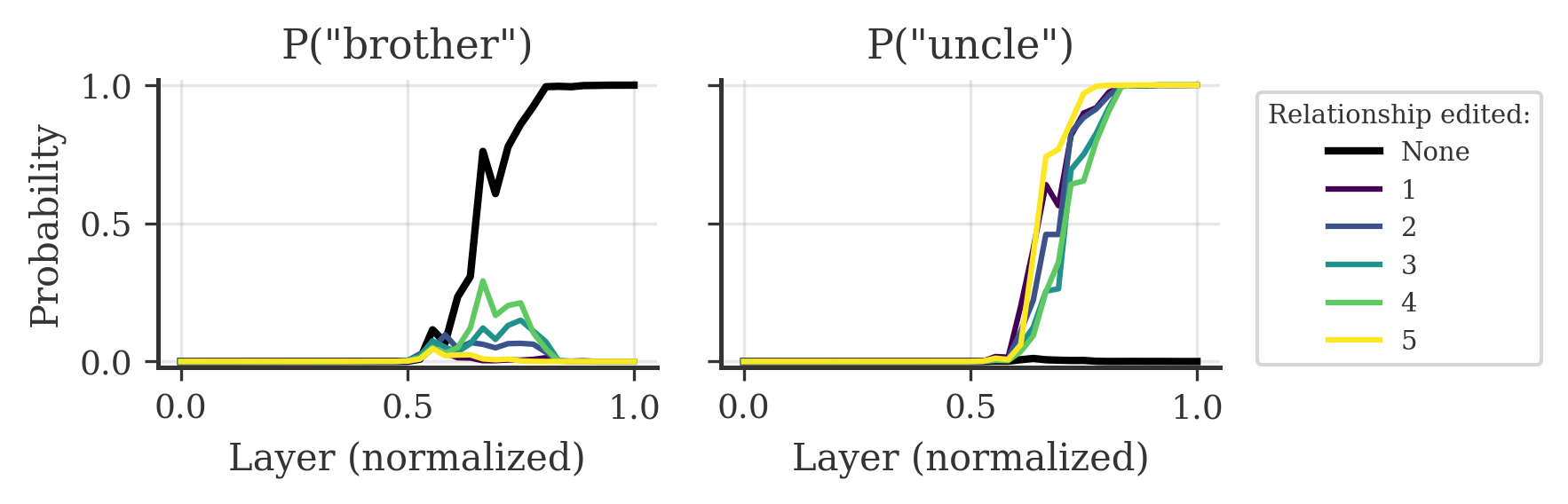}
        \vspace{-1.5em}
    \caption{Logit lens: LoRA finetuning}
    \label{fig:b}
  \end{subfigure}
  \caption{\textbf{Full causal patching and logit-lens trajectories} \small for a 5-hop example of finetuned GPT2-large versions.}
  \label{fig:causalpatch}
\end{figure}

\textbullet{}{} \textbf{Observation 5: \textit{Fully finetuned models show evidence for adaptive depth use in logit lens decoding; LoRA finetuned models do not.}} Turning to the key question of whether models learn to use their depth adaptively, the logit lens results in \cref{fig:compare-correct} indicate a clear split between the two regimes. The LoRA finetuned models remain closely aligned with their base model counterparts, solving tasks at similar depths across all hop counts, providing little evidence of depth use adapting to task difficulty. The fully finetuned models tell a different story: accuracy for harder tasks rises later in the network, especially for larger models, suggesting that these models do learn to process the task more iteratively with depth. This may also explain why fully finetuned models fail to length-generalise: if a model has learned to use its full depth for the hop counts seen during training, it has no capacity in reserve for longer stories.

\begin{figure}[t]
\vspace{-2em}
  \centering
  \begin{subfigure}{0.99\textwidth}
    \centering
       \includegraphics[width=\linewidth]{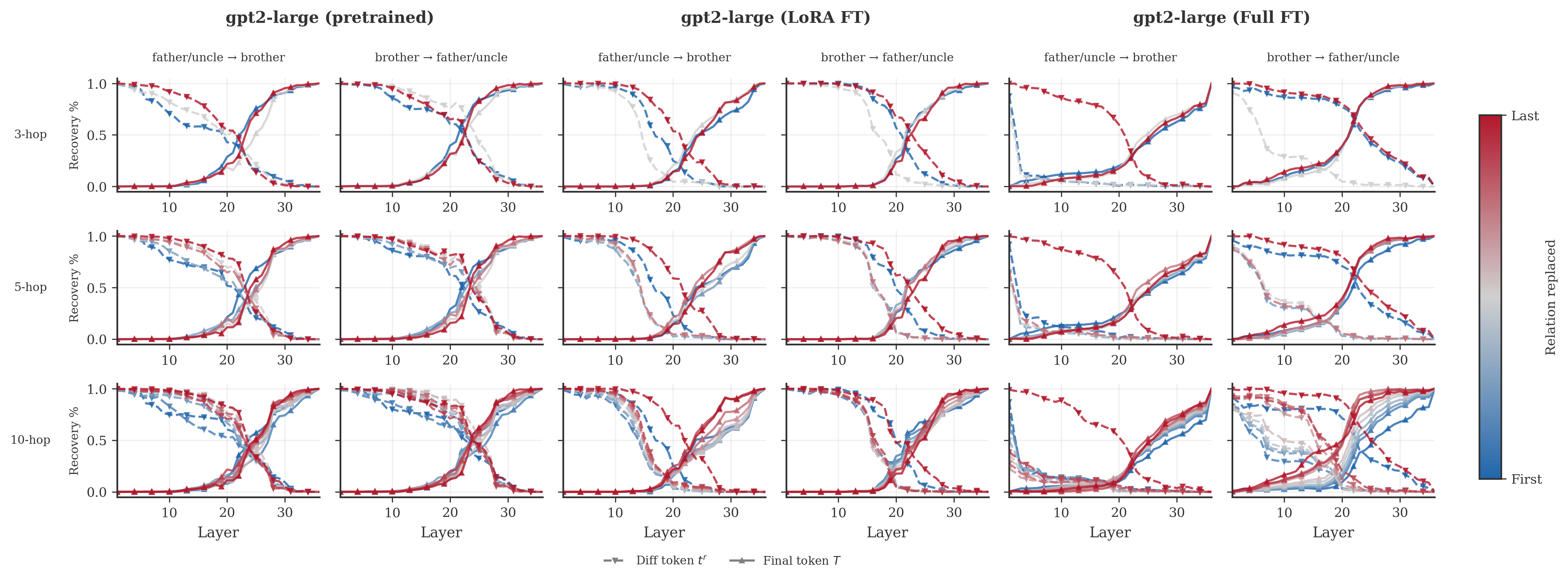}
       \vspace{-1em}
    \caption{Average recovery score by replaced relation across different hops (rows), colored by relation replaced.}
    \label{fig:a}
  \end{subfigure}%
  \hfill
  \begin{subfigure}{0.99\textwidth}
    \centering
     \includegraphics[width=\linewidth]{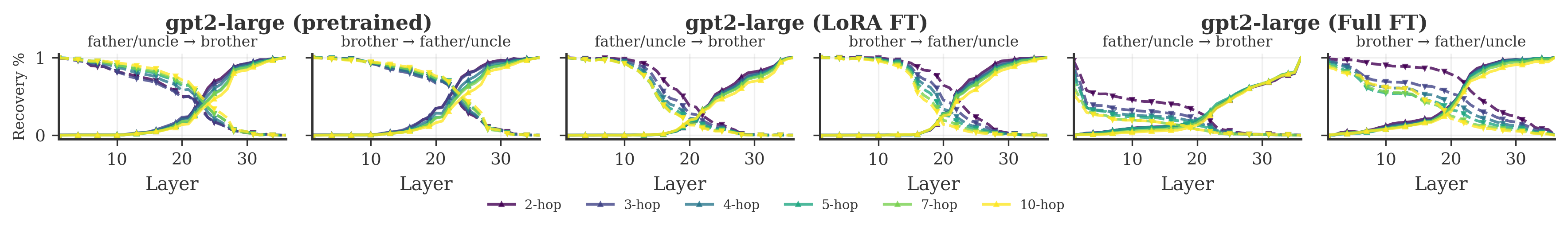}
     \vspace{-1em}
    \caption{Average recovery score across all relation replacement positions, colored by number of hops.}
    \label{fig:b}
  \end{subfigure}
  \caption{\textbf{Causal patching results for GPT2-large}, pretrained and finetuned. Average recovery score at the replaced token $t^r$ (dashed lines) and the final token $T$ (solid lines) by model depth.}
  \label{fig:cp-gpt2}
\end{figure}

\textbullet{}{} \textbf{Observation 6: \textit{All finetuned models learn to speculatively integrate relationship information across tokens, with fully finetuned models devoting more of their total depth to this.}} We further investigate depth use via causal patching. Unlike pretrained models, which were not trained on the task and are unlikely to anticipate the final query, finetuned models could in principle learn to use the hidden representations of all tokens $t<T$ to speculatively resolve relationship compositions as they appear in the text. This is precisely what we observe in \cref{fig:causalpatch} (see Appendix \ref{app:cp-fine} for additional hop counts): information about earlier relationships appears to be integrated into the residual streams of subsequent family tokens, as well as some intermediate tokens. The fully finetuned models learn to use the residual stream of all tokens at all depths for this purpose, while the LoRA finetuned models retain the detokenization phase of the base model, during which no semantic information is mixed across tokens. Recovery scores at intermediate family tokens are larger for the fully finetuned models than for the LoRA models, indicating that more information is routed via intermediate tokens. Strikingly, patching the hidden states of the modified story from uncle back to brother is much less successful than patching in the opposite direction. In \cref{fig:attn-patterns} in the Appendix we show that this is because the models have learned to attend differently (generally less) to brother/sister relations---likely because these are easier to resolve\footnote{In most cases, adding a sibling relationship to a chain does not change the final answer, whereas adding a parent relationship will.}---while uncle/father relations receive substantially more attention at future layers and tokens. Patching a single hidden state from uncle to brother therefore has little effect, since information about the harder relationship has already propagated forward; the reverse patch is far more effective precisely because limited information about the easier relation would have been passed forward in the first place.

\textbullet{}{} \textbf{ Observation 7: \textit{Both finetuning regimes use more depth for harder tasks, but fully finetuned models use substantially more depth overall for cross-token mixing and show clearer ordering of information integration.}} In \cref{fig:cp-gpt2}(a), we aggregate across more patching trajectories and confirm that these patterns generalise (with results for other models in Appendix \ref{app:cp-fine}). The LoRA finetuned models use approximately the same range of layers as the pretrained models to mix information across tokens, but change the order in which information is integrated, incorporating information about the first relationship substantially later. The fully finetuned models use much more of their depth for cross-token mixing (i.e. layers where recovery scores at both $t^r$ and $T$ are low), and devote more depth to processing some relationships (father/uncle) than others (siblings). We also find that the fully finetuned models most clearly integrate (i) information about the first relation last and (ii) information about the last relation first into the final prediction, while in the LoRA case this separation is less pronounced. In \cref{fig:cp-gpt2}(b), we examine how information integration varies with task difficulty. Both the LoRA and fully finetuned models learn to use more of their depth for harder tasks, mixing information earlier and completing integration later for longer chains.

\section{Conclusion and Discussion}
In this work, we provide empirical evidence that transformers can adapt their use of depth to task difficulty, using a controlled multi-hop relational reasoning task based on family stories. Investigating pretrained models, we find that some models use fewer layers to arrive at plausible answers for easier stories, and that models generally use more of their depth to integrate information across tokens for longer relationship chains. Investigating finetuned models, we find that models trained directly on the task use more layers to integrate information for harder tasks, and alter the order in which information is integrated relative to the base model. Models finetuned more aggressively, without regard for preserving general language modeling ability, use substantially more of their total depth for the task compared to those finetuned with LoRA on attention parameters only, and show clearer evidence of arriving at plausible answers earlier for easier tasks.

While our results provide some evidence of adaptive depth use in pretrained models, our finetuning experiments also highlight that the need to preserve general language modeling ability likely constrains which layers are available to perform additional computation for harder reasoning tasks (e.g. needing to reserve depth for the apparent detokenization function of earlier layers). At the same time, our decodability results reveal that different model families arrive at a semantically meaningful region of the residual stream (one that is directly decodable via the language model head) at markedly different depths. Since these models are architecturally similar, this suggests that differences in training give rise to different pressures for the residual stream to remain aligned (or become misaligned) with the output embeddings. What drives this (mis)alignment, and whether the proportion of the network devoted to each stage of inference is steerable, are interesting questions for future work.

\bibliography{bib}
\bibliographystyle{iclr2026_conference}
\newpage
\appendix
\section{Implementation details}\label{app:impl}
All models are accessed through the huggingface transformers library \citep{wolf2020transformers}; see \cref{tab:models} for their links. The logit lens analyses were implemented by accessing the intermediate hidden states and LM-head modules using the functionalities present in the transformers library directly, while the causal patching analyses were run using the nnsight \citep{fiottokaufman2024nnsight} library.

\begin{table}[h]
\centering
\small
\caption{All pretrained models used in this study, with HuggingFace identifiers, parameter counts, and number of transformer layers.}
\label{tab:models}
\begin{tabular}{llrr}
\toprule
\textbf{Family} & \textbf{HuggingFace identifier} & \textbf{Parameters} & \textbf{Layers} \\
\midrule
\multirow{4}{*}{GPT-2}
  & \href{https://huggingface.co/openai-community/gpt2}{\texttt{gpt2}}                     & 117M  & 12 \\
  & \href{https://huggingface.co/openai-community/gpt2-medium}{\texttt{gpt2-medium}}       & 345M  & 24 \\
  & \href{https://huggingface.co/openai-community/gpt2-large}{\texttt{gpt2-large}}         & 774M  & 36 \\
  & \href{https://huggingface.co/openai-community/gpt2-xl}{\texttt{gpt2-xl}}               & 1.5B  & 48 \\
\midrule
\multirow{4}{*}{Pythia}
  & \href{https://huggingface.co/EleutherAI/pythia-160m-deduped}{\texttt{pythia-160m-deduped}}   & 160M  & 12 \\
  & \href{https://huggingface.co/EleutherAI/pythia-410m-deduped}{\texttt{pythia-410m-deduped}}   & 410M  & 24 \\
  & \href{https://huggingface.co/EleutherAI/pythia-1.4b-deduped}{\texttt{pythia-1.4b-deduped}}   & 1.4B  & 24 \\
  & \href{https://huggingface.co/EleutherAI/pythia-2.8b-deduped}{\texttt{pythia-2.8b-deduped}}   & 2.8B  & 32 \\
\midrule
\multirow{3}{*}{Phi}
  & \href{https://huggingface.co/microsoft/phi-1}{\texttt{microsoft/phi-1}}                     & 1.3B  & 24 \\
  & \href{https://huggingface.co/microsoft/phi-1_5}{\texttt{microsoft/phi-1.5}}                 & 1.3B  & 24 \\
  & \href{https://huggingface.co/microsoft/phi-2}{\texttt{microsoft/phi-2}}                     & 2.7B  & 32 \\
  & \href{https://huggingface.co/microsoft/phi-4}{\texttt{microsoft/phi-4}}                     & 14B   & 40 \\
  & \href{https://huggingface.co/microsoft/phi-4-reasoning}{\texttt{microsoft/phi-4-reasoning}} & 14B   & 40 \\
\midrule
\multirow{5}{*}{Qwen2}
  & \href{https://huggingface.co/Qwen/Qwen2-0.5B}{\texttt{Qwen/Qwen2-0.5B}}                     & 494M  & 24 \\
  & \href{https://huggingface.co/Qwen/Qwen2-1.5B}{\texttt{Qwen/Qwen2-1.5B}}                     & 1.5B  & 28 \\
  & \href{https://huggingface.co/Qwen/Qwen2-Math-1.5B}{\texttt{Qwen/Qwen2-Math-1.5B}}           & 1.5B  & 28 \\
  & \href{https://huggingface.co/Qwen/Qwen2-7B}{\texttt{Qwen/Qwen2-7B}}                         & 7.6B  & 28 \\
  & \href{https://huggingface.co/Qwen/Qwen2-Math-7B}{\texttt{Qwen/Qwen2-Math-7B}}               & 7.6B  & 28 \\
\midrule
\multirow{8}{*}{Qwen2.5}
  & \href{https://huggingface.co/Qwen/Qwen2.5-0.5B}{\texttt{Qwen/Qwen2.5-0.5B}}                 & 494M  & 24 \\
  & \href{https://huggingface.co/Qwen/Qwen2.5-1.5B}{\texttt{Qwen/Qwen2.5-1.5B}}                 & 1.5B  & 28 \\
  & \href{https://huggingface.co/Qwen/Qwen2.5-Math-1.5B}{\texttt{Qwen/Qwen2.5-Math-1.5B}}       & 1.5B  & 28 \\
  & \href{https://huggingface.co/deepseek-ai/DeepSeek-R1-Distill-Qwen-1.5B}{\texttt{deepseek-ai/DeepSeek-R1-Distill-Qwen-1.5B}} & 1.5B & 28 \\
  & \href{https://huggingface.co/Qwen/Qwen2.5-3B}{\texttt{Qwen/Qwen2.5-3B}}                     & 3.1B  & 36 \\
  & \href{https://huggingface.co/Qwen/Qwen2.5-7B}{\texttt{Qwen/Qwen2.5-7B}}                     & 7.6B  & 28 \\
  & \href{https://huggingface.co/Qwen/Qwen2.5-Math-7B}{\texttt{Qwen/Qwen2.5-Math-7B}}           & 7.6B  & 28 \\
  & \href{https://huggingface.co/deepseek-ai/DeepSeek-R1-Distill-Qwen-7B}{\texttt{deepseek-ai/DeepSeek-R1-Distill-Qwen-7B}} & 7.6B & 28 \\
\midrule
\multirow{3}{*}{LLaMA}
  & \href{https://huggingface.co/meta-llama/Llama-3.2-1B}{\texttt{meta-llama/Llama-3.2-1B}}     & 1.2B  & 16 \\
  & \href{https://huggingface.co/meta-llama/Llama-3.2-3B}{\texttt{meta-llama/Llama-3.2-3B}}     & 3.2B  & 28 \\
  & \href{https://huggingface.co/meta-llama/Llama-3.1-8B}{\texttt{meta-llama/Llama-3.1-8B}}     & 8.0B  & 32 \\
\bottomrule
\end{tabular}
\end{table}

\textbf{Finetuning implementation details.} All models are finetuned using the TRL SFTTrainer with custom completion-only supervision: the cross-entropy loss is computed on the answer token $T$ only. Training data is generated using the CLUTRR generator, producing 1000 examples consisting of an equal mixture of 2--5-hop or 2-10hop stories for each of the two settings. To prevent memorisation of individual names, all individuals are renamed to Person1 through PersonK, with the numerical assignment order randomised across examples. Stories are presented in plain completion format (ending with \textit{Therefore, PersonA is PersonB's''}) without a chat template. For LoRA finetuning, we use rank $r = 16$, $\alpha = 32$, and dropout $0.05$, with no bias adaptation, targeting the attention modules (using the QKV projection matrices through the merged \texttt{c\_attn} module for GPT-2 only; and the \texttt{query\_key\_value} \text{and} the \texttt{dense} (output projection) modules for Pythia\footnote{We observed that including the output projection lead to better performance for the pythia models}). Full finetuning updates all model parameters. In both regimes, we train for 20 epochs with a learning rate of $2 \times 10^{-4}$, a batch size of 2 with gradient accumulation over 4 steps (effective batch size 8), a warmup of 50 steps, and a maximum sequence length of 512 tokens.

\newpage
\section{Additional results}\label{app:results}
\setcounter{figure}{0}
\renewcommand{\thefigure}{B\arabic{figure}}

\subsection{Additional logit-lens results for pretrained models}\label{app:logitlens-pre}
\begin{figure}[h]
    \centering
    \includegraphics[width=0.99\linewidth]{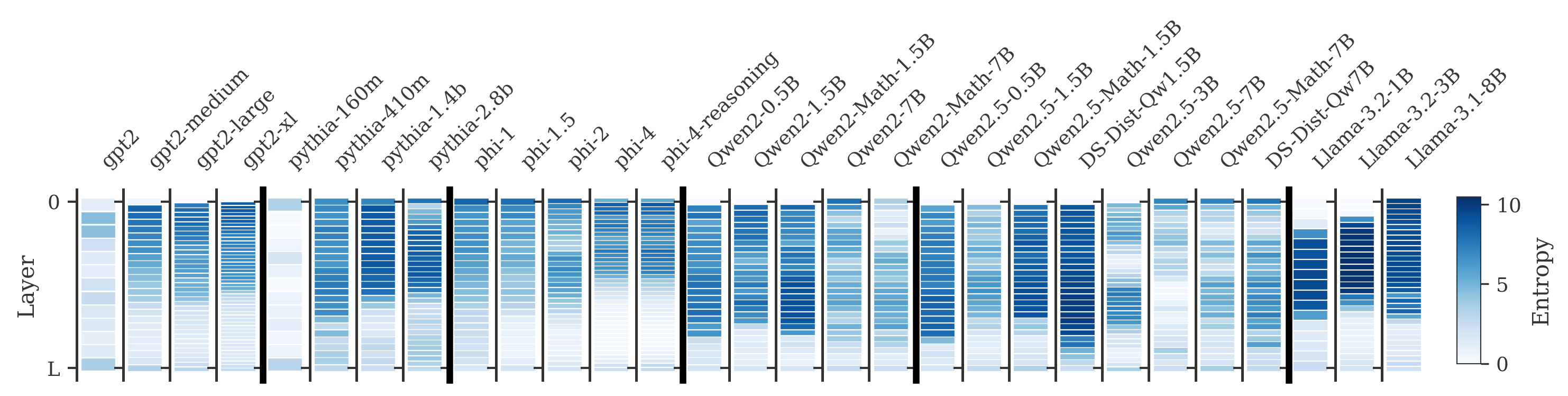}
    \caption{Entropy of the logit lens prediction $\tilde{p}_{l,T}$ by layer for pretrained models, averaged across all 900 CLUTRR 2-10hop test stories as in \cref{fig:family_rels}.}
    \label{fig:entropy}
\end{figure}

\begin{figure}[h]
  \centering
  \begin{subfigure}{0.8\textwidth}
    \centering
    \includegraphics[width=\linewidth]{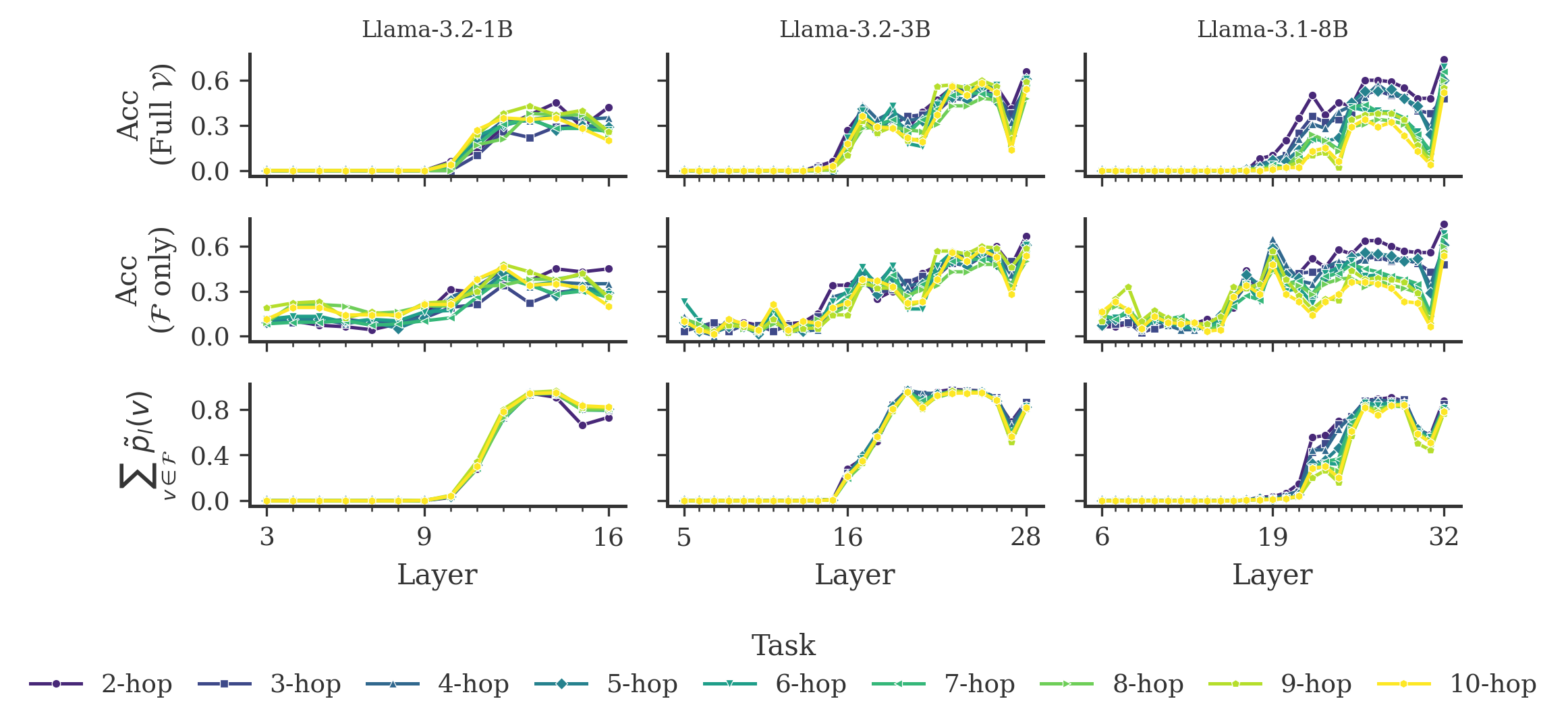}
    \caption{Llama3 Family}
    \label{fig:corr-llama}
  \end{subfigure}%
  \hfill
  \begin{subfigure}{0.8\textwidth}
    \centering
    \includegraphics[width=\linewidth]{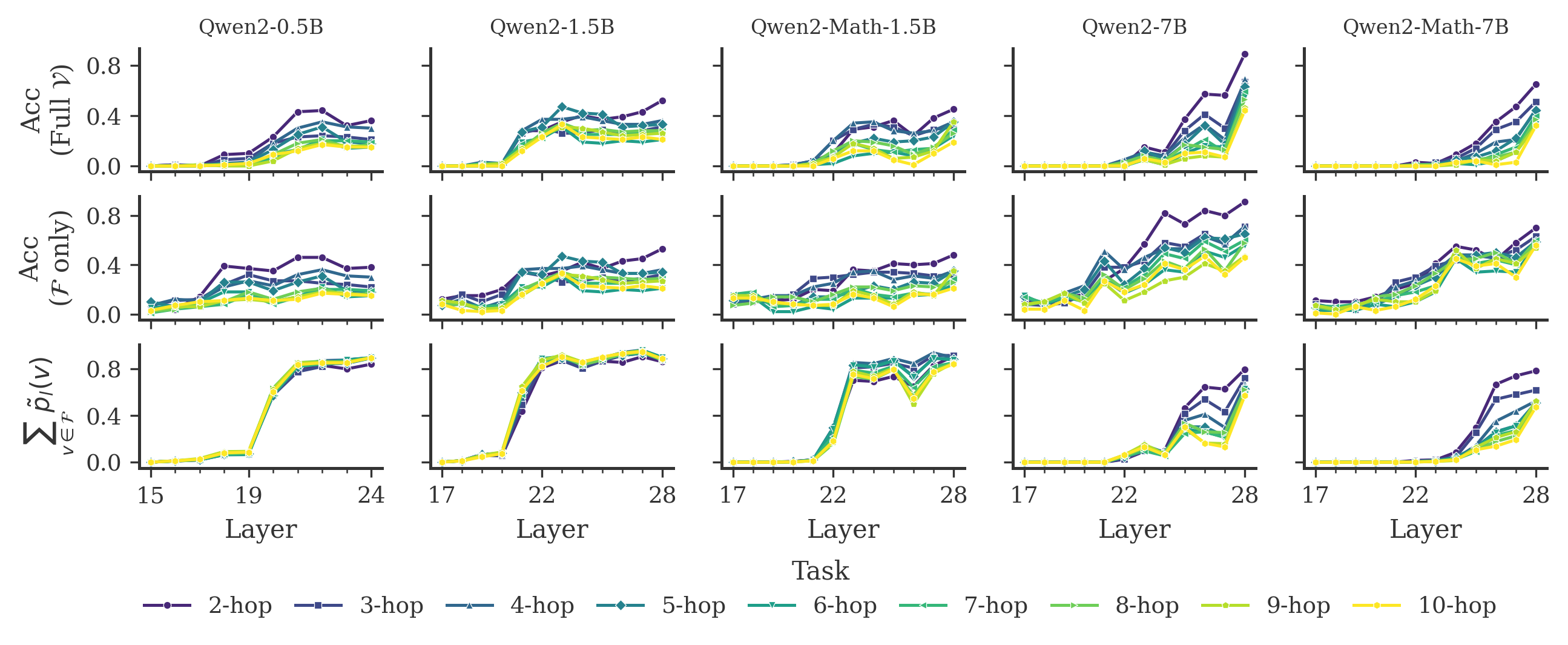}
    \caption{Qwen2 Family}
    \label{fig:corr-qwen2}
  \end{subfigure}
  \caption{Layerwise decoding results for Llama and Qwen2 families}
  \label{fig:correctness-other}
\end{figure}

\FloatBarrier
\newpage
\subsection{Additional causal patching results for pretrained models (siblings-only setting)}\label{app:cp-pre}

\begin{figure}[h]
  \centering
  \begin{subfigure}{0.32\textwidth}
    \centering
    \includegraphics[width=\linewidth]{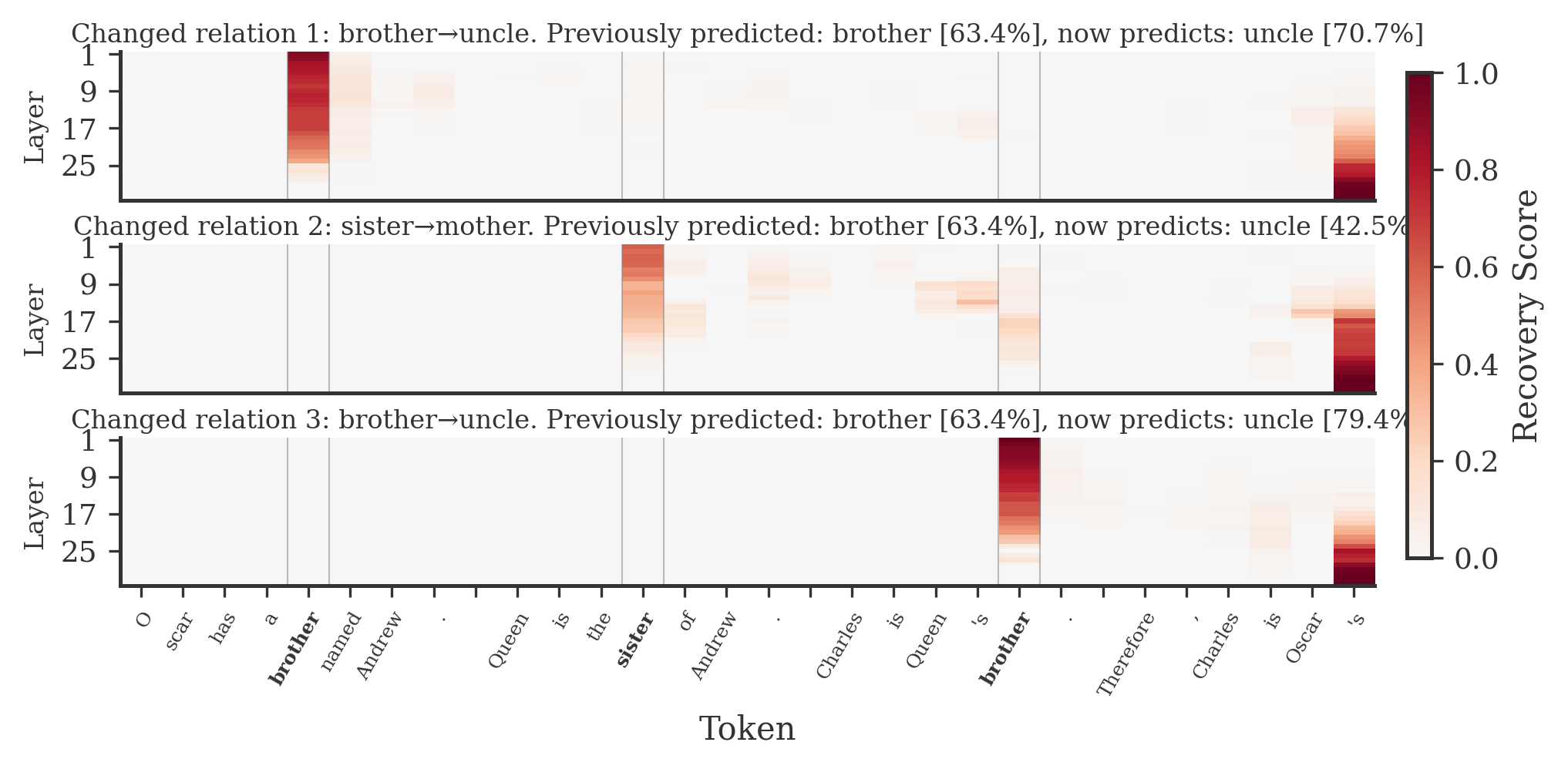}
    \caption{3-hop: Phi2}
    \label{fig:3hopphi}
  \end{subfigure}%
  \hfill
  \begin{subfigure}{0.32\textwidth}
    \centering
    \includegraphics[width=\linewidth]{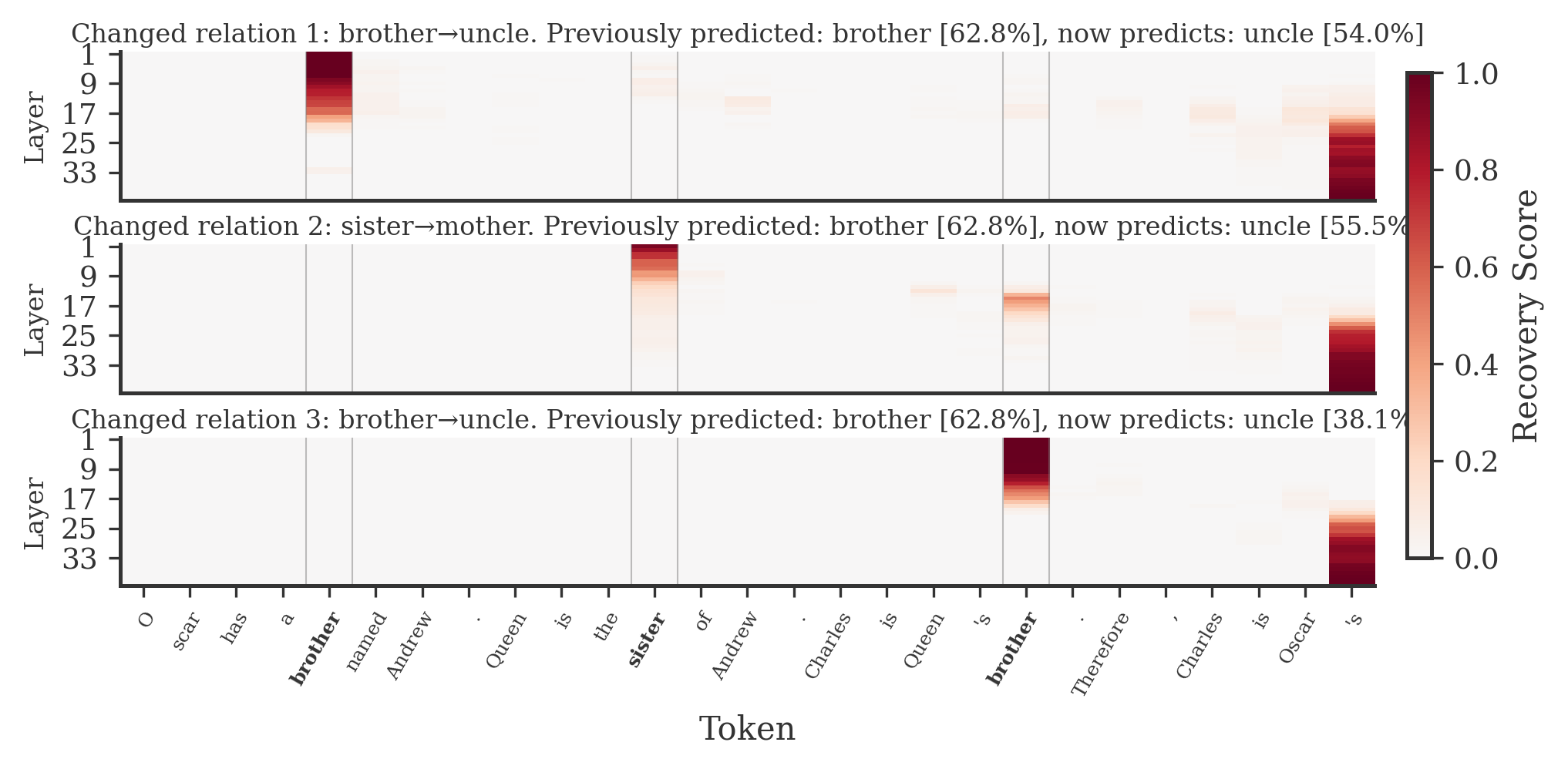}
    \caption{3-hop: Phi4}
    \label{fig:b}
  \end{subfigure}
    \hfill
  \begin{subfigure}{0.32\textwidth}
    \centering
    \includegraphics[width=\linewidth]{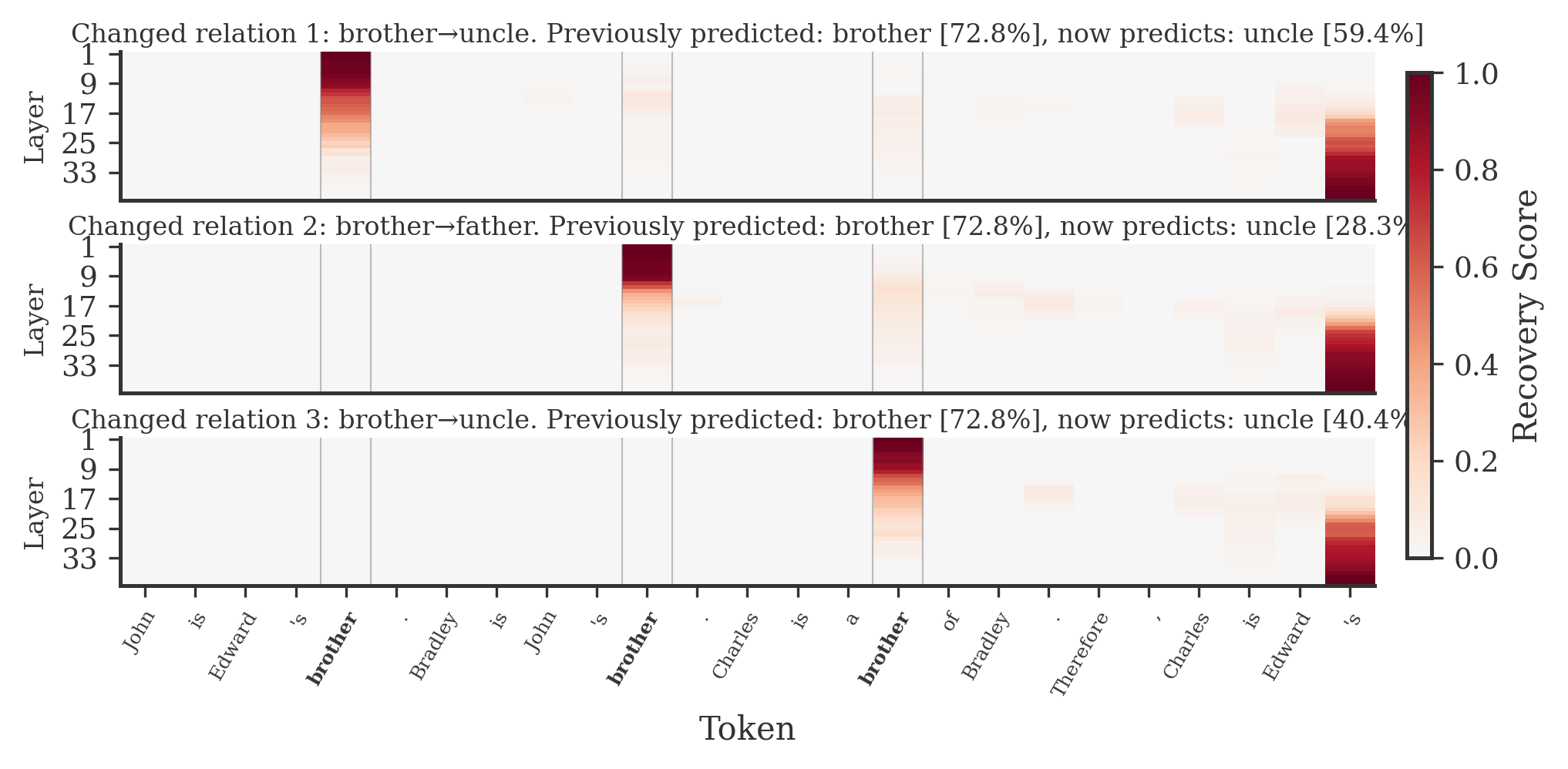}
    \caption{3-hop: Phi4-reasoning}
    \label{fig:b}
  \end{subfigure}
  \hfill
    \begin{subfigure}{0.32\textwidth}
    \centering
    \includegraphics[width=\linewidth]{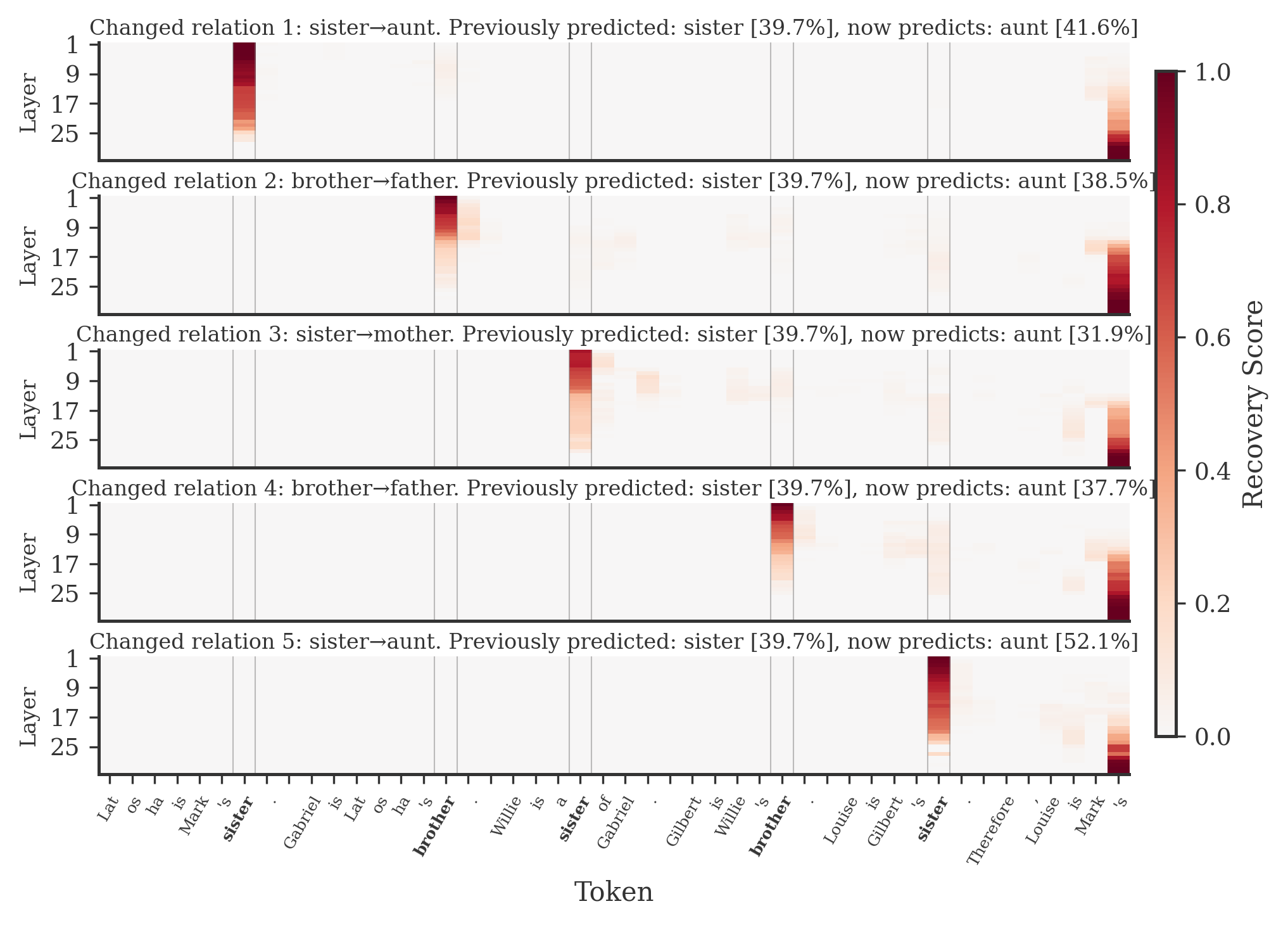}
    \caption{5-hop: Phi2}
    \label{fig:a}
  \end{subfigure}%
  \hfill
  \begin{subfigure}{0.32\textwidth}
    \centering
    \includegraphics[width=\linewidth]{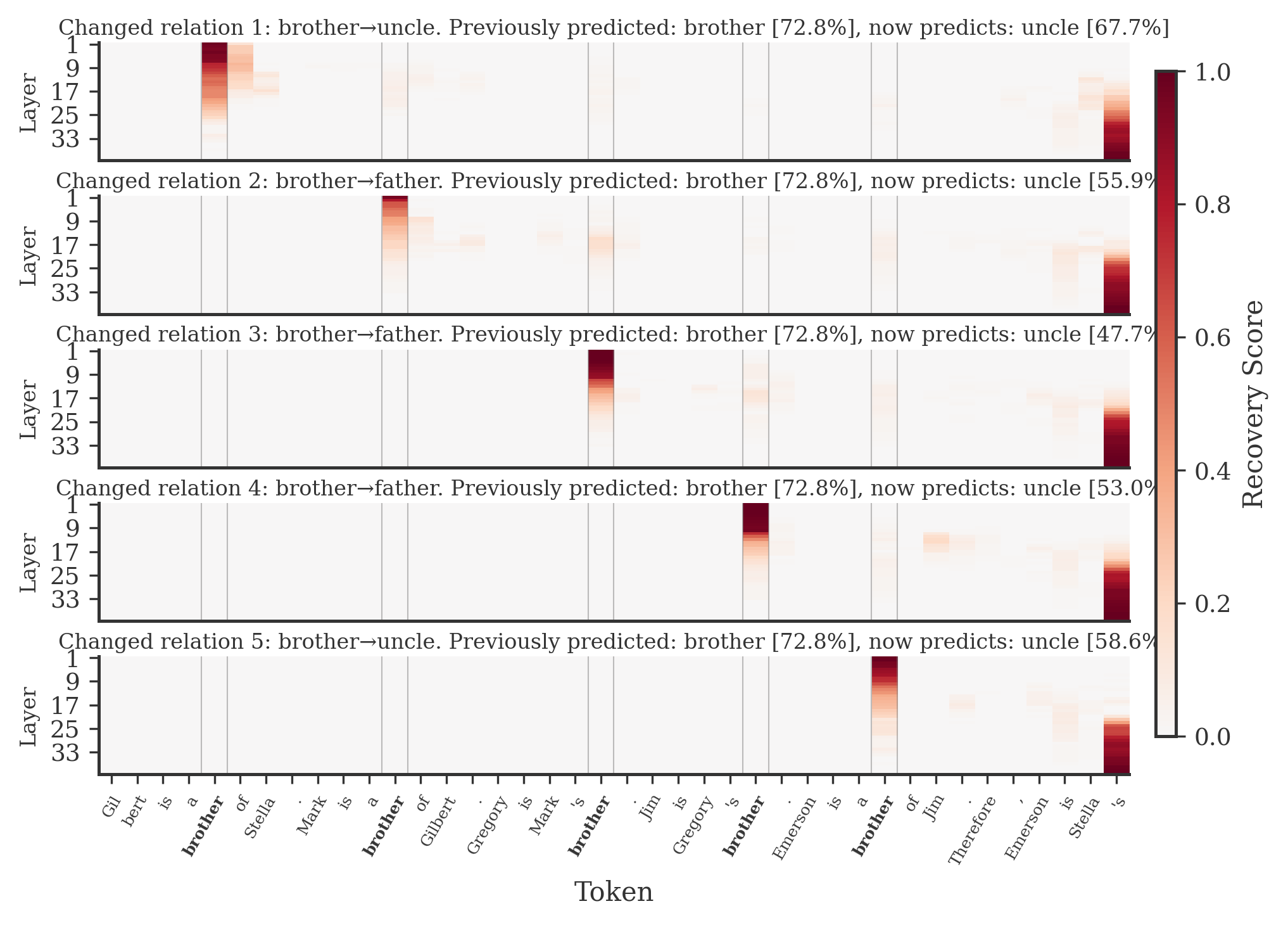}
    \caption{5-hop: Phi4}
    \label{fig:b}
  \end{subfigure}
    \hfill
  \begin{subfigure}{0.32\textwidth}
    \centering
    \includegraphics[width=\linewidth]{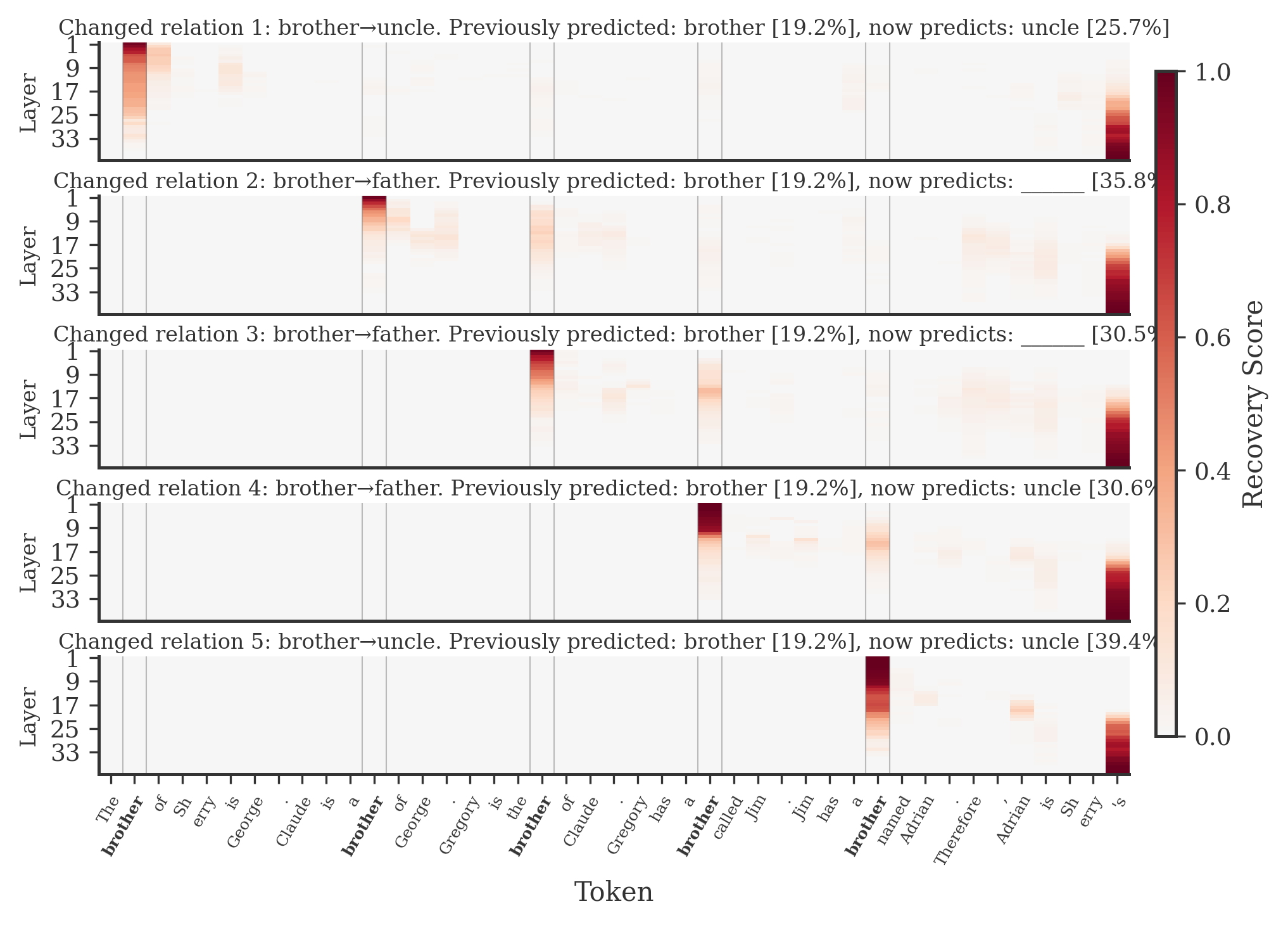}
    \caption{5-hop: Phi4-reasoning}
    \label{fig:b}
  \end{subfigure}
\hfill
    \begin{subfigure}{0.32\textwidth}
    \centering
    \includegraphics[width=\linewidth]{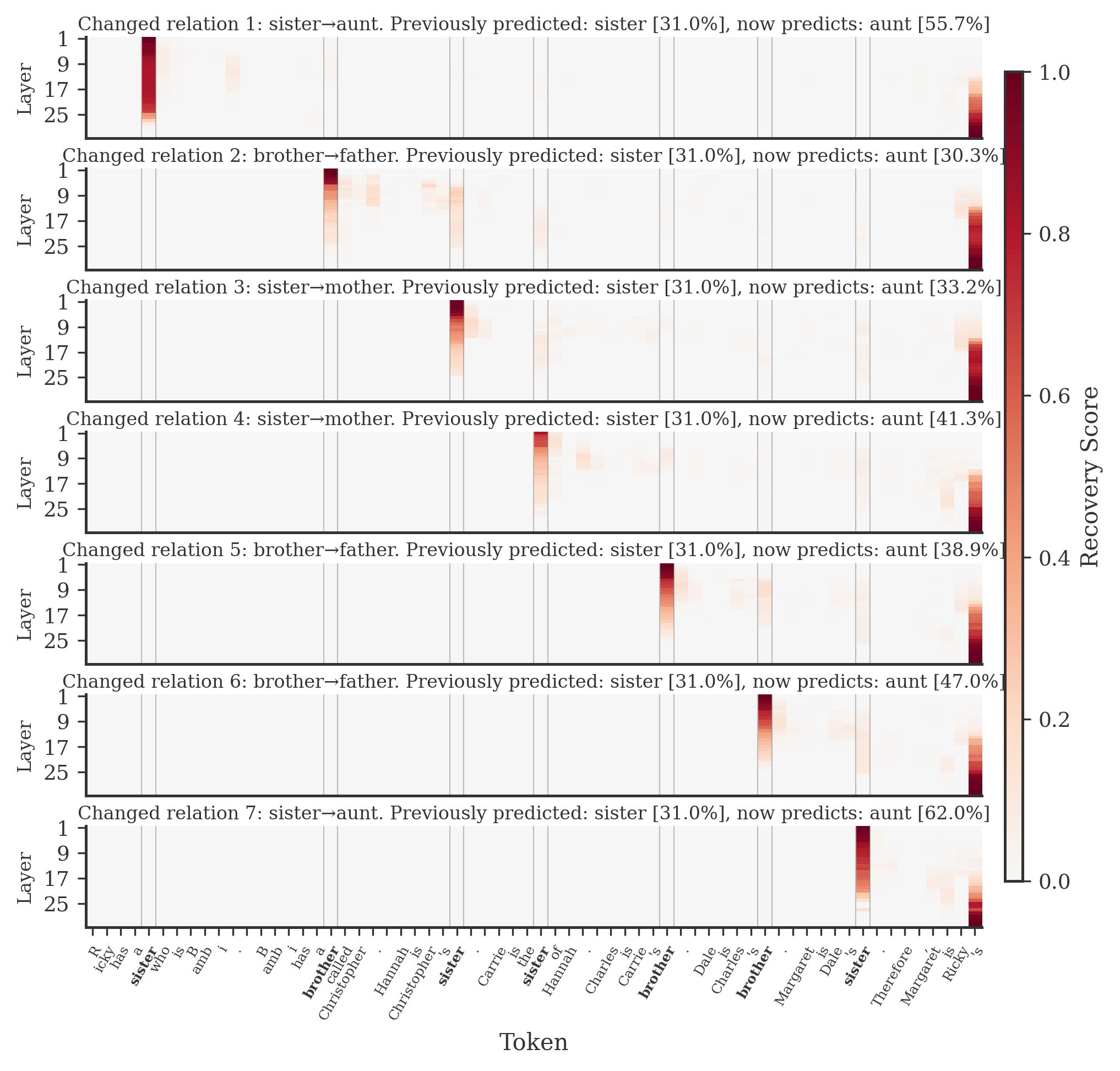}
    \caption{7-hop: Phi2}
    \label{fig:a}
  \end{subfigure}%
  \hfill
  \begin{subfigure}{0.32\textwidth}
    \centering
    \includegraphics[width=\linewidth]{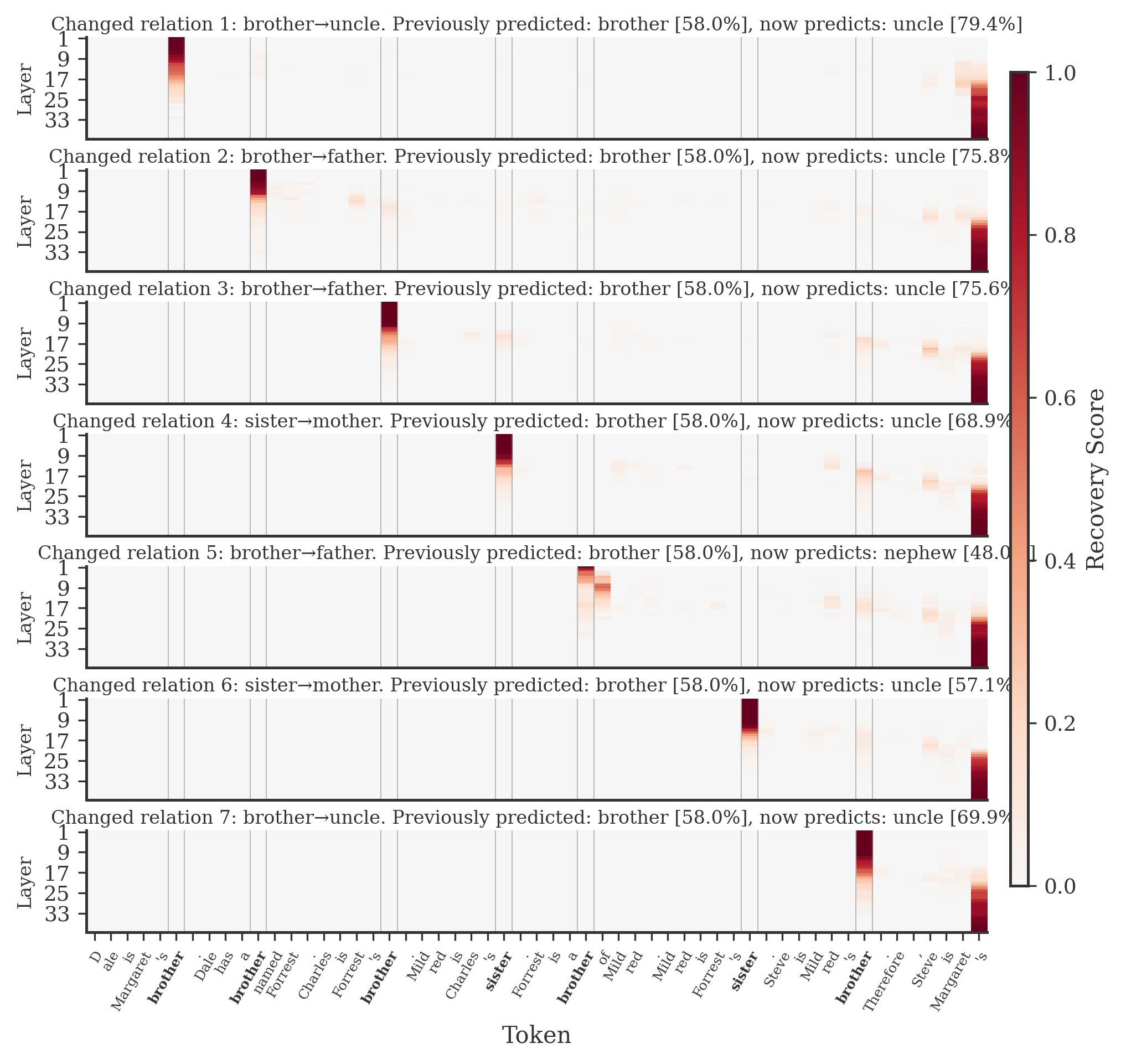}
    \caption{7-hop: Phi4}
    \label{fig:b}
  \end{subfigure}
    \hfill
  \begin{subfigure}{0.32\textwidth}
    \centering
    \includegraphics[width=\linewidth]{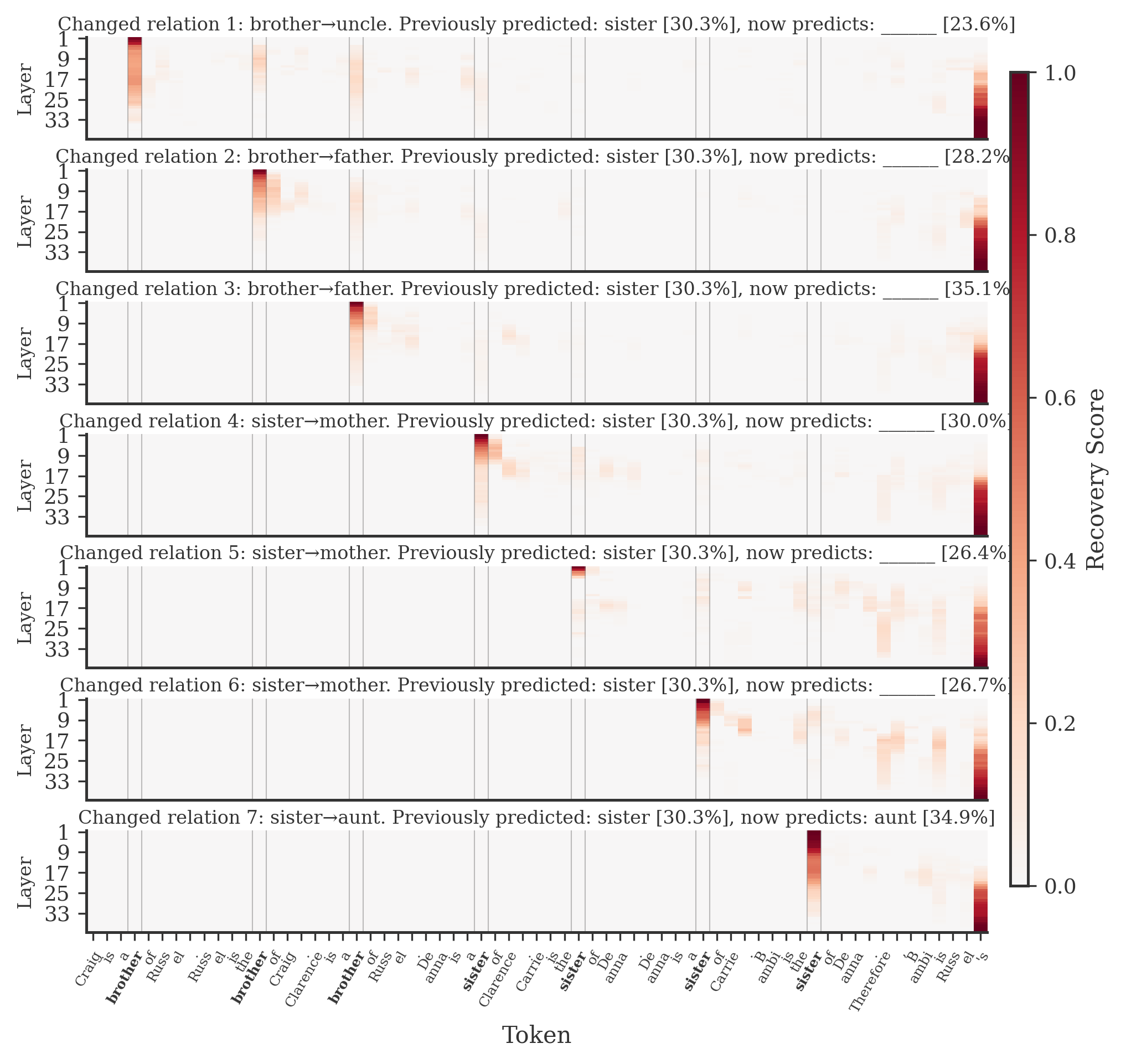}
    \caption{7-hop: Phi4-reasoning}
    \label{fig:b}
  \end{subfigure}
  \caption{Examples of causal patching trajectories for Phi models}
  \label{fig:patching-phi}
\end{figure}

\begin{figure}[h]
  \centering
  \begin{subfigure}{0.32\textwidth}
    \centering
    \includegraphics[width=\linewidth]{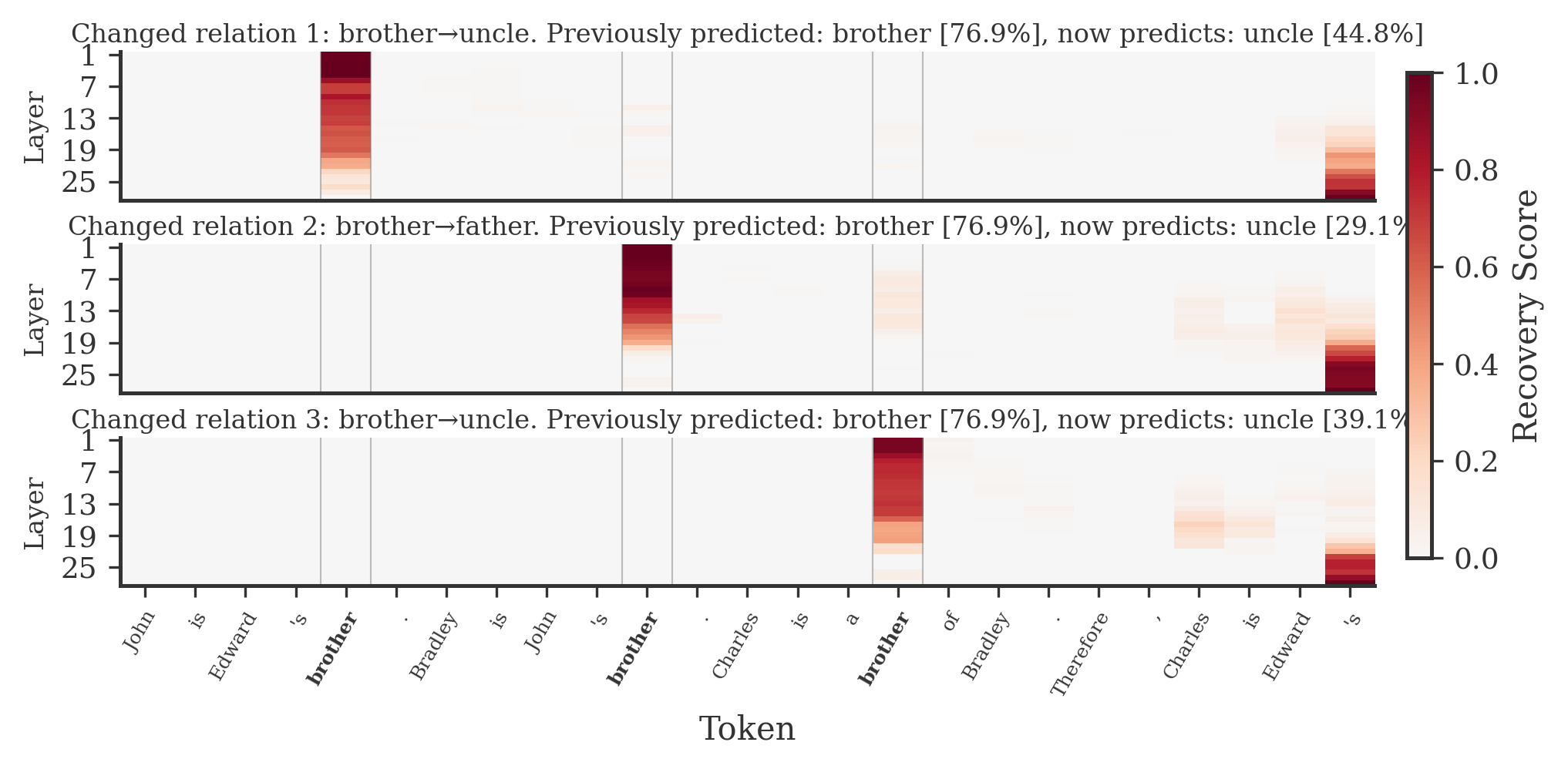}
    \caption{3-hop: Qwen2.5-1.5B}
    \label{fig:a}
  \end{subfigure}%
  \hfill
  \begin{subfigure}{0.32\textwidth}
    \centering
    \includegraphics[width=\linewidth]{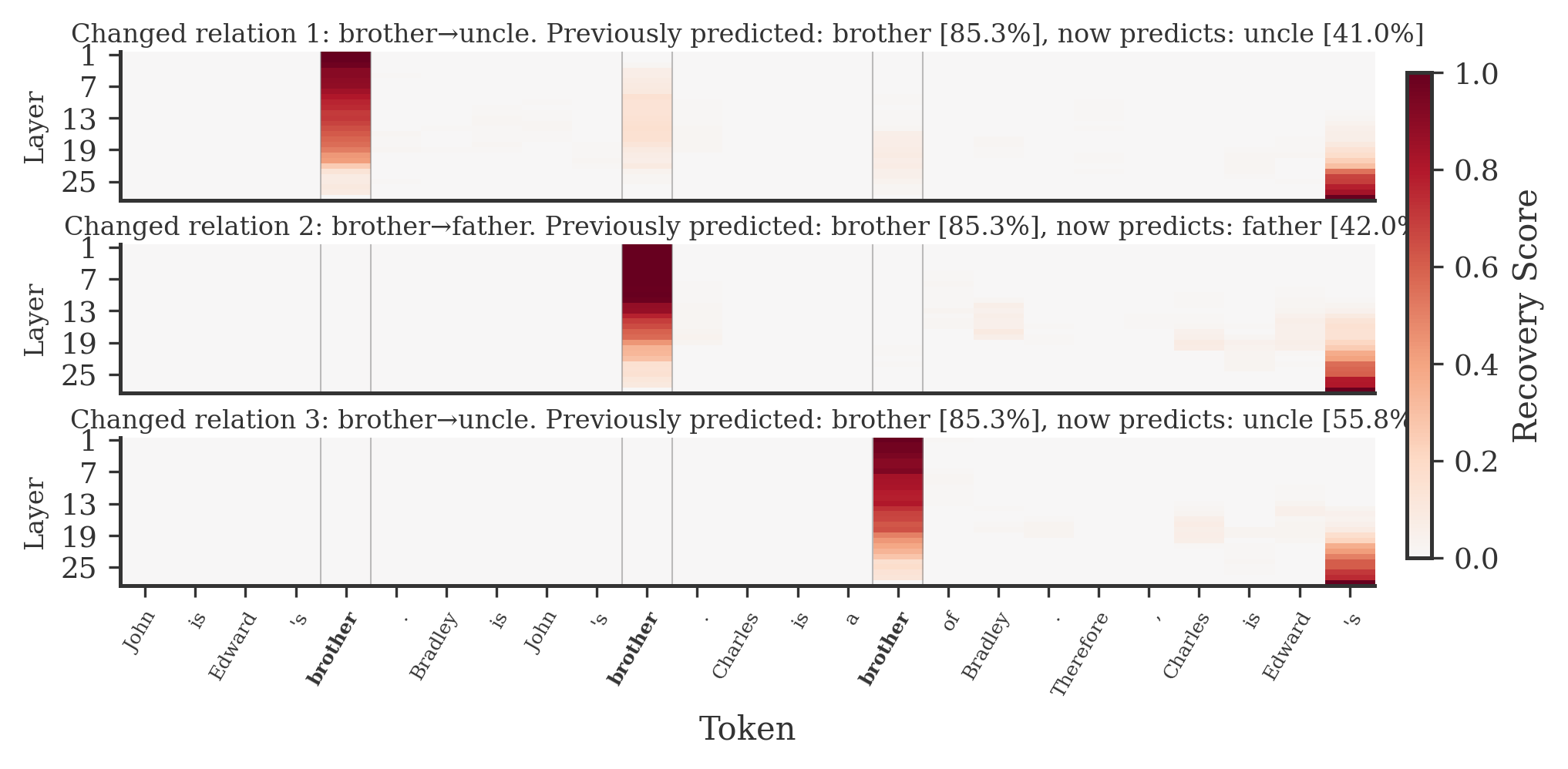}
    \caption{3-hop: Qwen2.5-7B}
    \label{fig:b}
  \end{subfigure}
    \hfill
  \begin{subfigure}{0.32\textwidth}
    \centering
    \includegraphics[width=\linewidth]{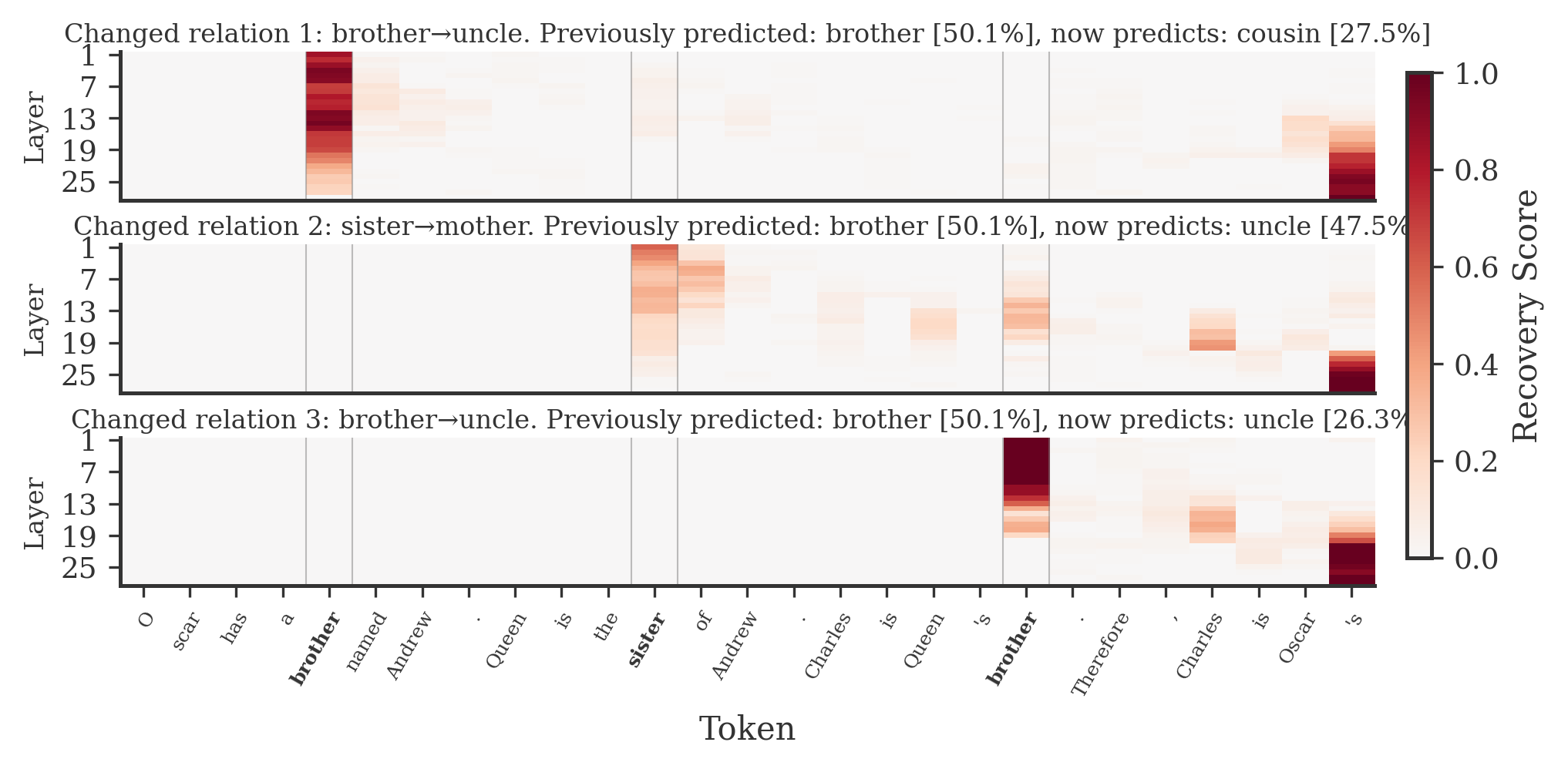}
    \caption{3-hop: Qwen2.5-Math-7B}
    \label{fig:b}
  \end{subfigure}
  \hfill
  \begin{subfigure}{0.32\textwidth}
    \centering
    \includegraphics[width=\linewidth]{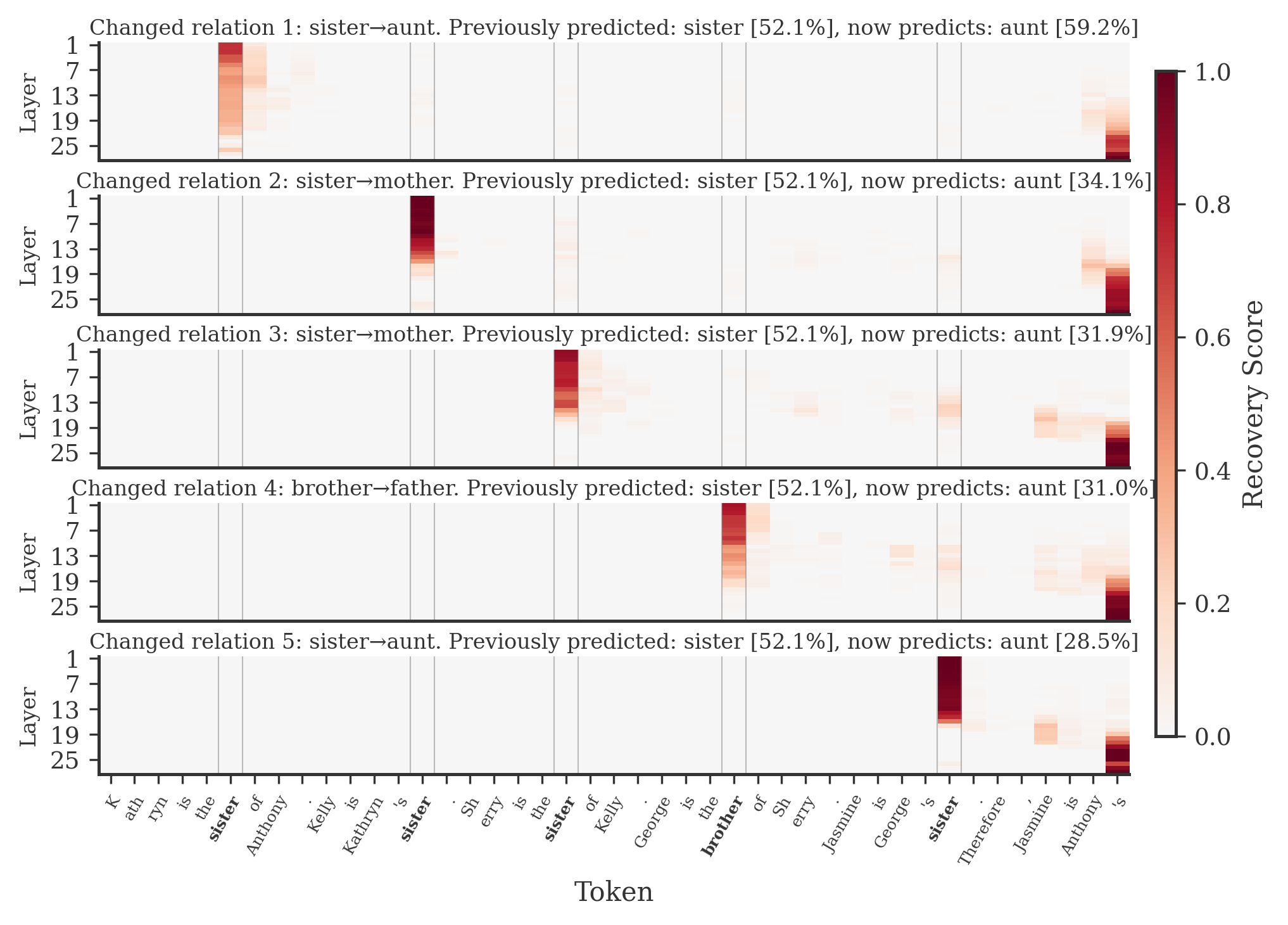}
    \caption{5-hop: Qwen2.5-1.5B}
    \label{fig:a}
  \end{subfigure}%
  \hfill
  \begin{subfigure}{0.32\textwidth}
    \centering
    \includegraphics[width=\linewidth]{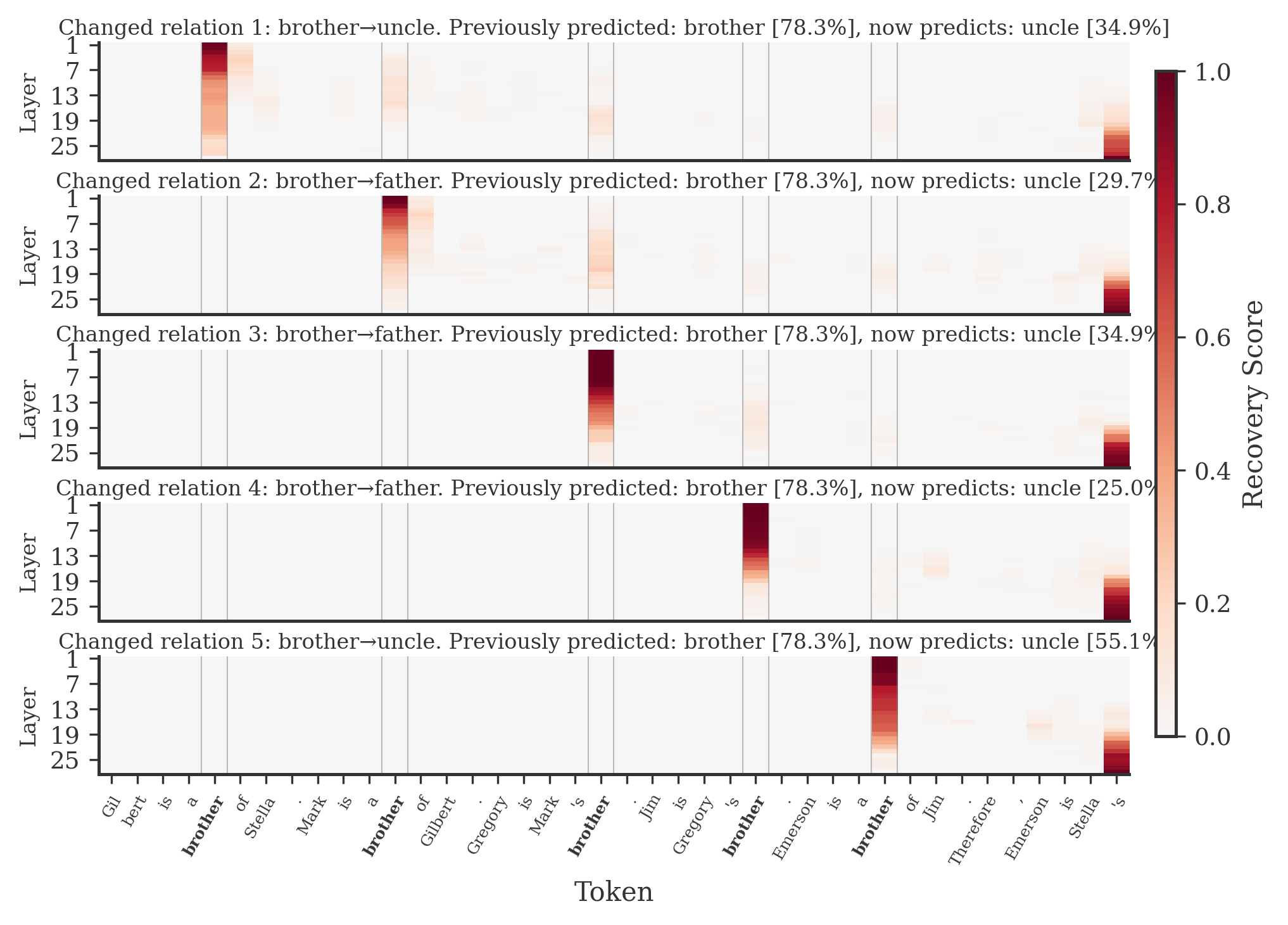}
    \caption{5-hop: Qwen2.5-7B}
    \label{fig:b}
  \end{subfigure}
    \hfill
  \begin{subfigure}{0.32\textwidth}
    \centering
    \includegraphics[width=\linewidth]{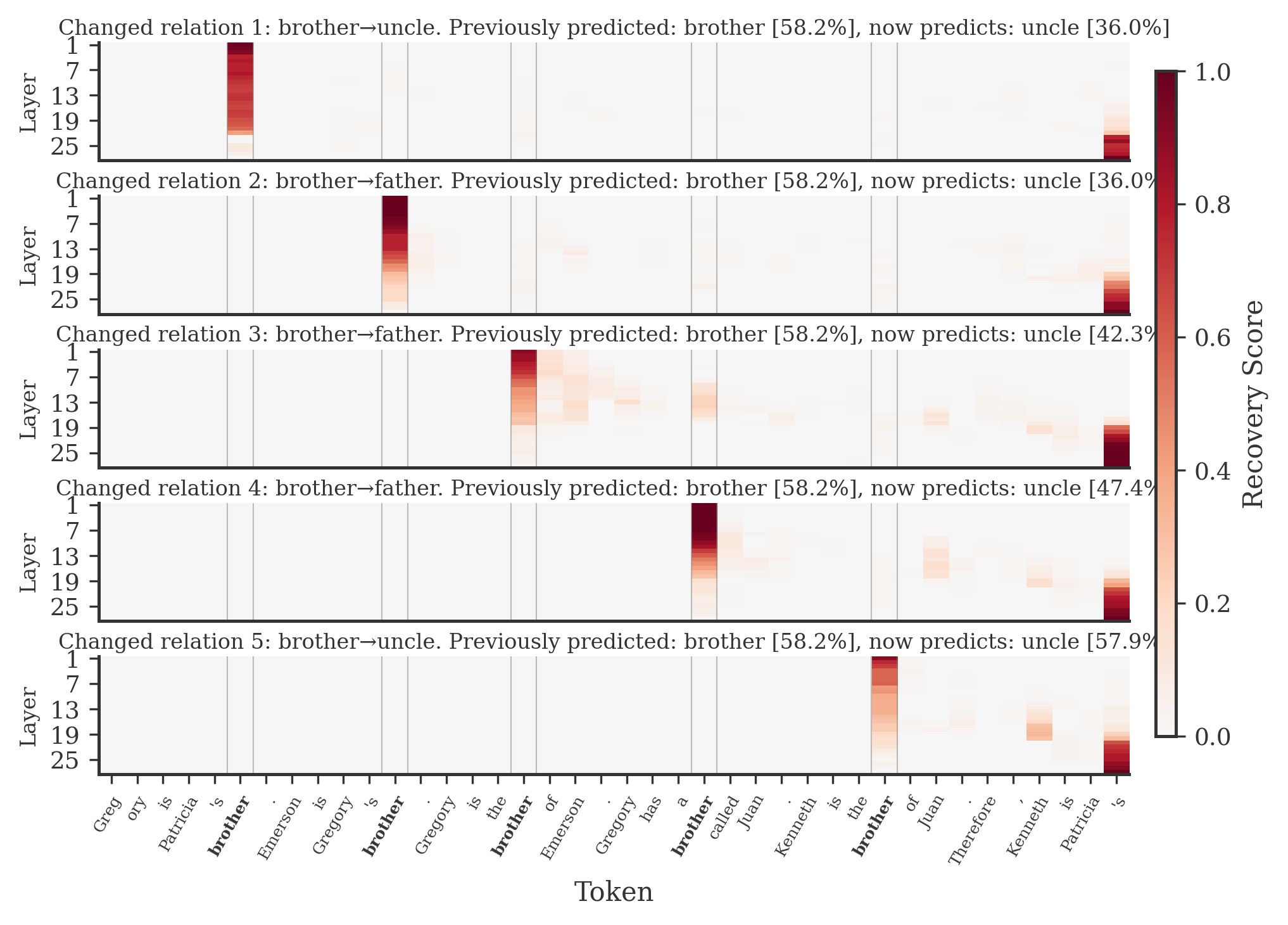}
    \caption{5-hop: Qwen2.5-Math-7B}
    \label{fig:b}
  \end{subfigure}
\hfill
  \begin{subfigure}{0.32\textwidth}
    \centering
    \includegraphics[width=\linewidth]{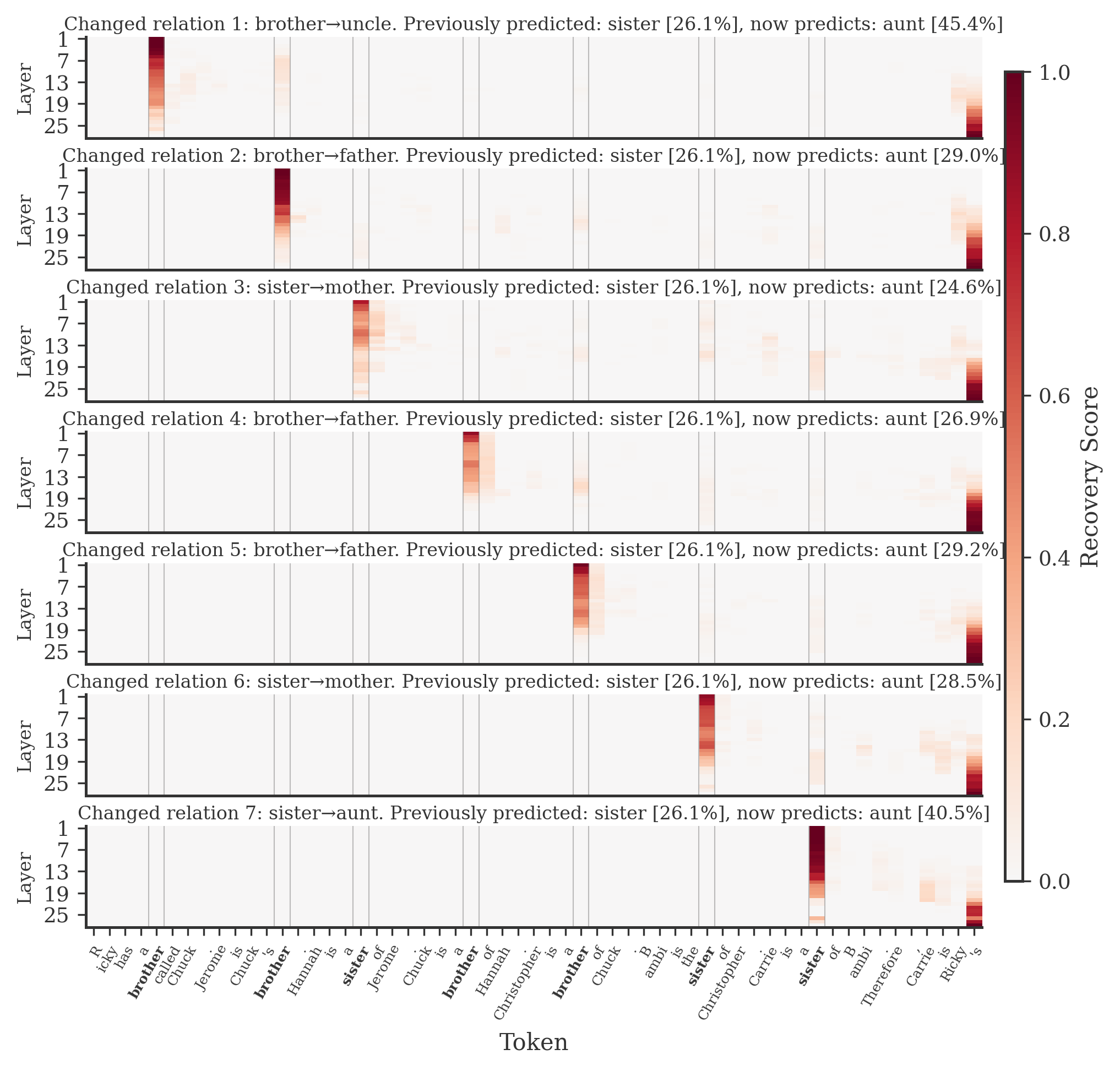}
    \caption{7-hop: Qwen2.5-1.5B}
    \label{fig:a}
  \end{subfigure}%
  \hfill
  \begin{subfigure}{0.32\textwidth}
    \centering
    \includegraphics[width=\linewidth]{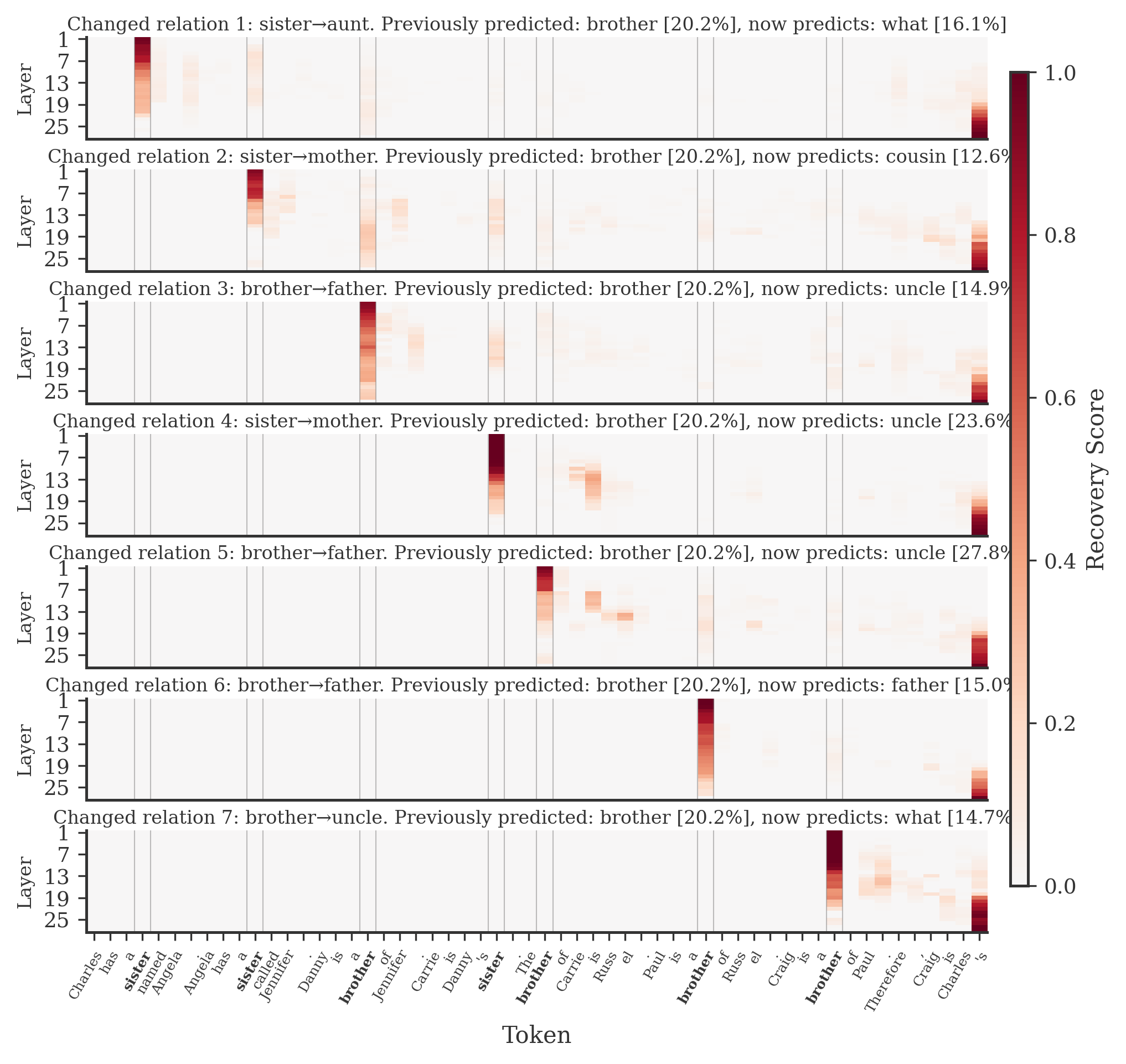}
    \caption{7-hop: Qwen2.5-7B}
    \label{fig:b}
  \end{subfigure}
    \hfill
  \begin{subfigure}{0.32\textwidth}
    \centering
    \includegraphics[width=\linewidth]{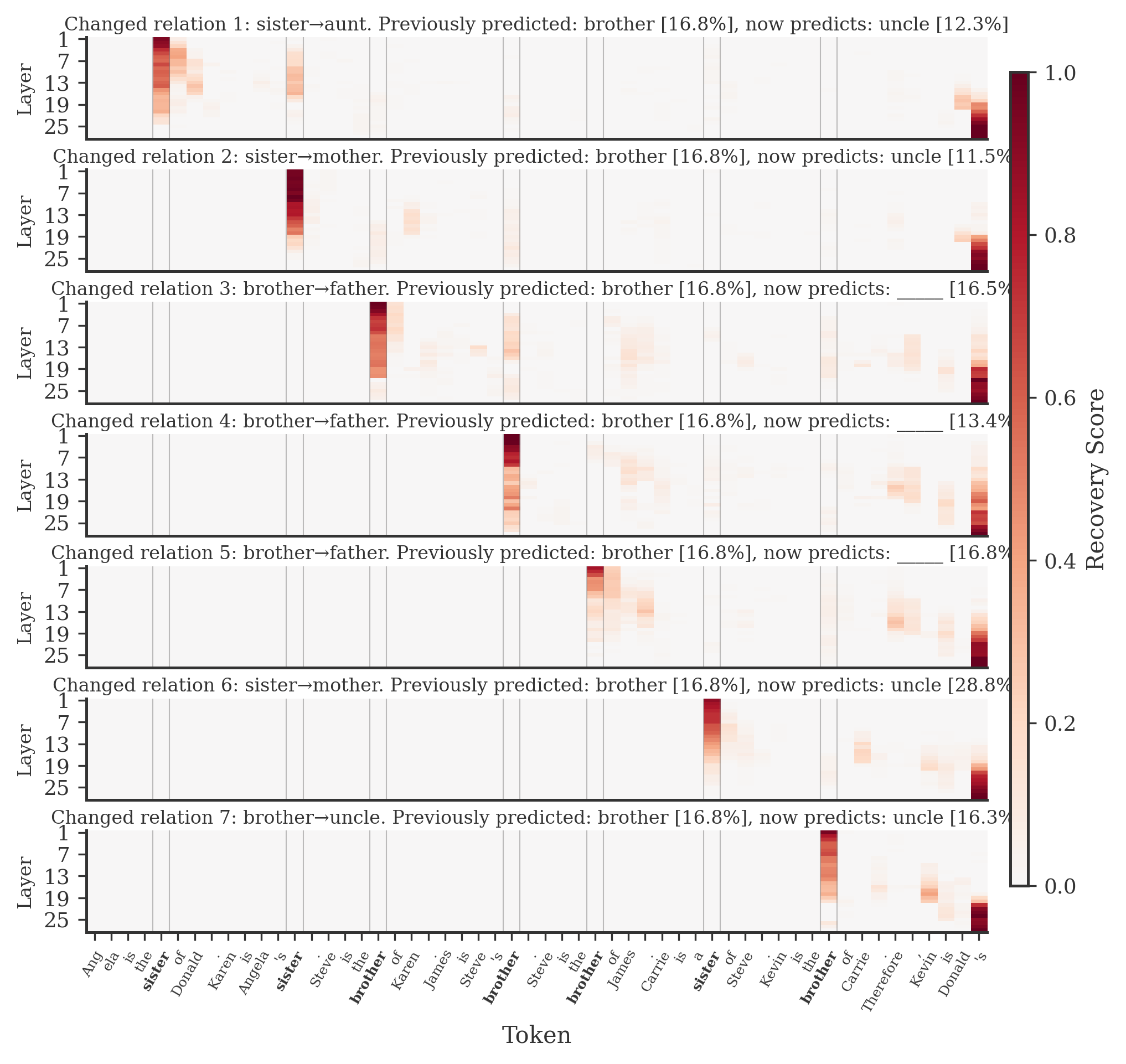}
    \caption{7-hop: Qwen2.5-Math-7B}
    \label{fig:b}
  \end{subfigure}
  \caption{Examples of causal patching trajectories for Qwen2.5 models}
  \label{fig:patching-qwen}
\end{figure}

\begin{figure}[h]
    \centering
    \includegraphics[width=0.99\linewidth]{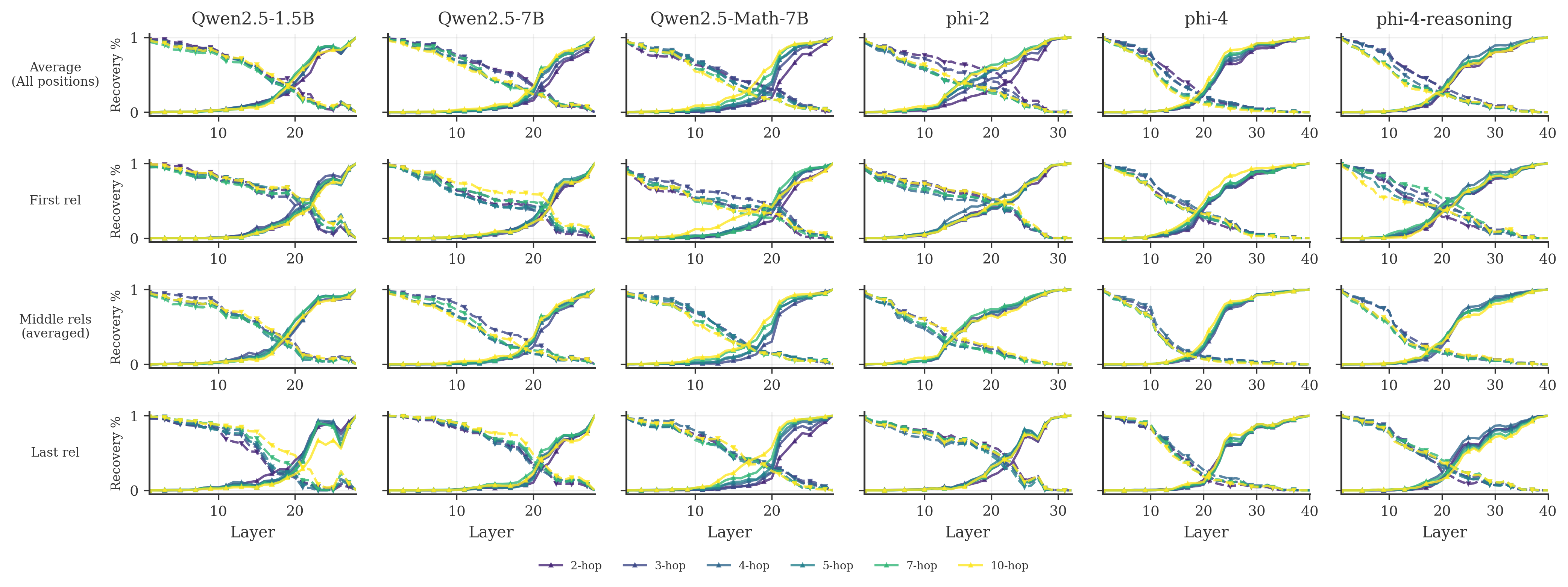}
    \caption{Average recovery score by relation replacement position, colored by number of hops, for stories which originally only had siblings. The first row is identical to \cref{fig:cp-pretrained}(b) in the main text; this figure supplements the main text figure by splitting the averaged results across replacement positions.}
    \label{fig:placeholder}
\end{figure}

\FloatBarrier
\newpage
\subsection{Additional logit lens results for finetuned models}\label{app:ll-fine}
\begin{figure}[h]
    \centering
    \includegraphics[width=0.990\linewidth]{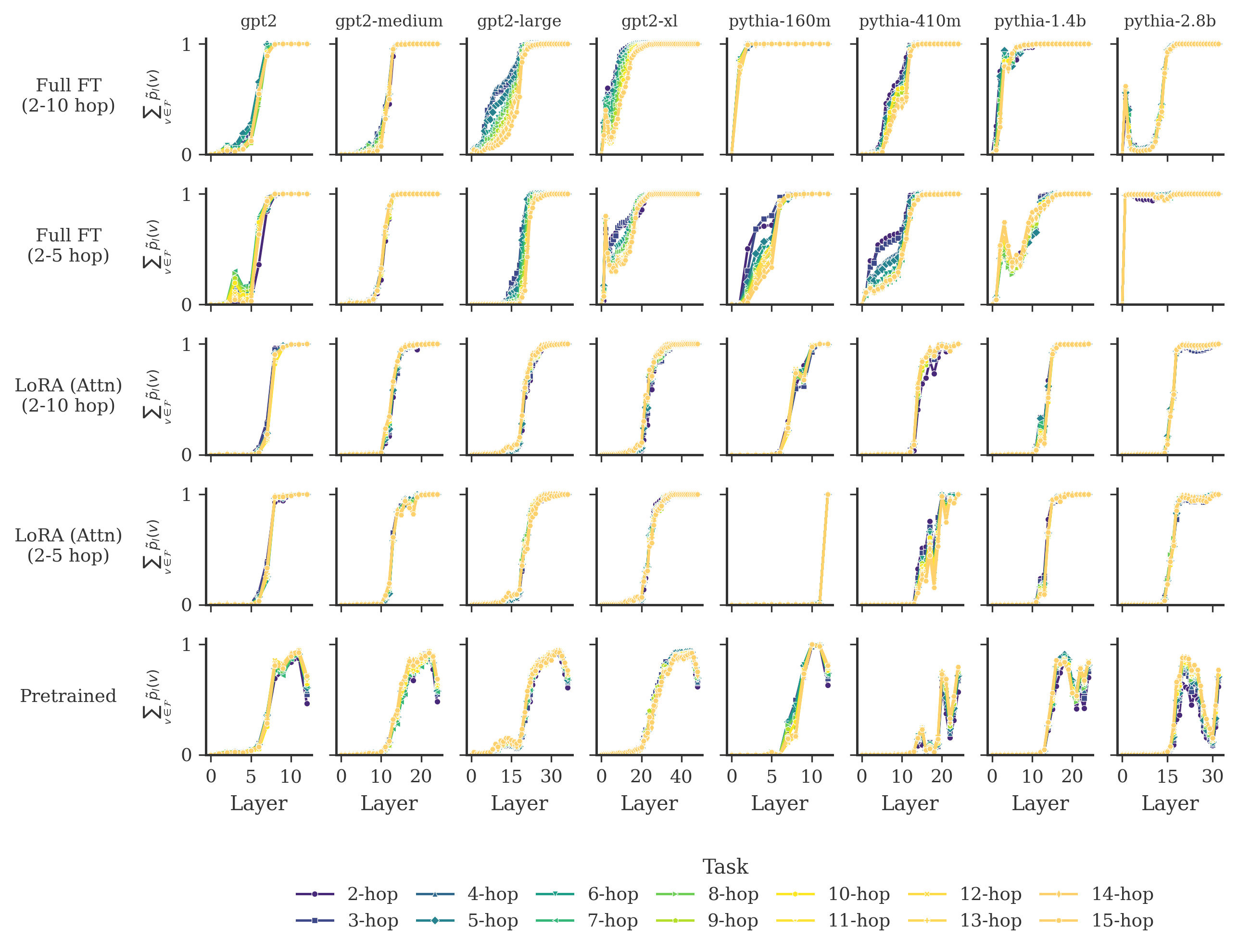}
    \caption{Probability assigned to family tokens by the logit lens prediction $\tilde{p}_{l,T}$ across layers and models by number of hops.}
    \label{fig:compare-probs}
\end{figure}

\FloatBarrier
\newpage
\subsection{Additional causal patching results for finetuned models (siblings-only setting) }\label{app:cp-fine}
\begin{figure}[ht]
  \centering
  \begin{subfigure}{0.48\textwidth}
    \centering
    \includegraphics[width=\linewidth]{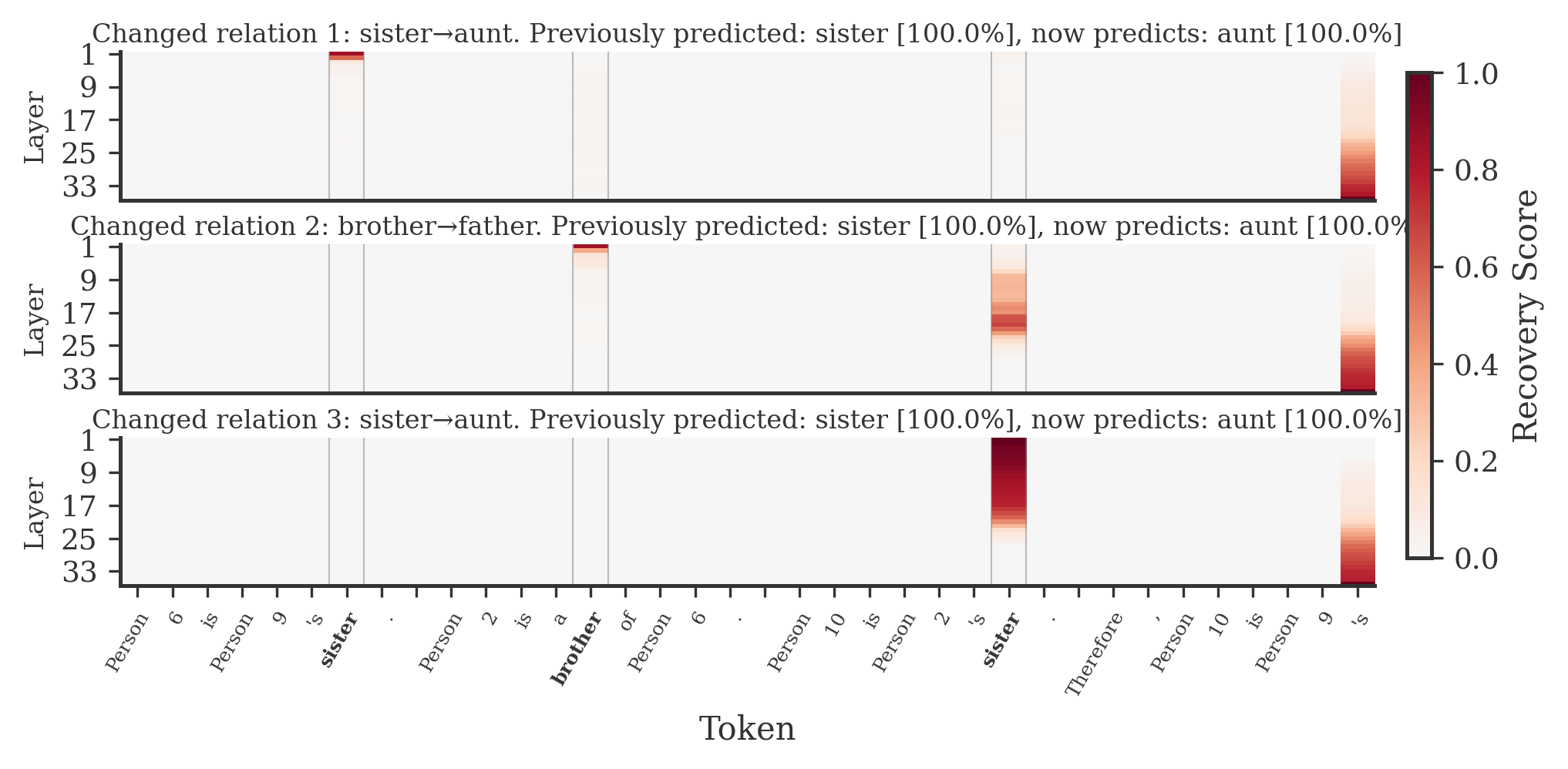}
    \caption{Patch: $uncle \rightarrow brother$, full finetuning}
    \label{fig:a}
  \end{subfigure}%
  \hfill
  \begin{subfigure}{0.48\textwidth}
    \centering
    \includegraphics[width=\linewidth]{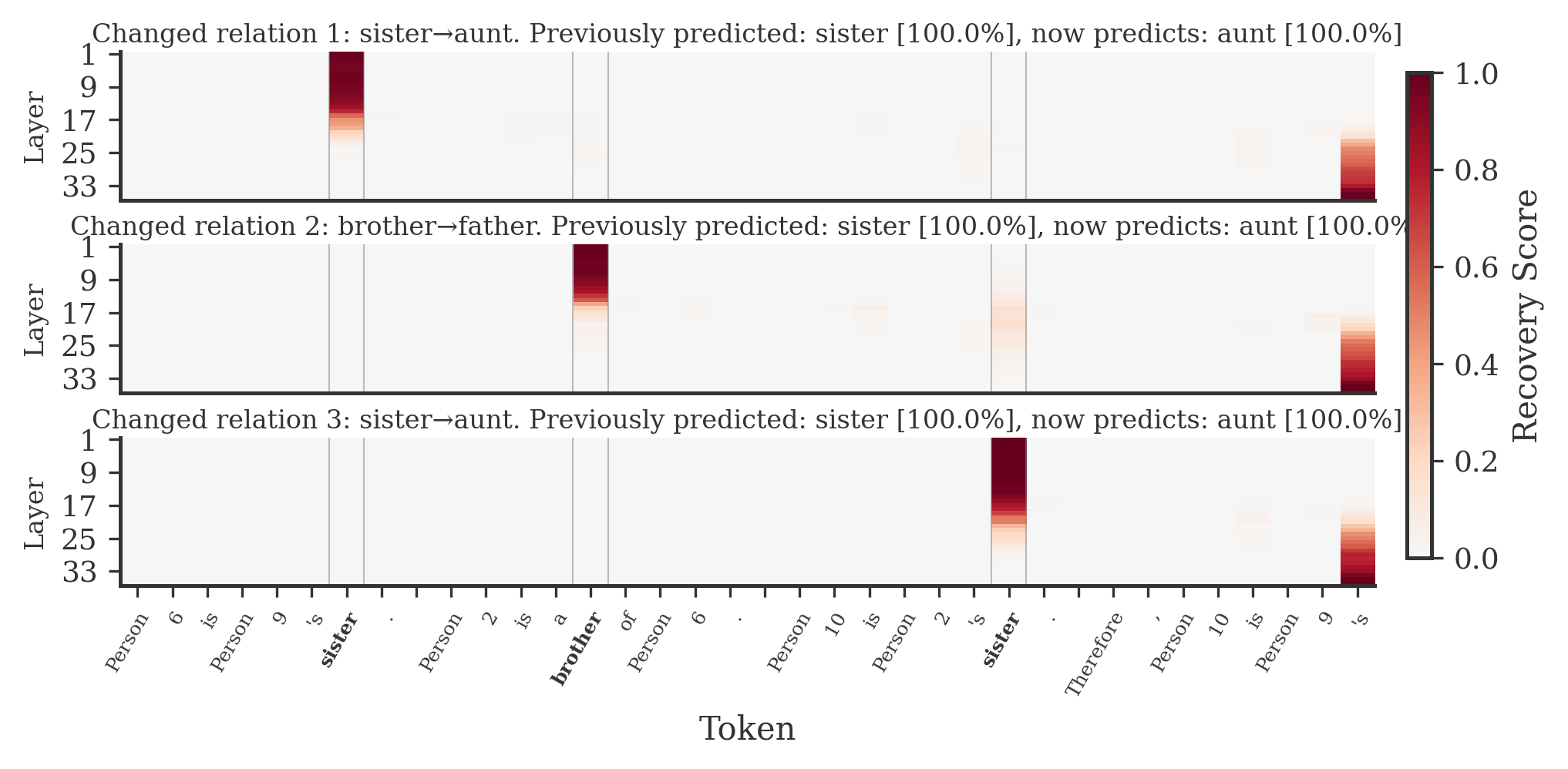}
    \caption{Patch:  $uncle \rightarrow brother$, LoRA finetuning}
    \label{fig:b}
  \end{subfigure}
    \begin{subfigure}{0.48\textwidth}
    \centering
    \includegraphics[width=\linewidth]{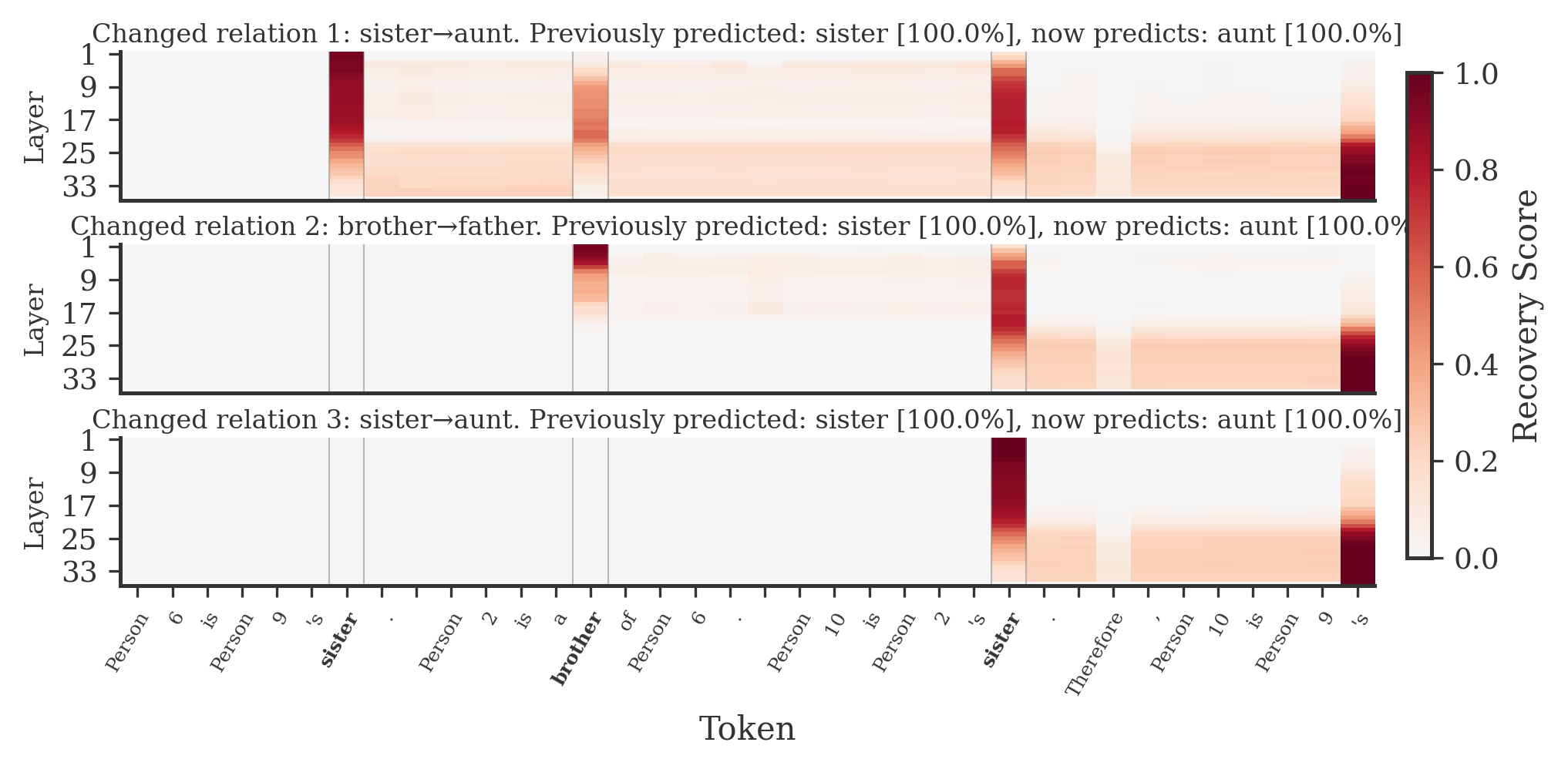}
    \caption{Patch $brother \rightarrow uncle$, full finetuning}
    \label{fig:a}
  \end{subfigure}%
  \hfill
  \begin{subfigure}{0.48\textwidth}
    \centering
    \includegraphics[width=\linewidth]{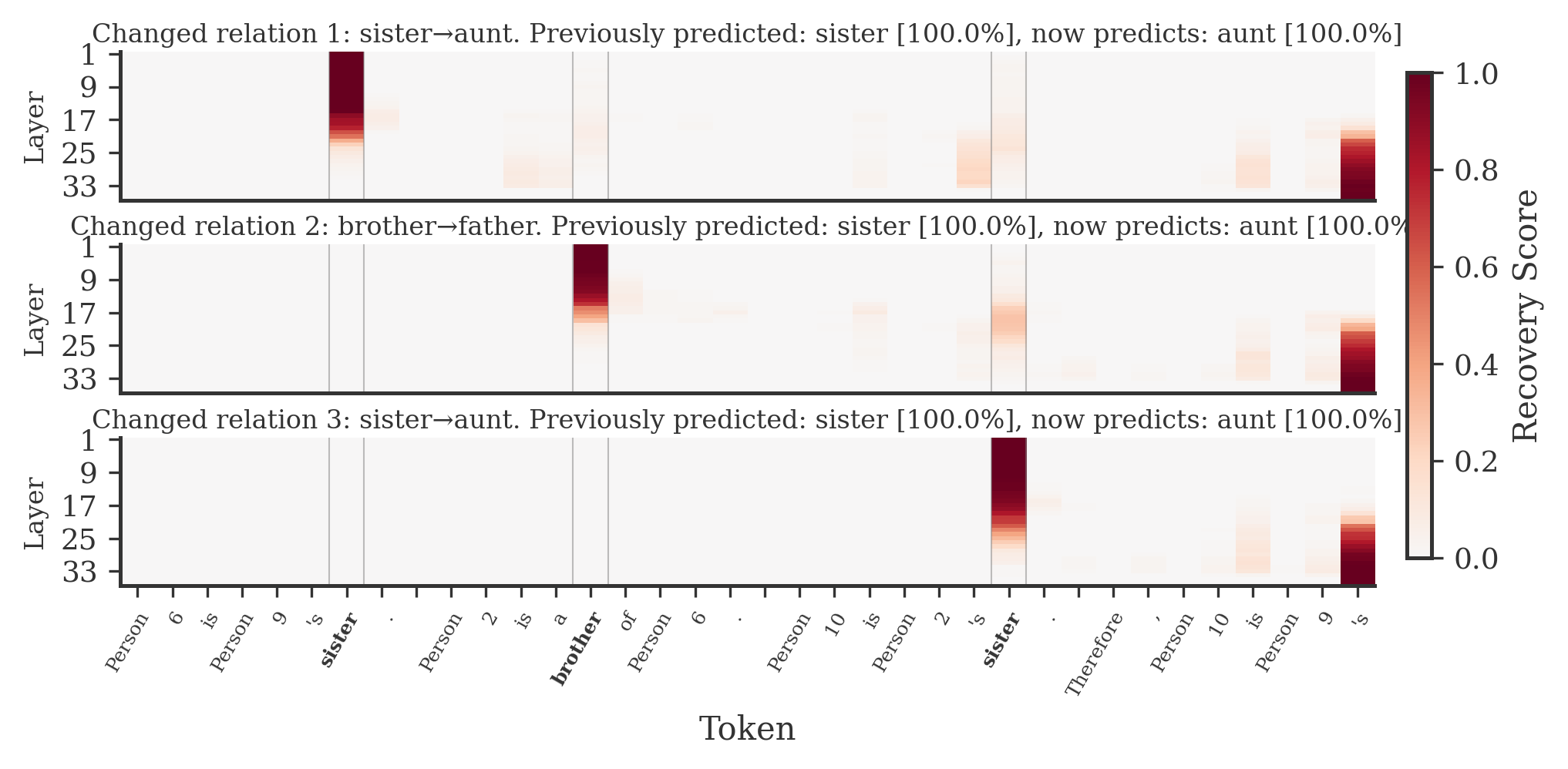}
    \caption{Patch $brother \rightarrow uncle$, LoRA finetuning}
    \label{fig:b}
  \end{subfigure}
    \begin{subfigure}{0.48\textwidth}
    \centering
    \includegraphics[width=\linewidth]{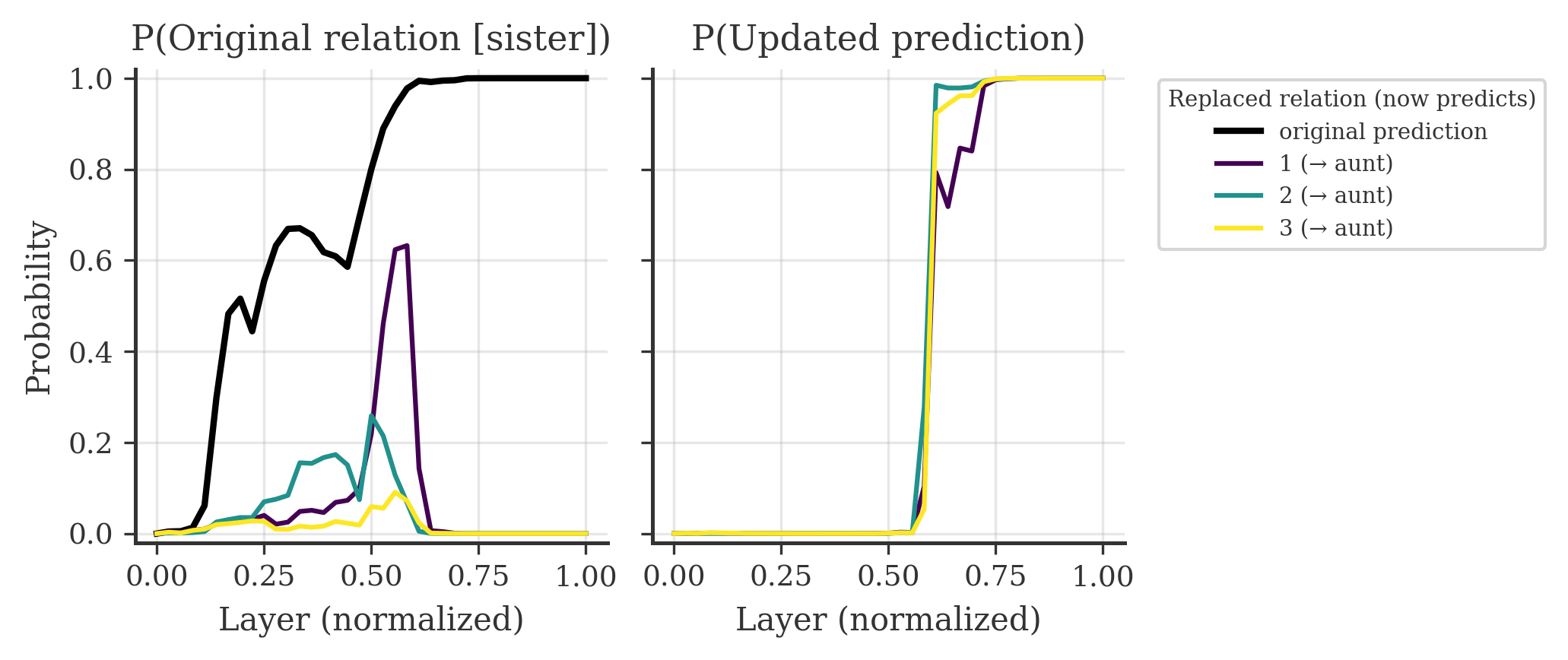}
    \caption{Logit Lens: Full finetuning}
    \label{fig:a}
  \end{subfigure}%
  \hfill
  \begin{subfigure}{0.48\textwidth}
    \centering
    \includegraphics[width=\linewidth]{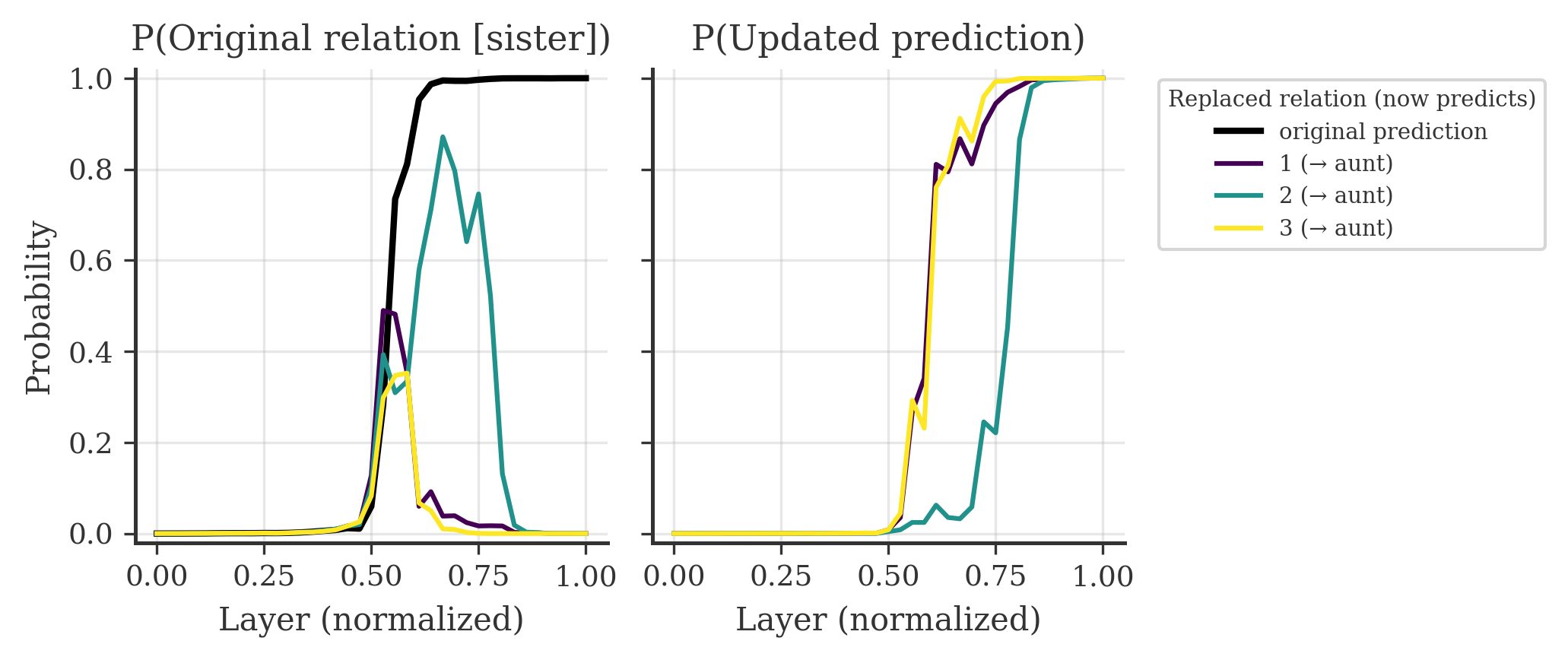}
    \caption{Logit lens: LoRA finetuning}
    \label{fig:b}
  \end{subfigure}
  \caption{Full causal patching and logit-lens results for a 3-hop example using GPT2-large.}
  \label{fig:causalpatch-hop3}
\end{figure}

\begin{figure}[ht]
  \centering
  \begin{subfigure}{0.48\textwidth}
    \centering
    \includegraphics[width=\linewidth]{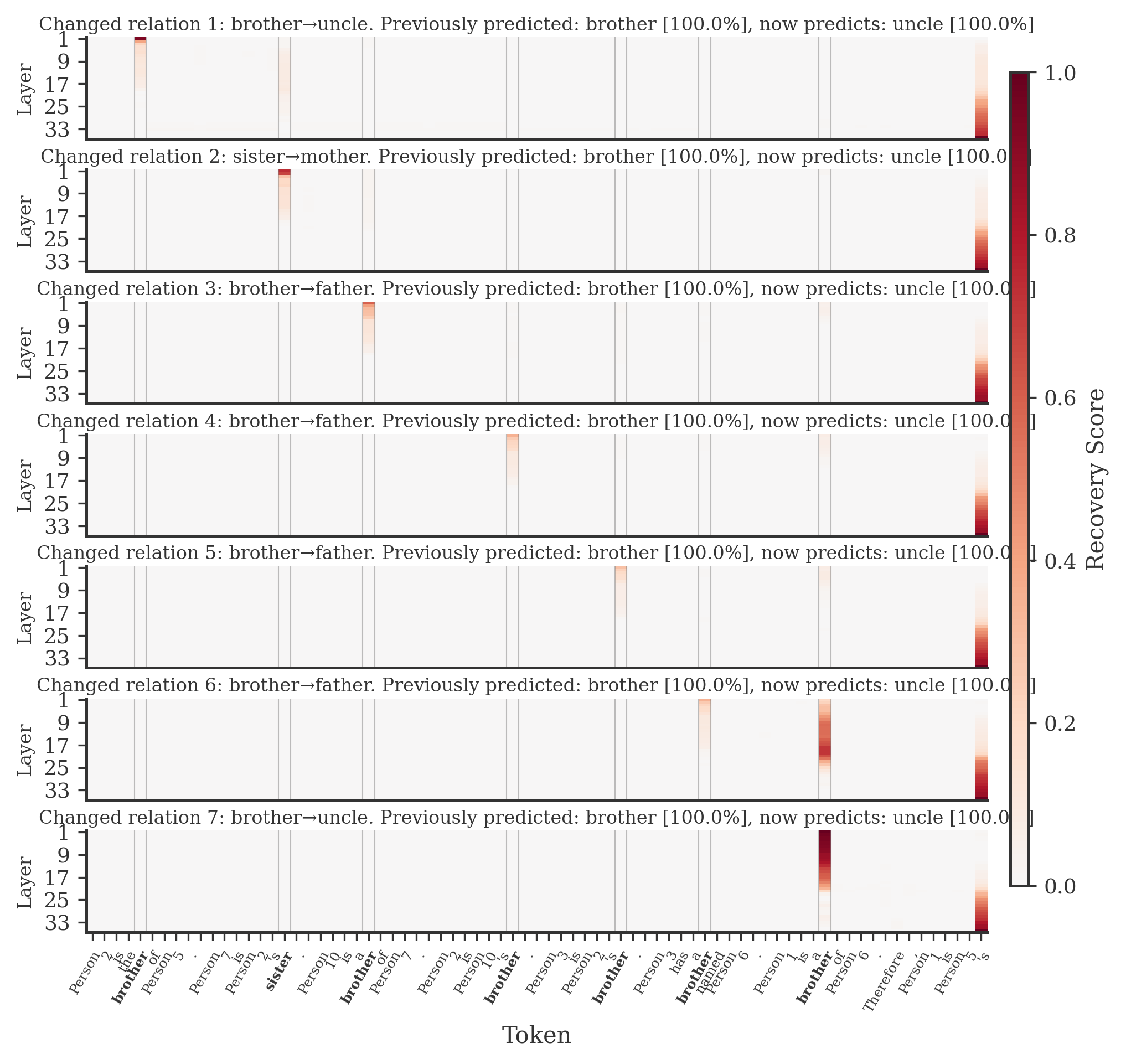}
    \caption{Patch: $uncle \rightarrow brother$, full finetuning}
    \label{fig:a}
  \end{subfigure}%
  \hfill
  \begin{subfigure}{0.48\textwidth}
    \centering
    \includegraphics[width=\linewidth]{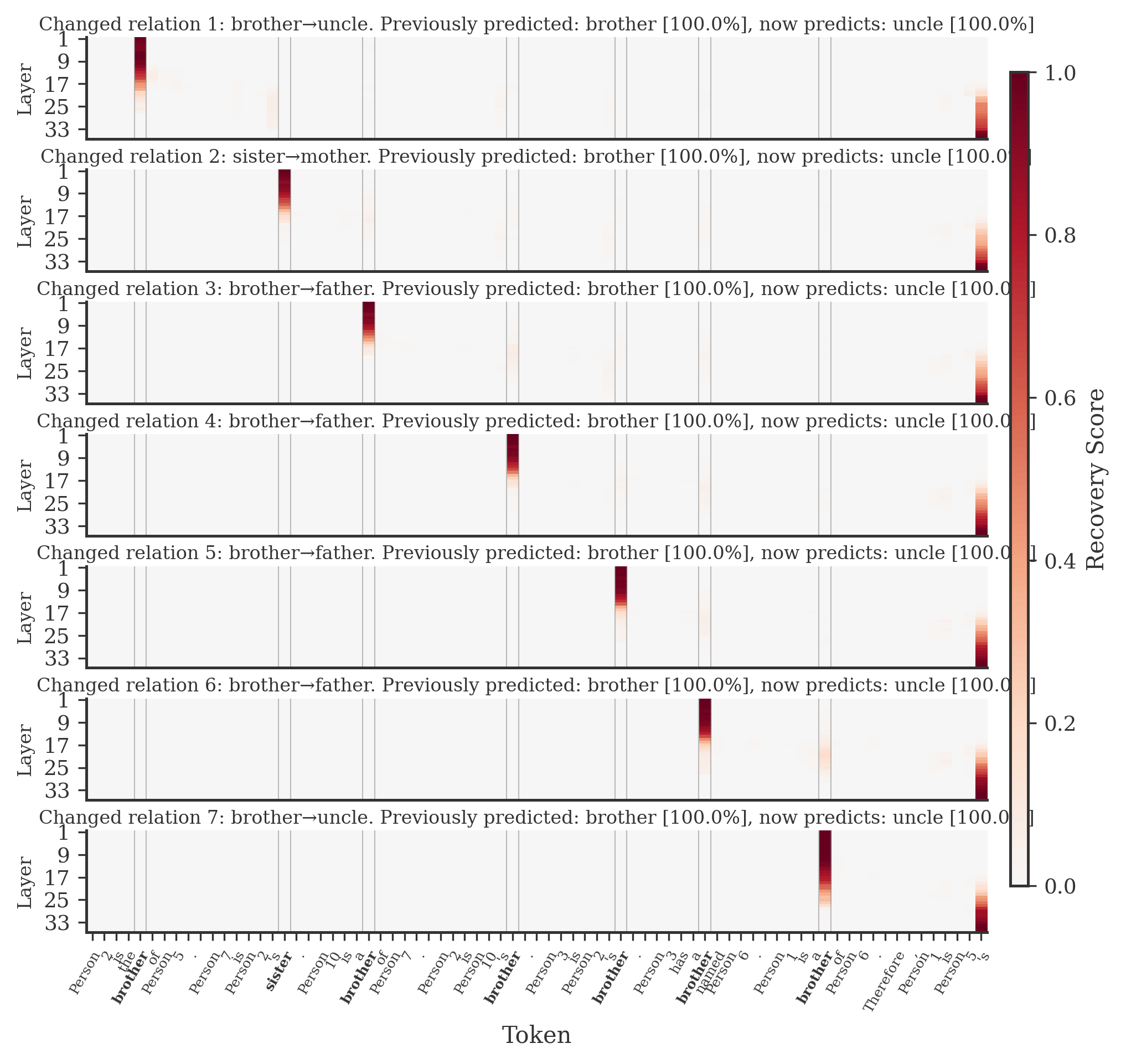}
    \caption{Patch:  $uncle \rightarrow brother$, LoRA finetuning}
    \label{fig:b}
  \end{subfigure}
    \begin{subfigure}{0.48\textwidth}
    \centering
    \includegraphics[width=\linewidth]{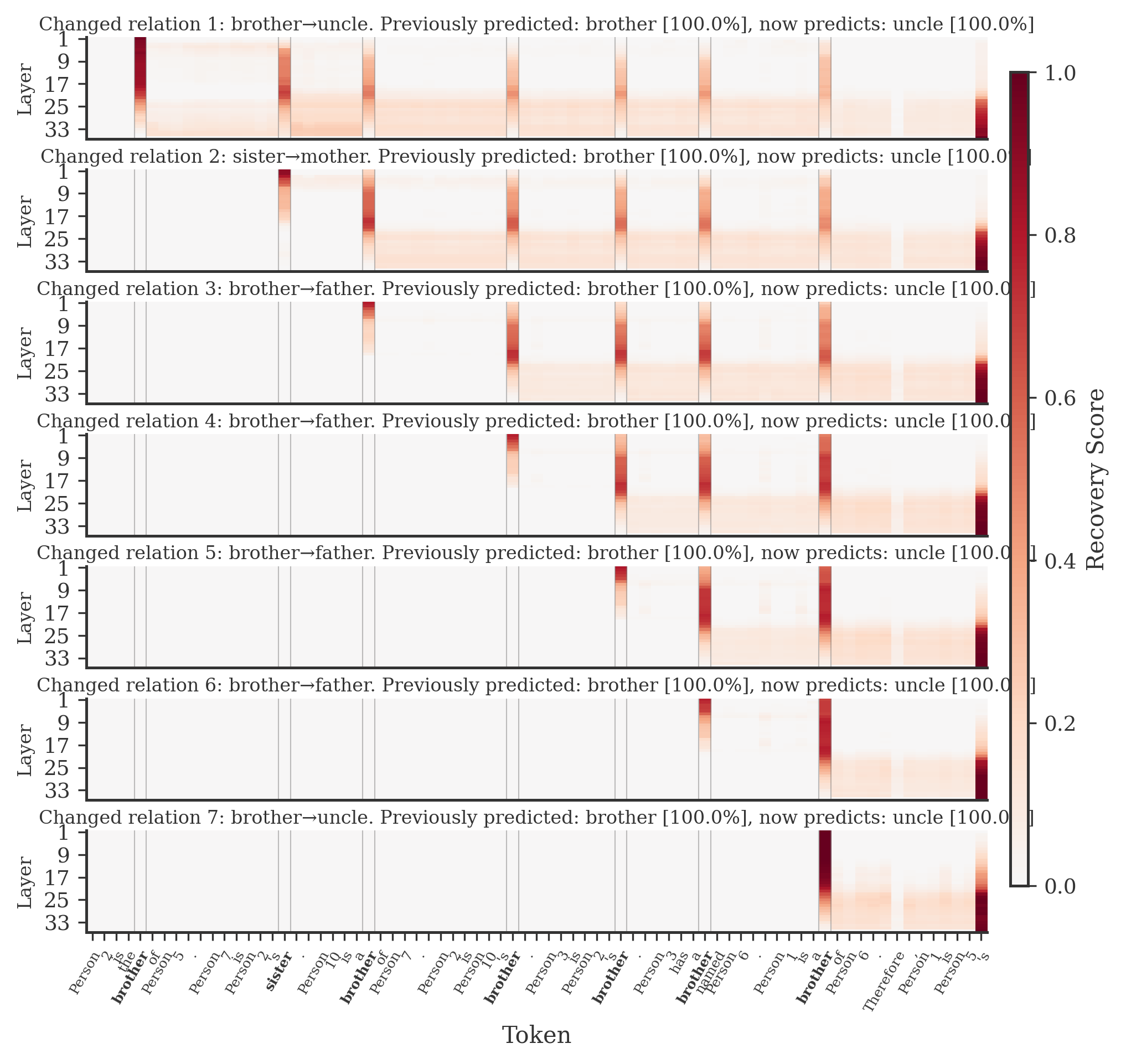}
    \caption{Patch $brother \rightarrow uncle$, full finetuning}
    \label{fig:a}
  \end{subfigure}%
  \hfill
  \begin{subfigure}{0.48\textwidth}
    \centering
    \includegraphics[width=\linewidth]{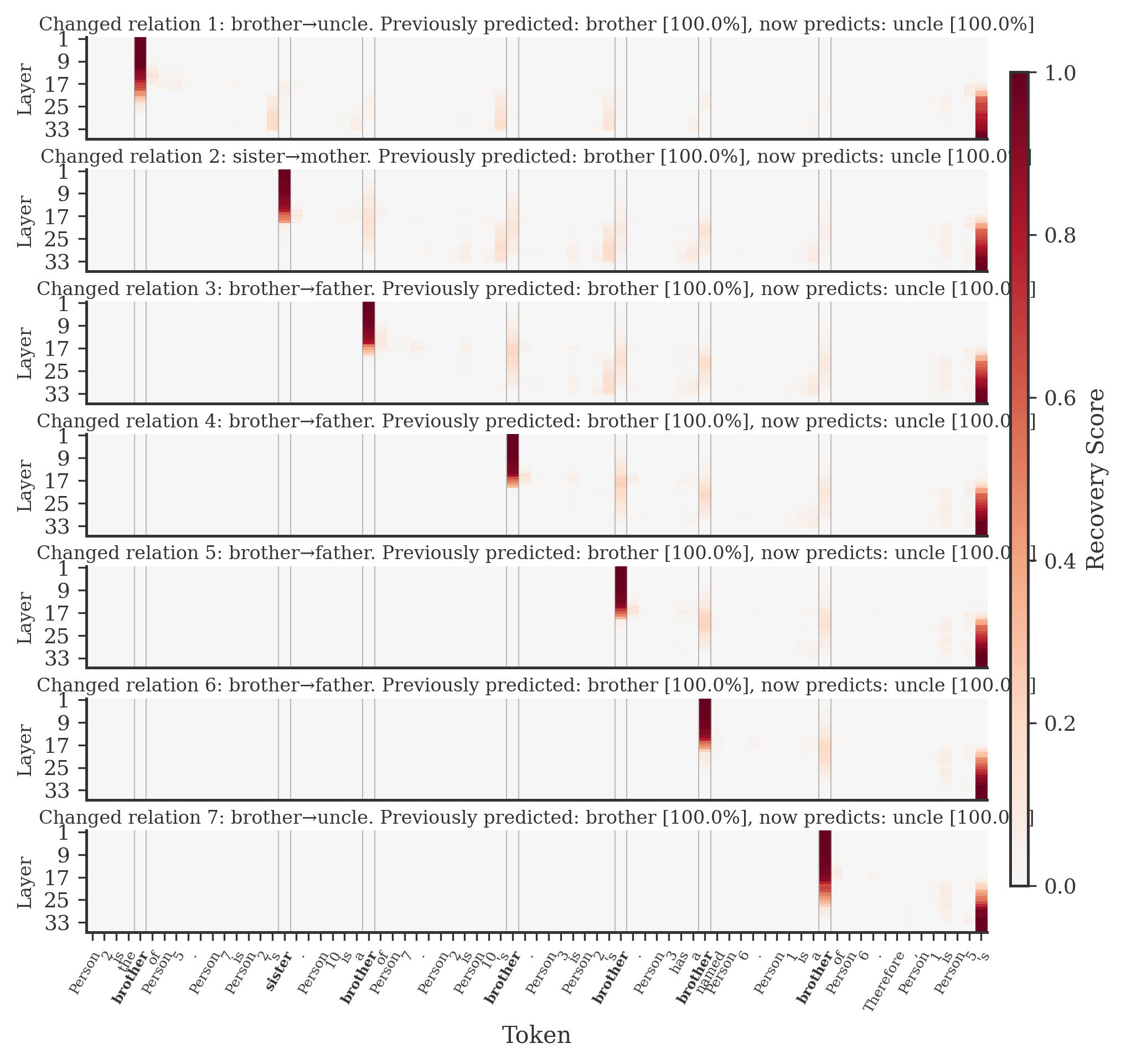}
    \caption{Patch $brother \rightarrow uncle$, LoRA finetuning}
    \label{fig:b}
  \end{subfigure}
    \begin{subfigure}{0.48\textwidth}
    \centering
    \includegraphics[width=\linewidth]{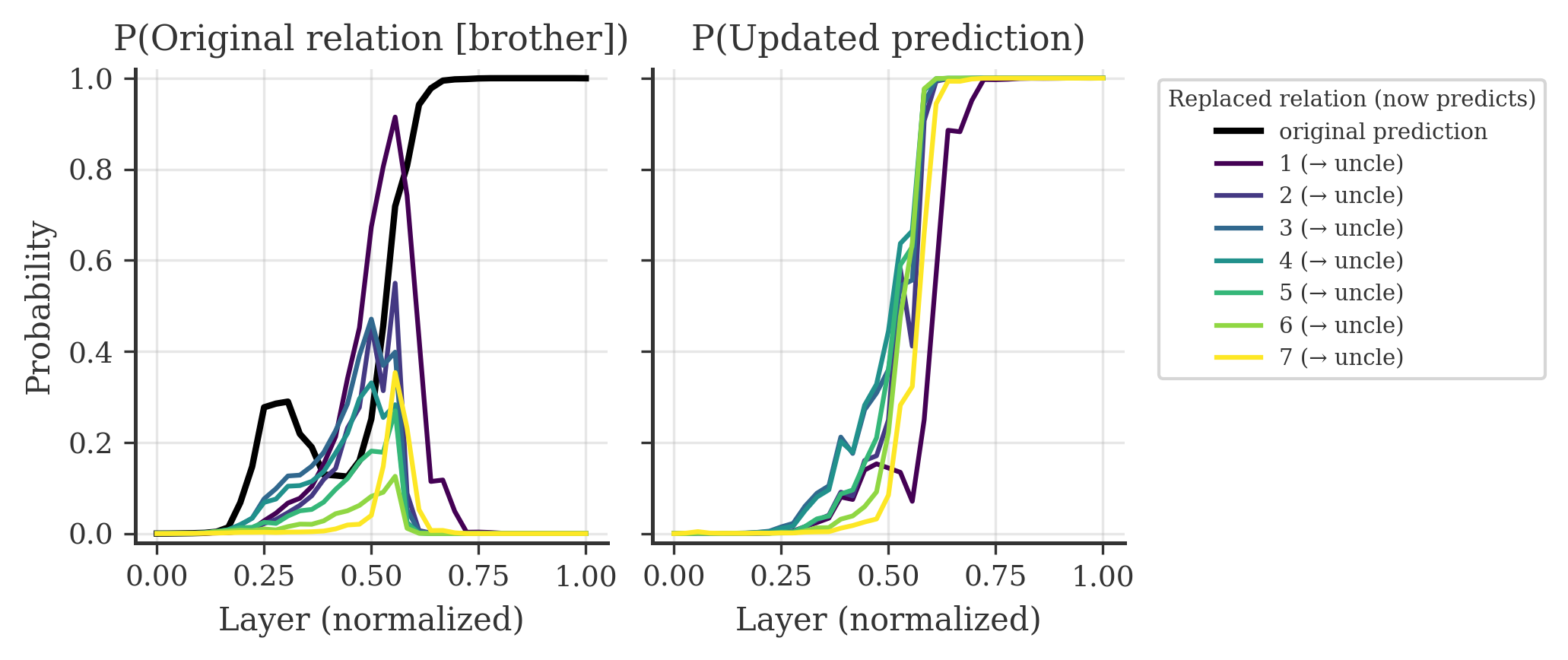}
    \caption{Logit Lens: Full finetuning}
    \label{fig:a}
  \end{subfigure}%
  \hfill
  \begin{subfigure}{0.48\textwidth}
    \centering
    \includegraphics[width=\linewidth]{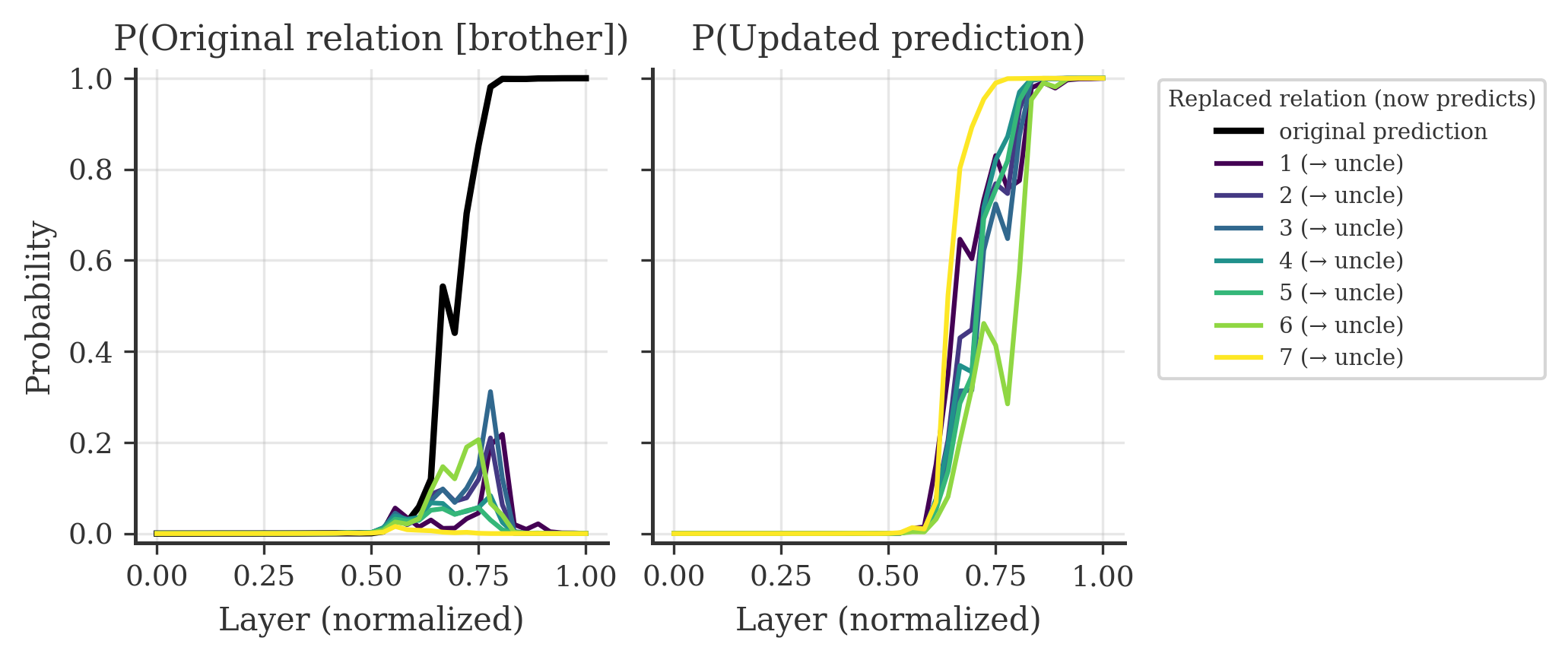}
    \caption{Logit lens: LoRA finetuning}
    \label{fig:b}
  \end{subfigure}
  \caption{Full causal patching and logit-lens results for a 7-hop example using GPT2-large.}
  \label{fig:causalpatch-hop7}
\end{figure}

\begin{figure}[h]
  \centering
    \begin{subfigure}{0.45\textwidth}
    \centering
       \includegraphics[width=\linewidth]{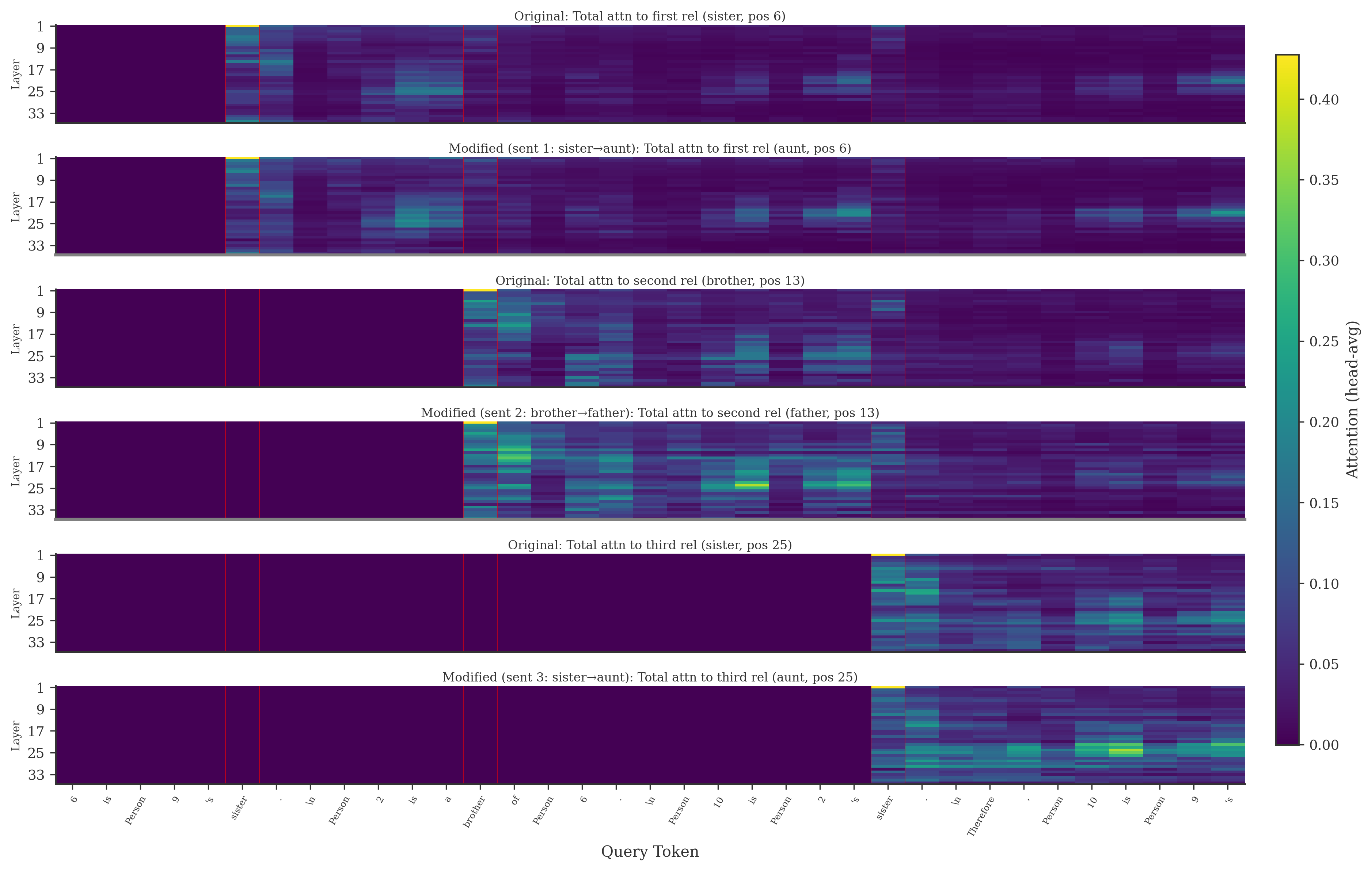}
    \caption{3-hop: LoRA finetuned}
    \label{fig:a}
  \end{subfigure}%
  \hfill
    \begin{subfigure}{0.45\textwidth}
    \centering
       \includegraphics[width=\linewidth]{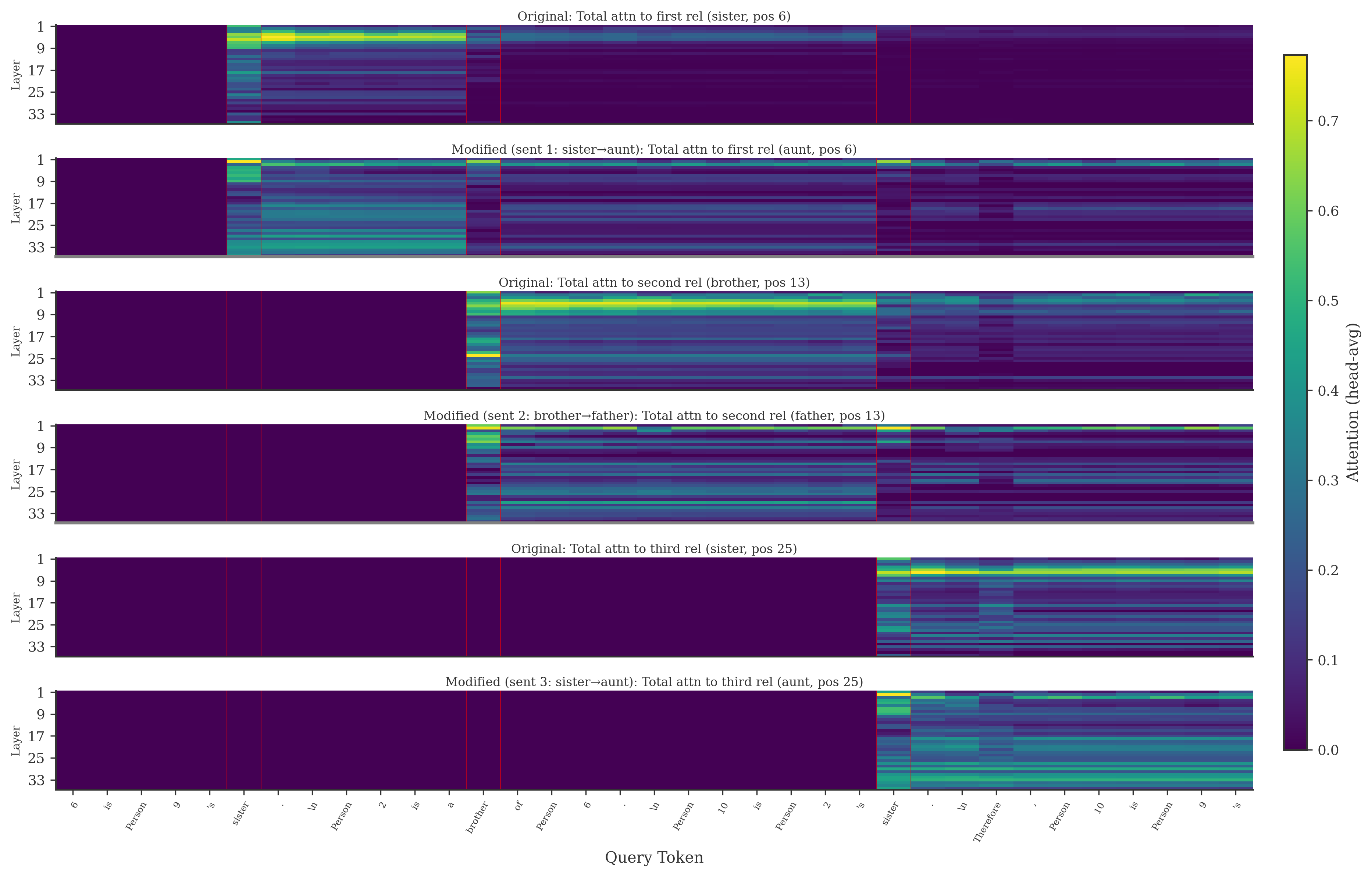}
    \caption{3-hop: Fully finetuned}
    \label{fig:a}
  \end{subfigure}%
  \hfill
      \begin{subfigure}{0.45\textwidth}
    \centering
       \includegraphics[width=\linewidth]{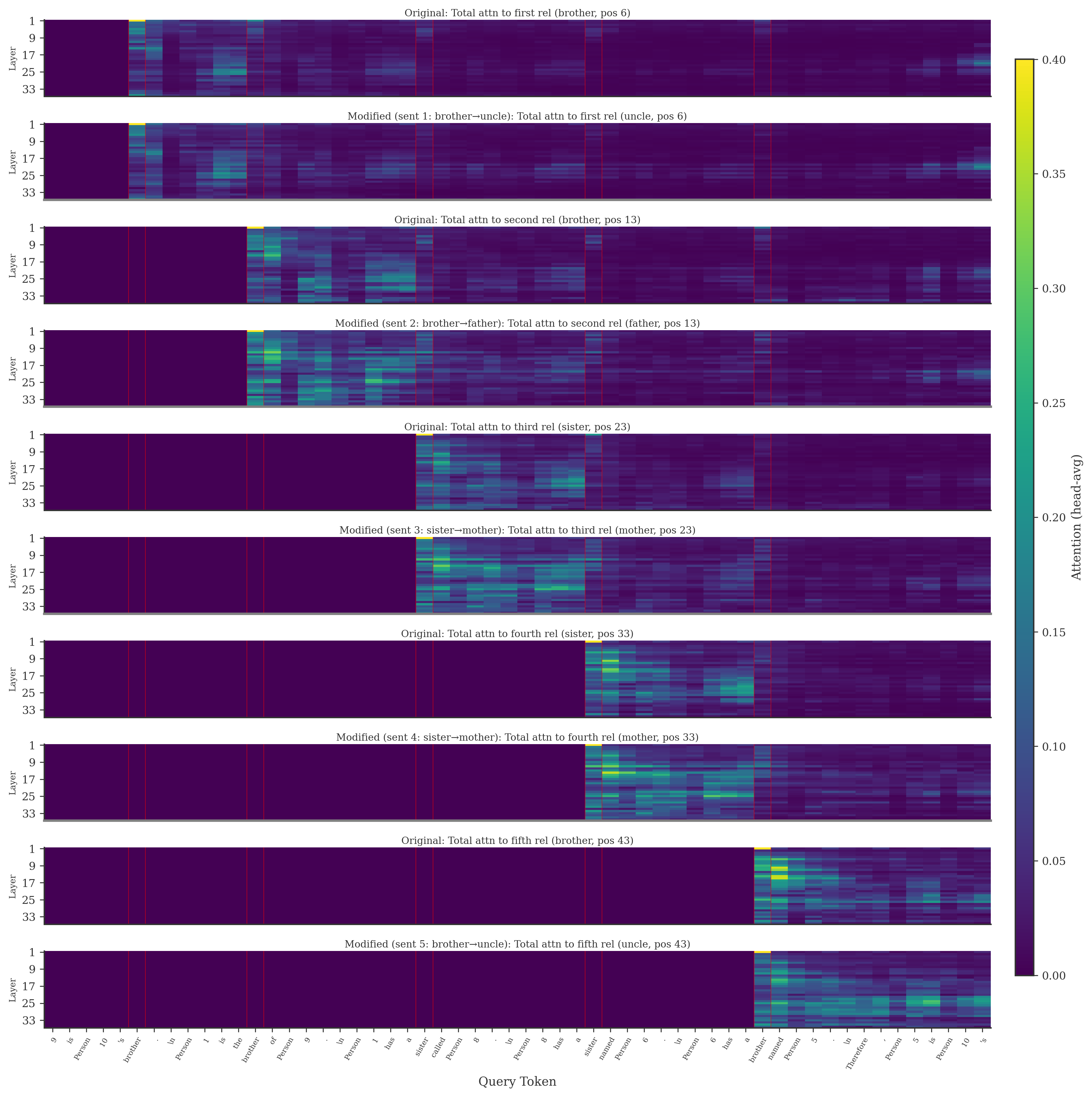}
    \caption{5-hop: LoRA finetuned}
    \label{fig:a}
  \end{subfigure}%
  \hfill
    \begin{subfigure}{0.45\textwidth}
    \centering
       \includegraphics[width=\linewidth]{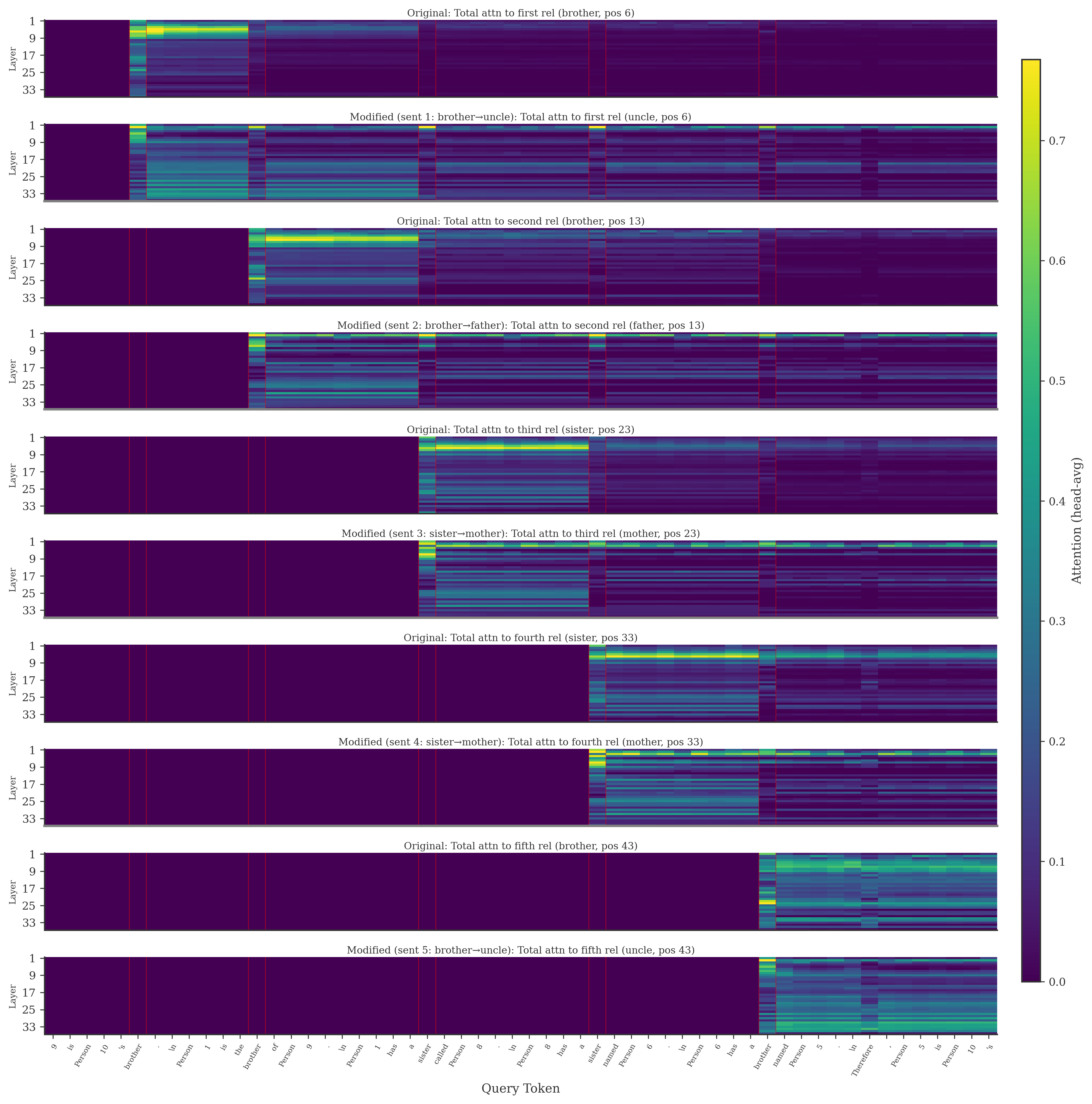}
    \caption{5-hop: Fully finetuned}
    \label{fig:a}
  \end{subfigure}%
  \hfill
      \begin{subfigure}{0.45\textwidth}
    \centering
       \includegraphics[width=\linewidth]{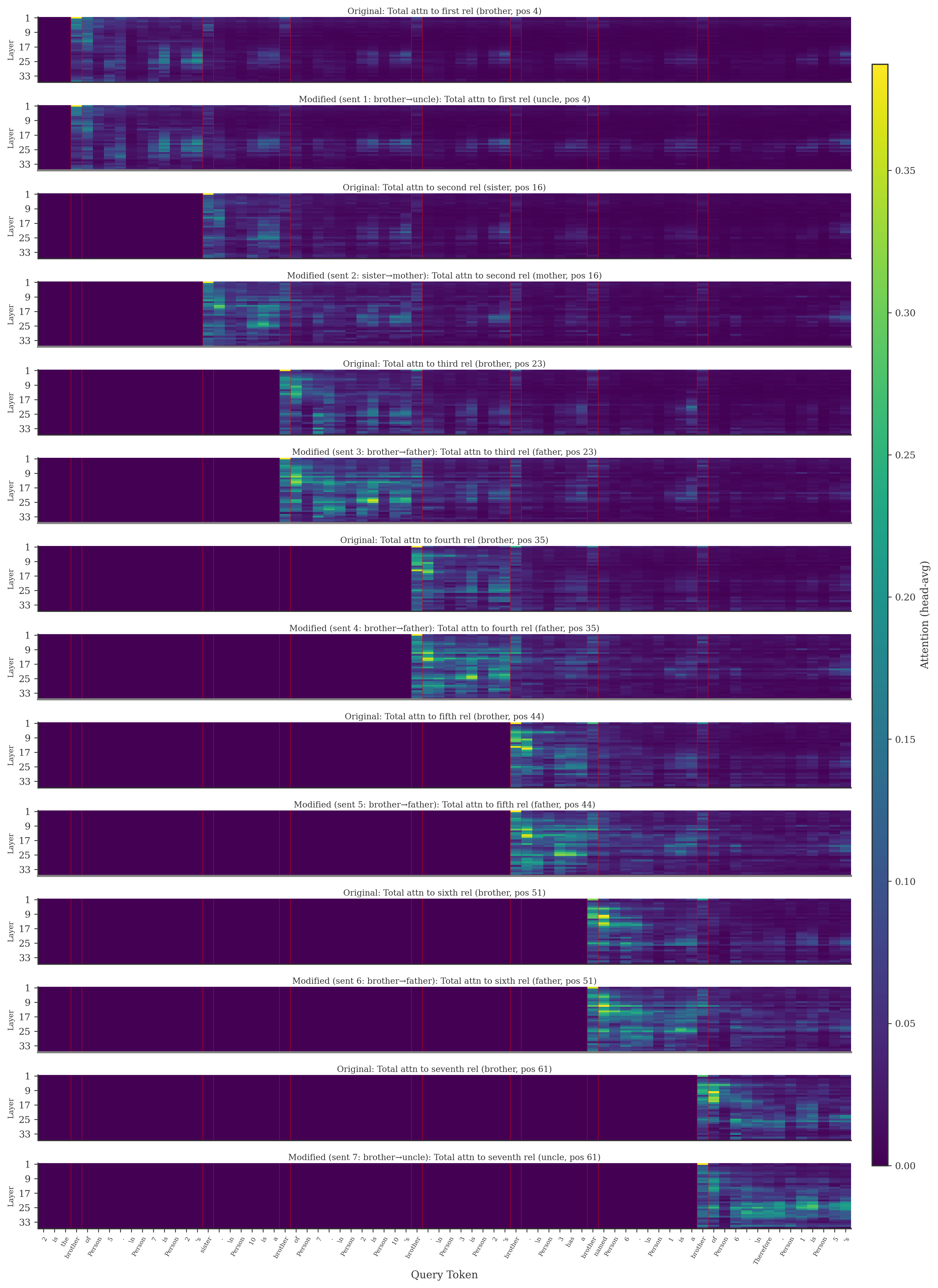}
    \caption{7-hop: LoRA finetuned}
    \label{fig:a}
  \end{subfigure}%
  \hfill
    \begin{subfigure}{0.45\textwidth}
    \centering
       \includegraphics[width=\linewidth]{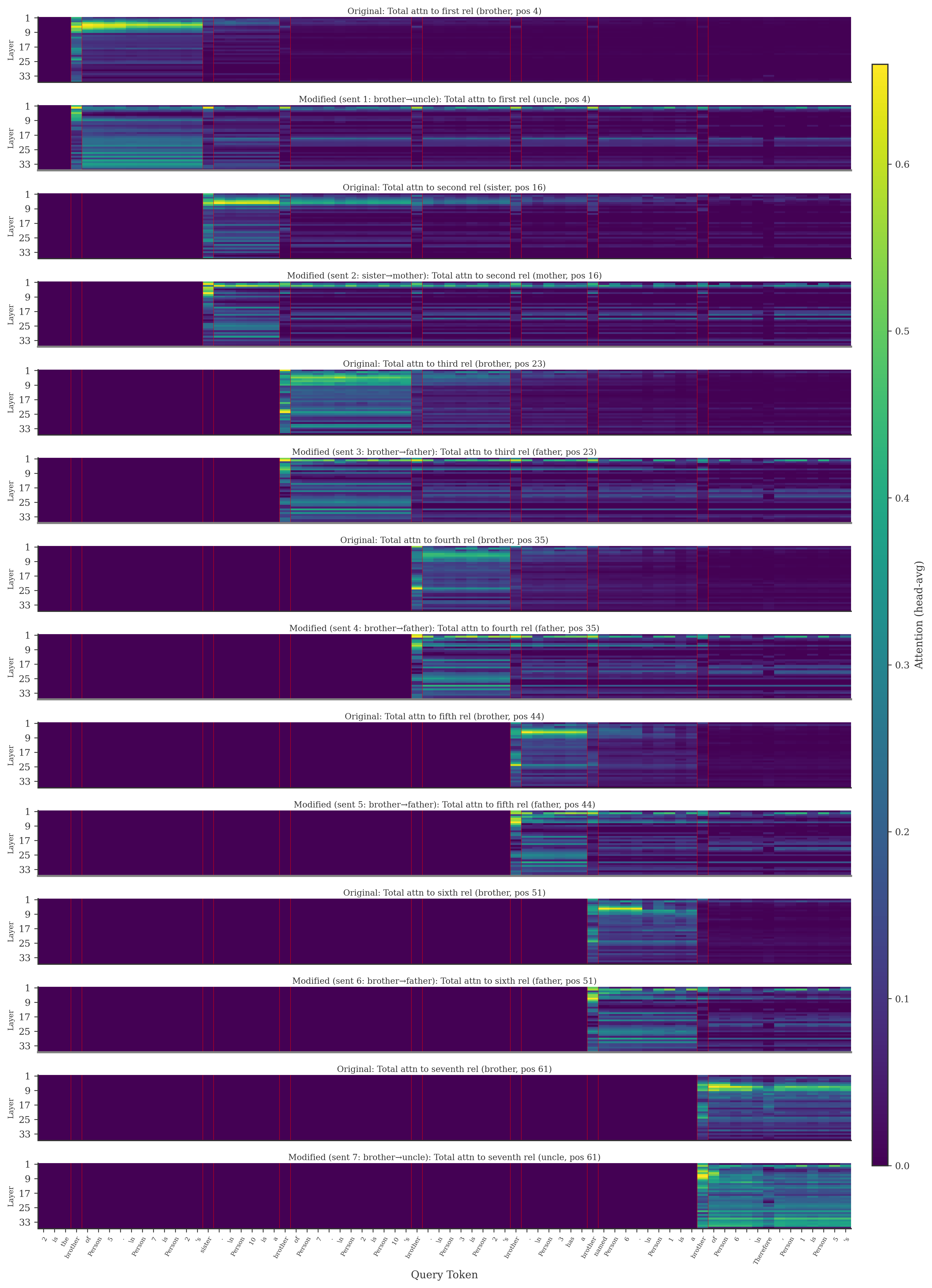}
    \caption{7-hop: Fully finetuned}
    \label{fig:a}
  \end{subfigure}%
  \hfill
  \caption{Attention to key token $t^r$ by layer and query token for the causal patching examples of GPT2-large. Attention patterns differ between sibling and mother/father/uncle/aunt relations (first and second row of each location), more markedly so for the fully finetuned models.}
  \label{fig:attn-patterns}
\end{figure}

\begin{figure}[h]
    \centering
    \includegraphics[width=0.99\linewidth]{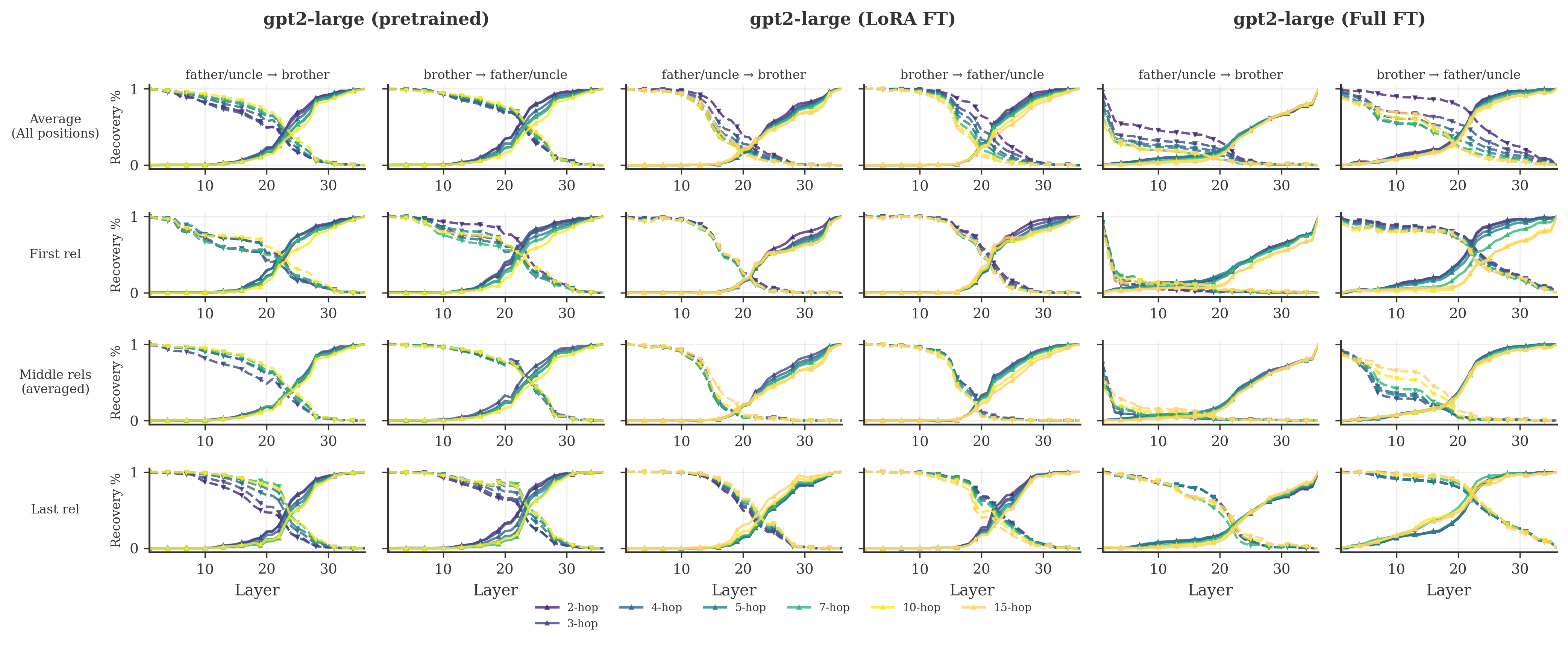}
    \caption{Average recovery score by relation replacement position for different versions of gpt2-large, colored by number of hops, for stories which originally only had siblings. The first row is identical to \cref{fig:cp-gpt2}(b) in the main text; this figure supplements the main text figure by splitting the averaged results across replacement positions.}
    \label{fig:placeholder}
\end{figure}

\begin{figure}[h]
  \centering
  \begin{subfigure}{0.99\textwidth}
    \centering
       \includegraphics[width=\linewidth]{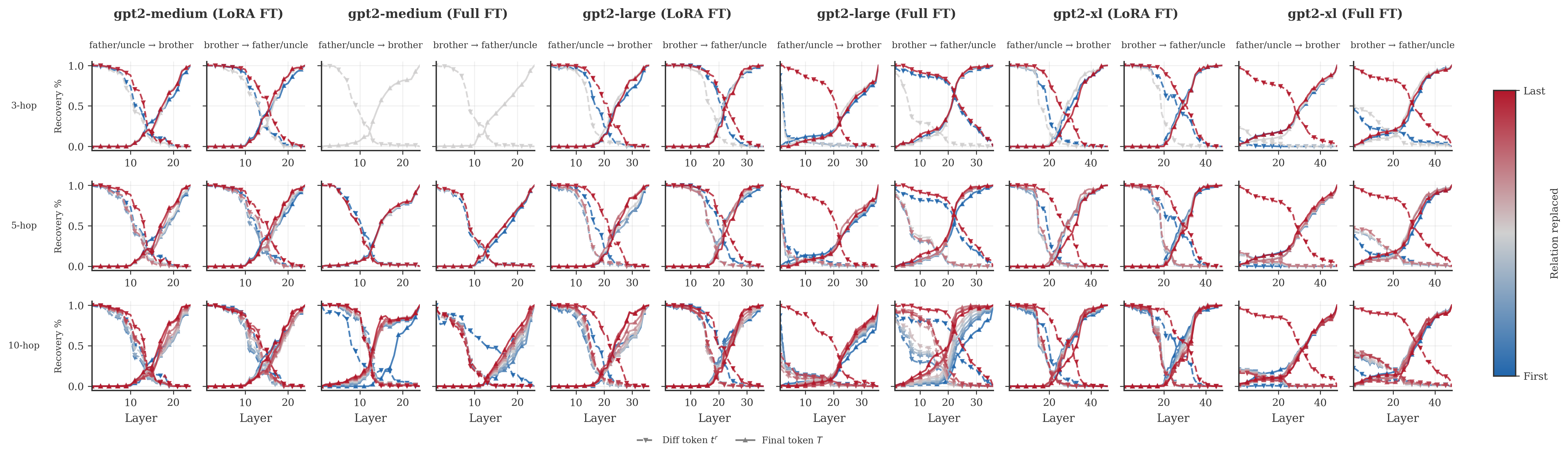}
    \caption{Average recovery score by replaced relation across different hops (rows), colored by relation replaced.}
    \label{fig:a}
  \end{subfigure}%
  \hfill
  \begin{subfigure}{0.99\textwidth}
    \centering
     \includegraphics[width=\linewidth]{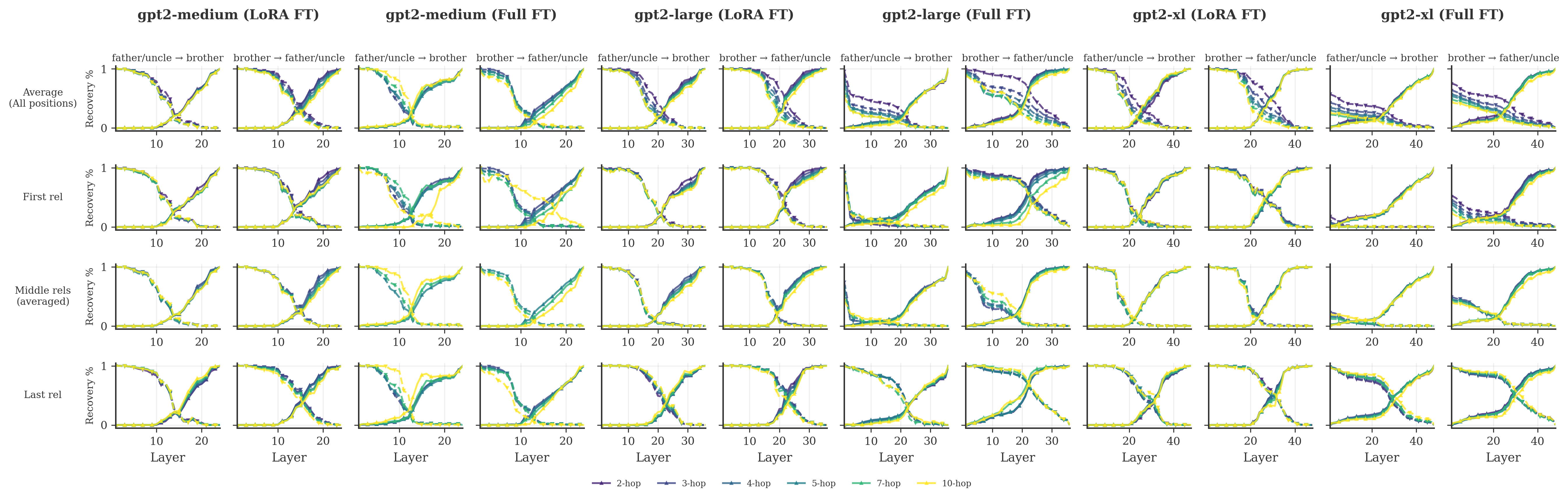}
    \caption{Average recovery score across all relation replacement positions, colored by number of hops.}
    \label{fig:b}
  \end{subfigure}
  \caption{Causal patching analyses for finetuned GPT2 models of different sizes. Average recovery score at the replaced token $t^r$ (dashed line) and the final token $T$ (solid line) by model depth, for stories with only sibling relations. Note: For the finetuned GPT2-medium, the top prediction never flipped for some replaced relationships, leading to missing lines in both plots.}
  \label{fig:cp-gpt2all-siblng}
\end{figure}

\begin{figure}[ht]
  \centering
  \begin{subfigure}{0.48\textwidth}
    \centering
    \includegraphics[width=\linewidth]{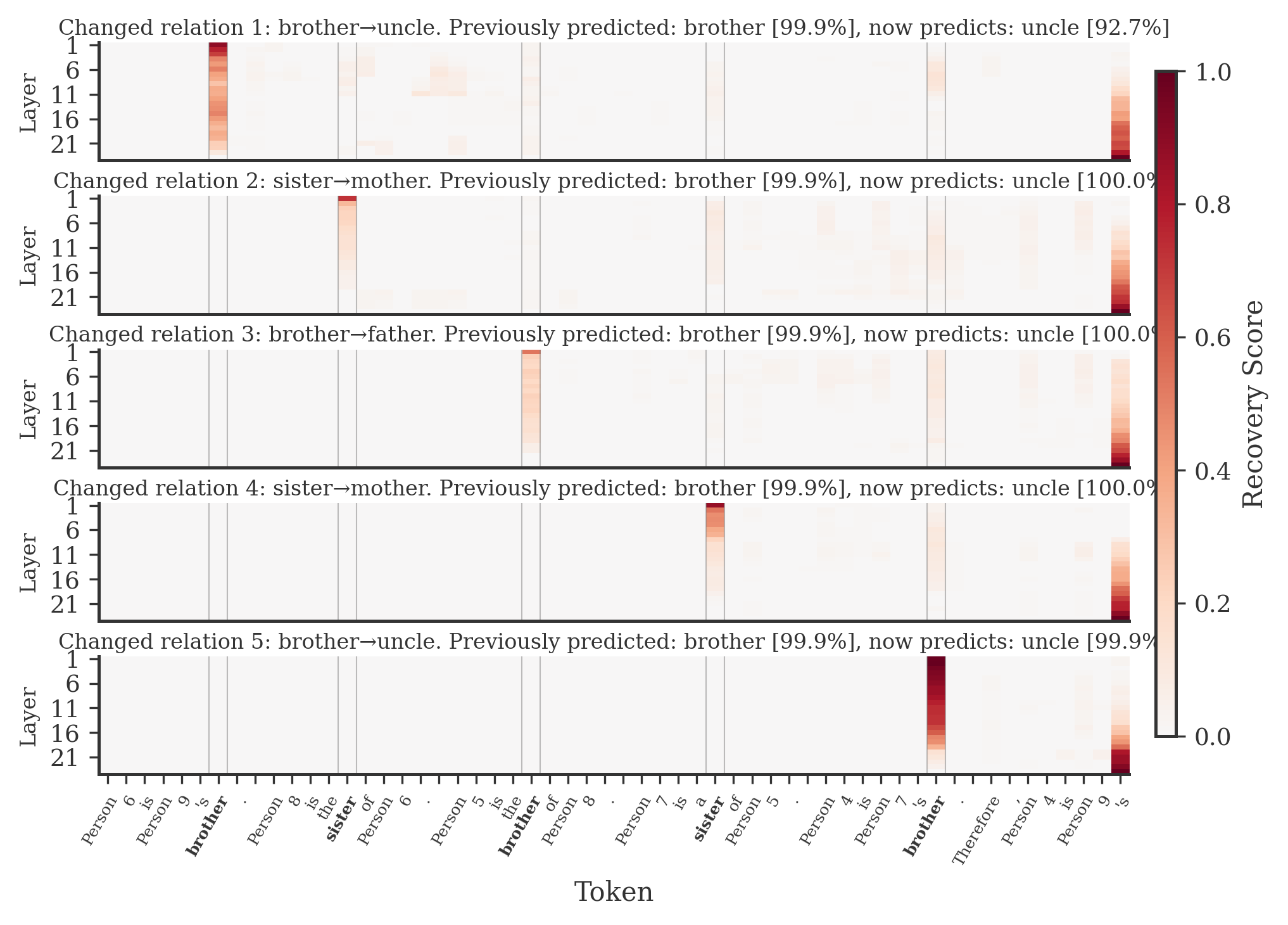}
    \caption{Patch: $uncle \rightarrow brother$, full finetuning}
    \label{fig:a}
  \end{subfigure}%
  \hfill
  \begin{subfigure}{0.48\textwidth}
    \centering
    \includegraphics[width=\linewidth]{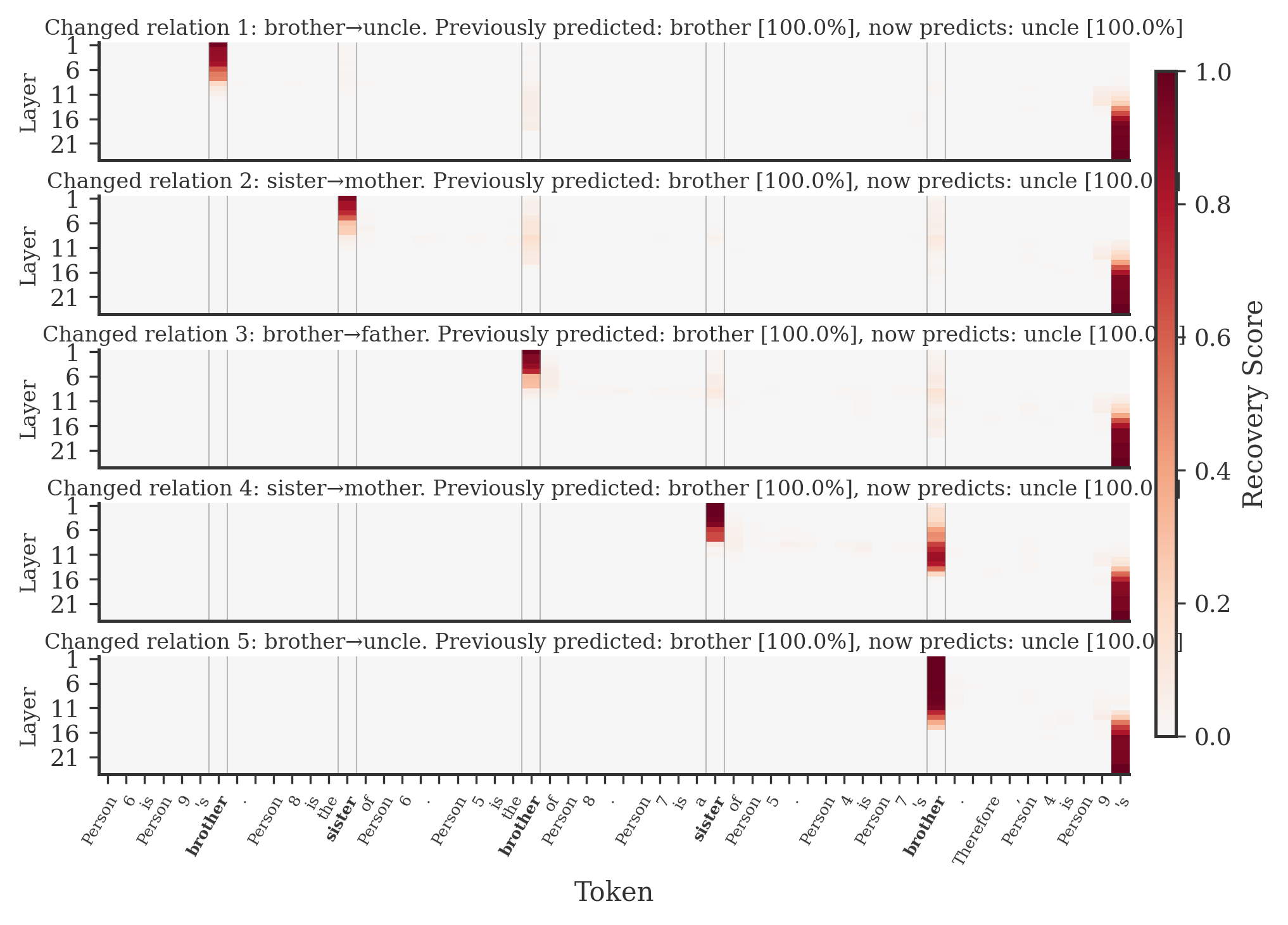}
    \caption{Patch:  $uncle \rightarrow brother$, LoRA finetuning}
    \label{fig:b}
  \end{subfigure}
    \begin{subfigure}{0.48\textwidth}
    \centering
    \includegraphics[width=\linewidth]{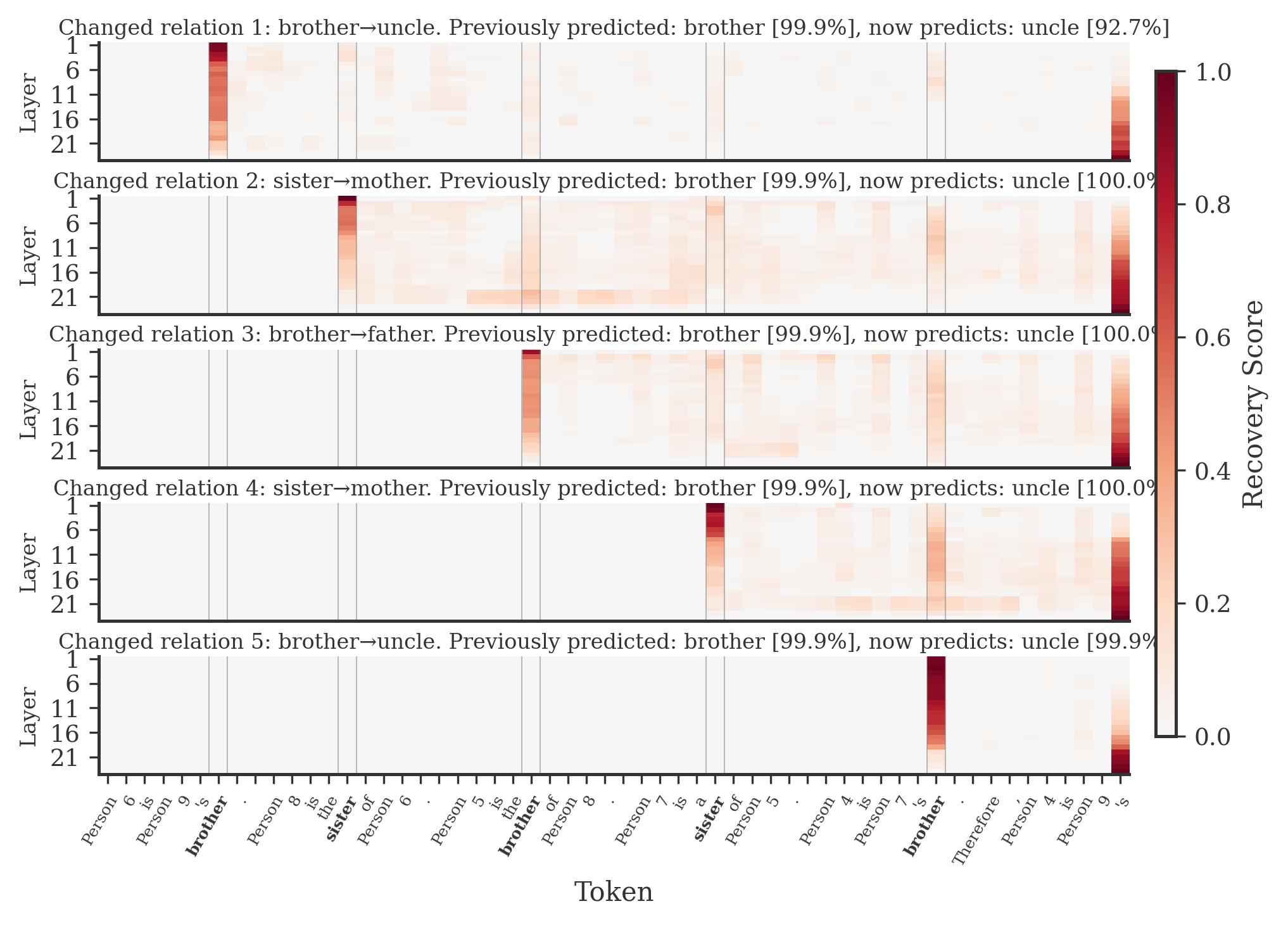}
    \caption{Patch $brother \rightarrow uncle$, full finetuning}
    \label{fig:a}
  \end{subfigure}%
  \hfill
  \begin{subfigure}{0.48\textwidth}
    \centering
    \includegraphics[width=\linewidth]{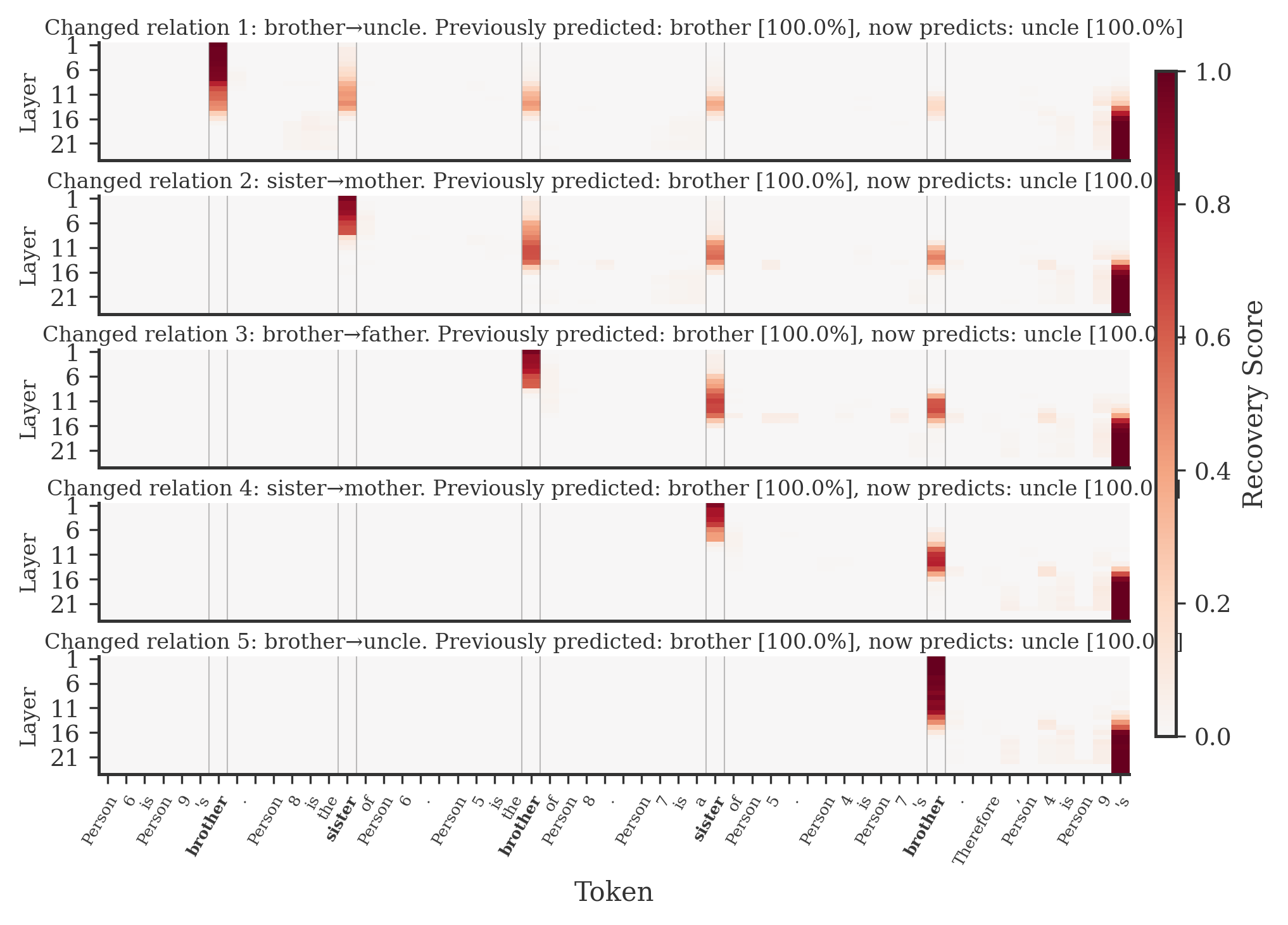}
    \caption{Patch $brother \rightarrow uncle$, LoRA finetuning}
    \label{fig:b}
  \end{subfigure}
    \begin{subfigure}{0.48\textwidth}
    \centering
    \includegraphics[width=\linewidth]{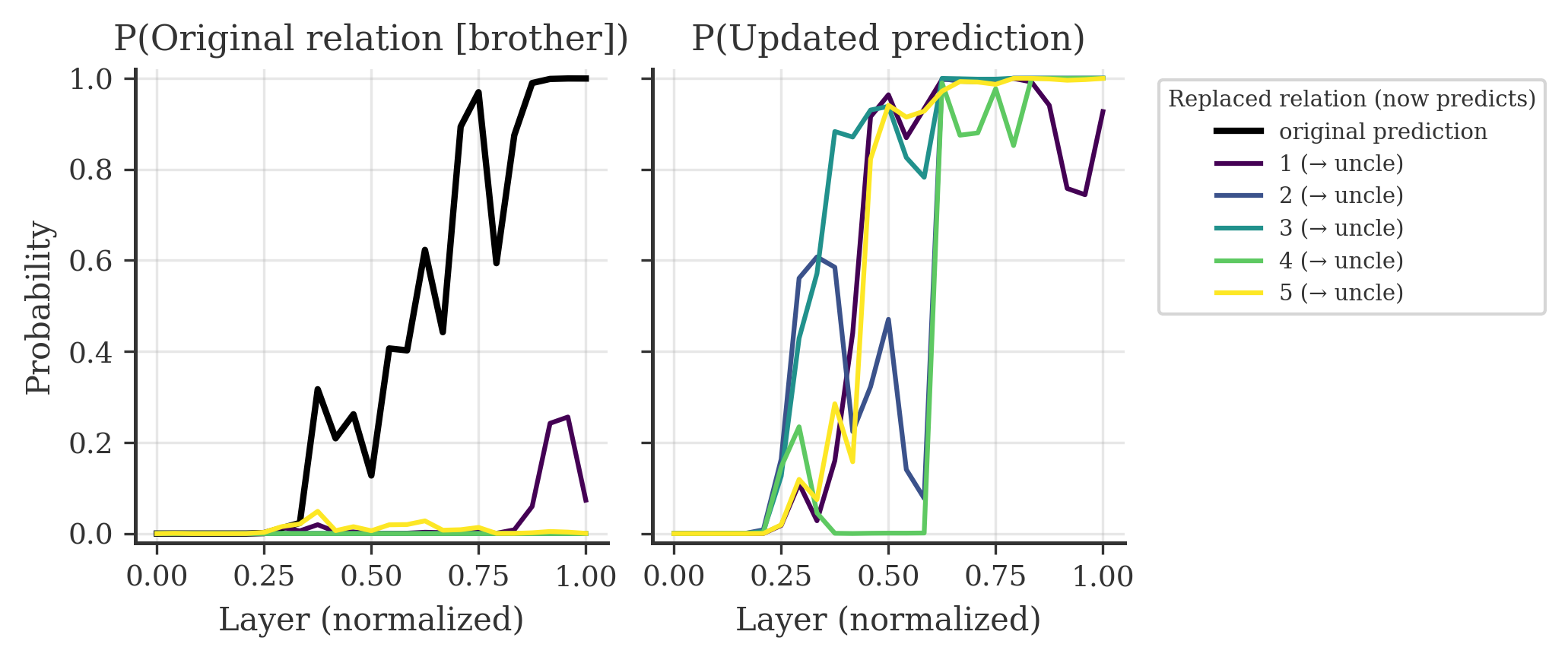}
    \caption{Logit Lens: Full finetuning}
    \label{fig:a}
  \end{subfigure}%
  \hfill
  \begin{subfigure}{0.48\textwidth}
    \centering
    \includegraphics[width=\linewidth]{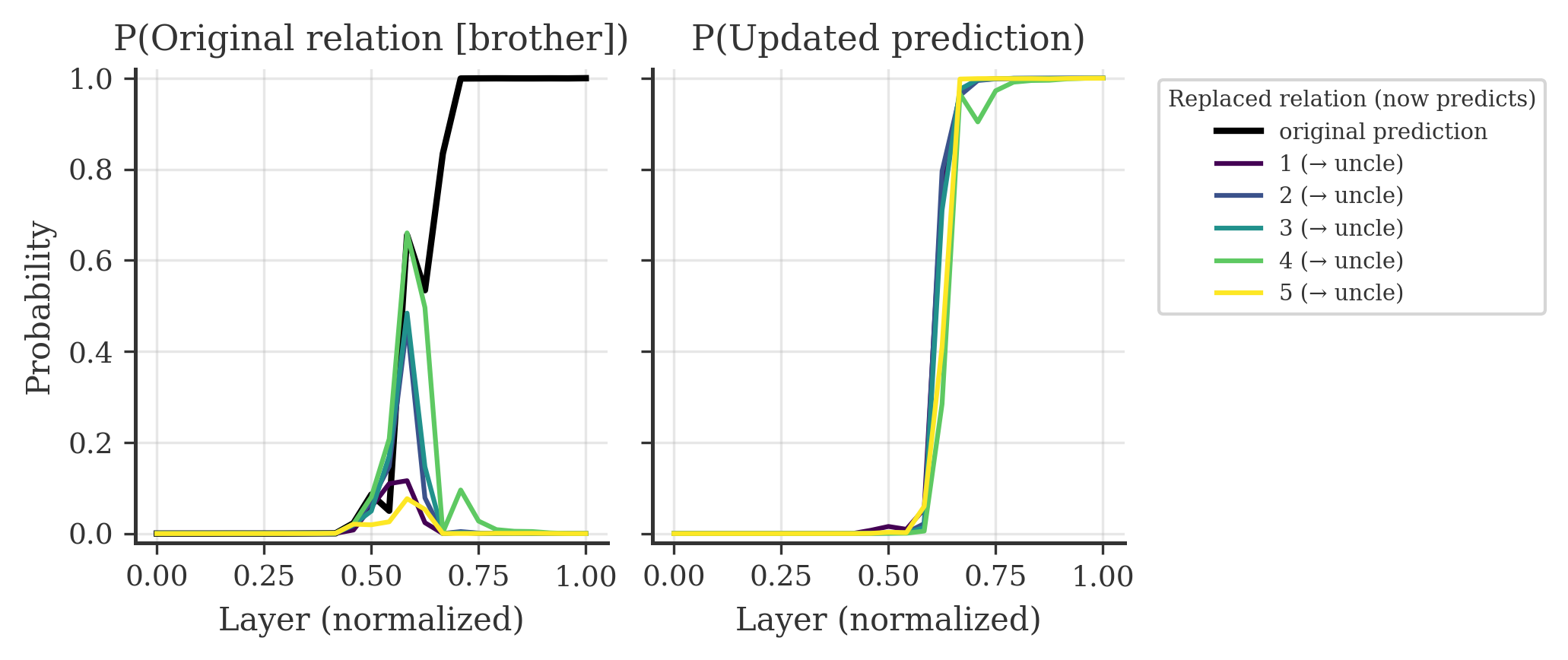}
    \caption{Logit lens: LoRA finetuning}
    \label{fig:b}
  \end{subfigure}
  \caption{Full causal patching and logit-lens results for a 5-hop example using pythia-1.4B.}
  \label{fig:causalpatch-pythia}
\end{figure}

\begin{figure}[h]
  \centering
  \begin{subfigure}{0.99\textwidth}
    \centering
       \includegraphics[width=\linewidth]{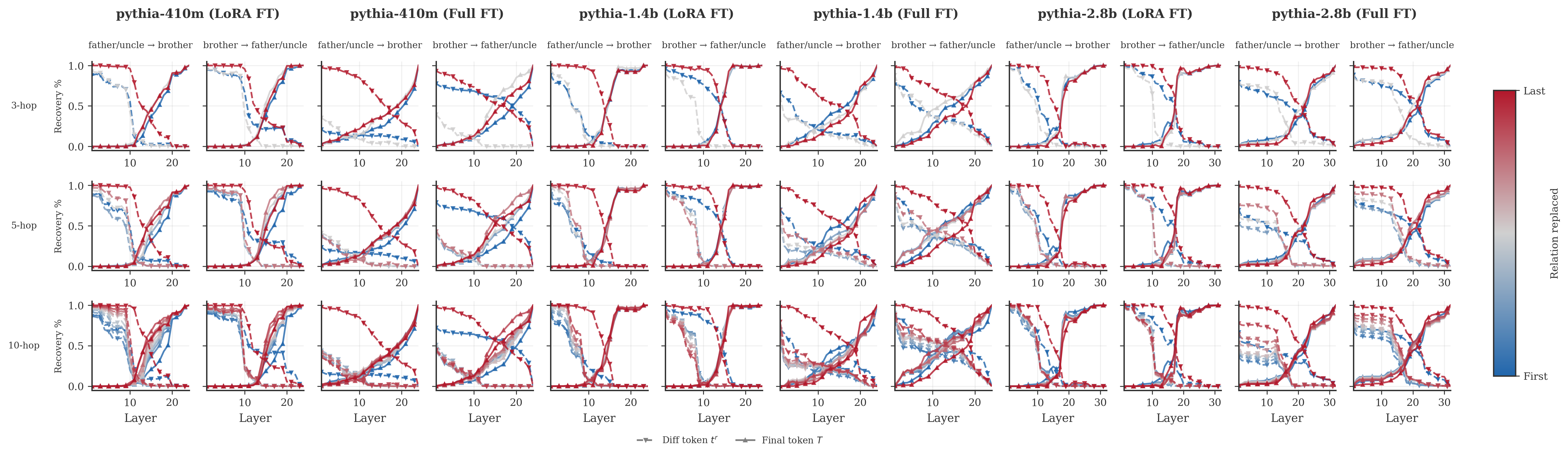}
    \caption{Average recovery score by replaced relation across different hops (rows), colored by relation replaced.}
    \label{fig:a}
  \end{subfigure}%
  \hfill
  \begin{subfigure}{0.99\textwidth}
    \centering
     \includegraphics[width=\linewidth]{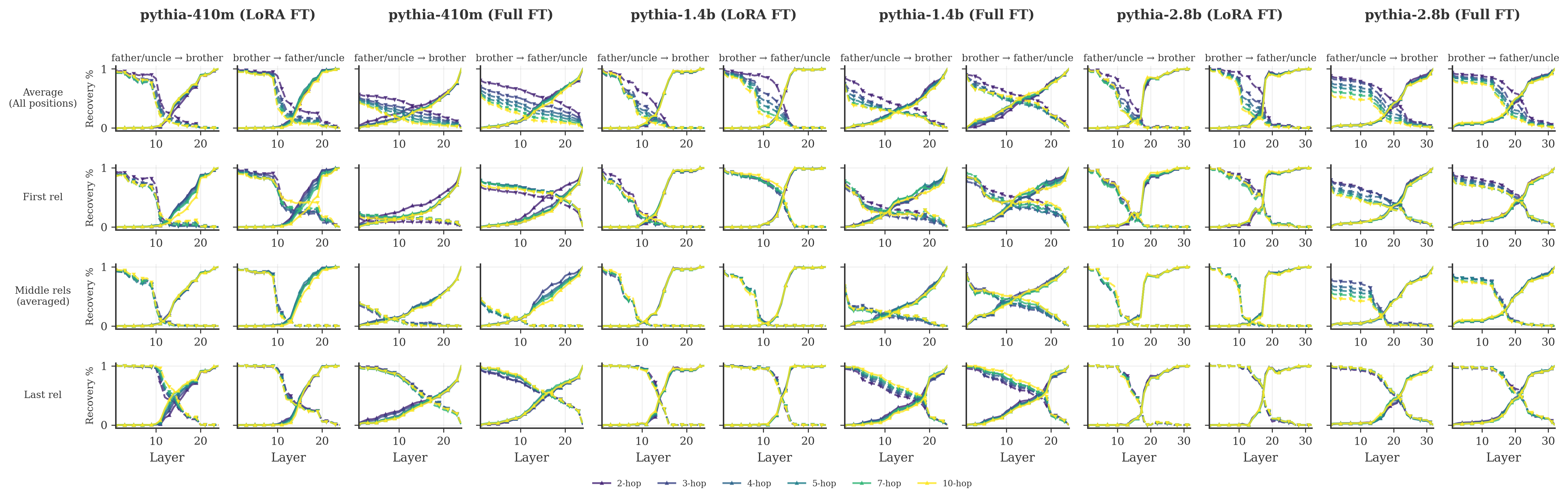}
    \caption{Average recovery score across all relation replacement positions, colored by number of hops.}
    \label{fig:b}
  \end{subfigure}
  \caption{Causal patching analyses for finetuned pythia models. Average recovery score at the replaced token $t^r$ (dashed line) and the final token $T$ (solid line) by model depth, for stories with only sibling relations.}
  \label{fig:cp-pythia-siblng}
\end{figure}

\FloatBarrier
\subsection{Additional causal patching results with more complex family relation stories}\label{app:morestories}
\begin{figure}[h]
  \centering
  \begin{subfigure}{0.99\textwidth}
    \centering
       \includegraphics[width=\linewidth]{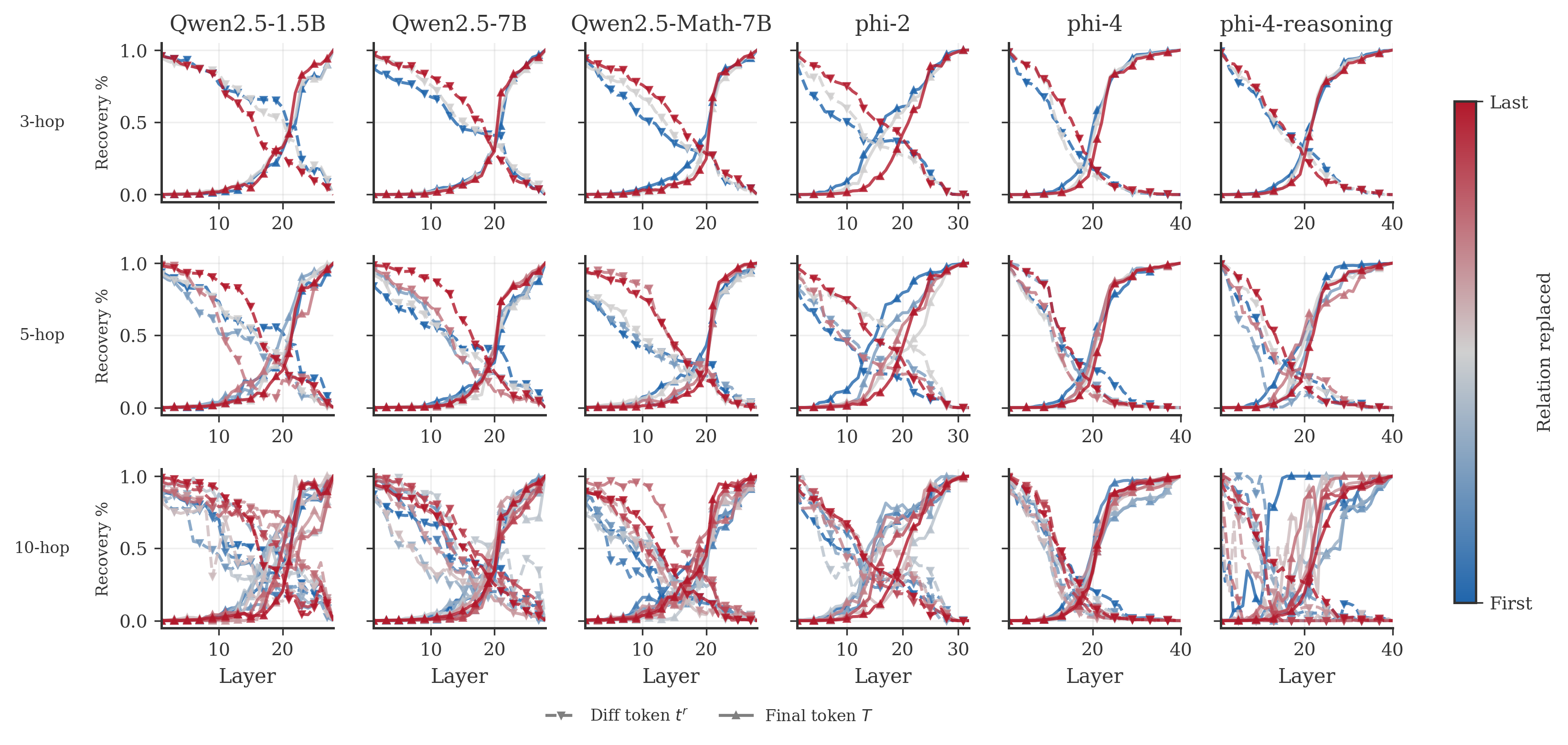}
    \caption{Average recovery score by replaced relation across different hops (rows), colored by relation replaced.}
    \label{fig:a}
  \end{subfigure}%
  \hfill
  \begin{subfigure}{0.99\textwidth}
    \centering
     \includegraphics[width=\linewidth]{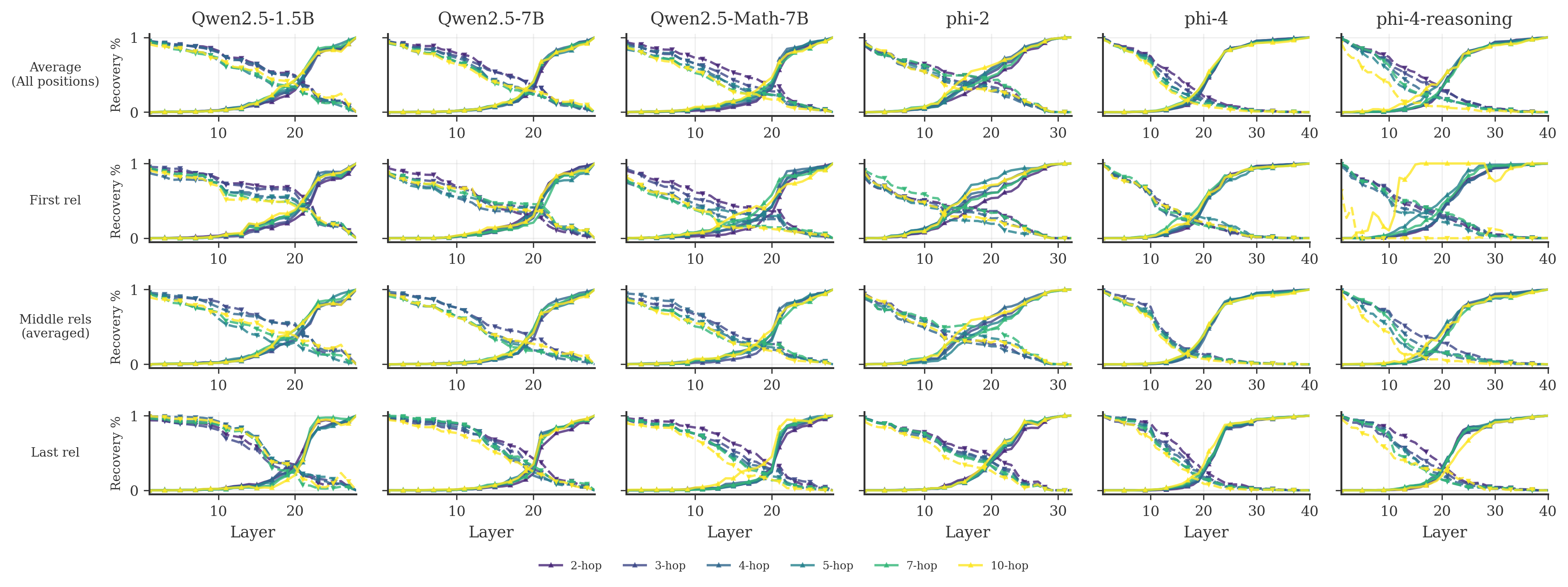}
    \caption{Average recovery score across all relation replacement positions, colored by number of hops.}
    \label{fig:b}
  \end{subfigure}
  \caption{Causal patching analyses for pretrained models. Average recovery score at the replaced token $t^r$ (dashed line) and the final token $T$ (solid line) by model depth, for stories with all standard CLUTRR relation types, where existing relations are now replaced with "brother/sister". That is, this figure replicates the analyses in \cref{fig:cp-pretrained} without relying on stories that contain brother/sister relations only -- with consistent trends.}
  \label{fig:cp-pretrained-allrels}
\end{figure}

\begin{figure}[h]
  \centering
  \begin{subfigure}{0.99\textwidth}
    \centering
       \includegraphics[width=\linewidth]{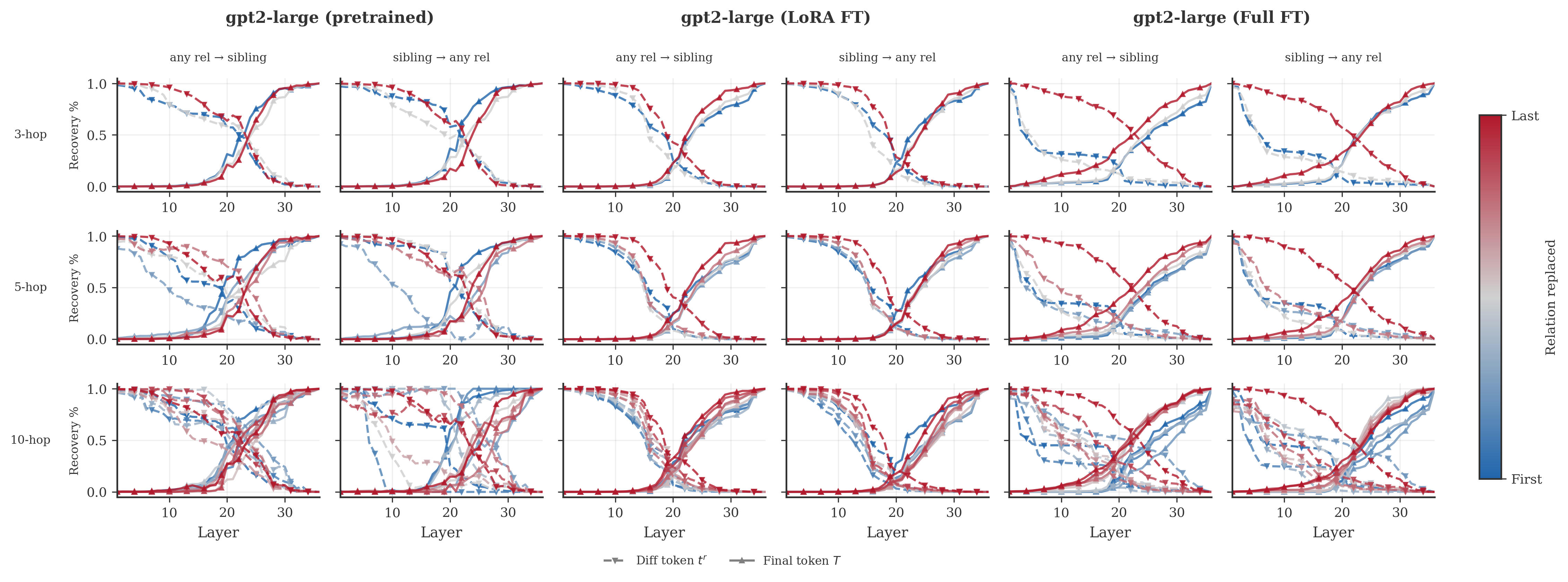}
    \caption{Average recovery score by replaced relation across different hops (rows), colored by relation replaced.}
    \label{fig:a}
  \end{subfigure}%
  \hfill
  \begin{subfigure}{0.99\textwidth}
    \centering
     \includegraphics[width=\linewidth]{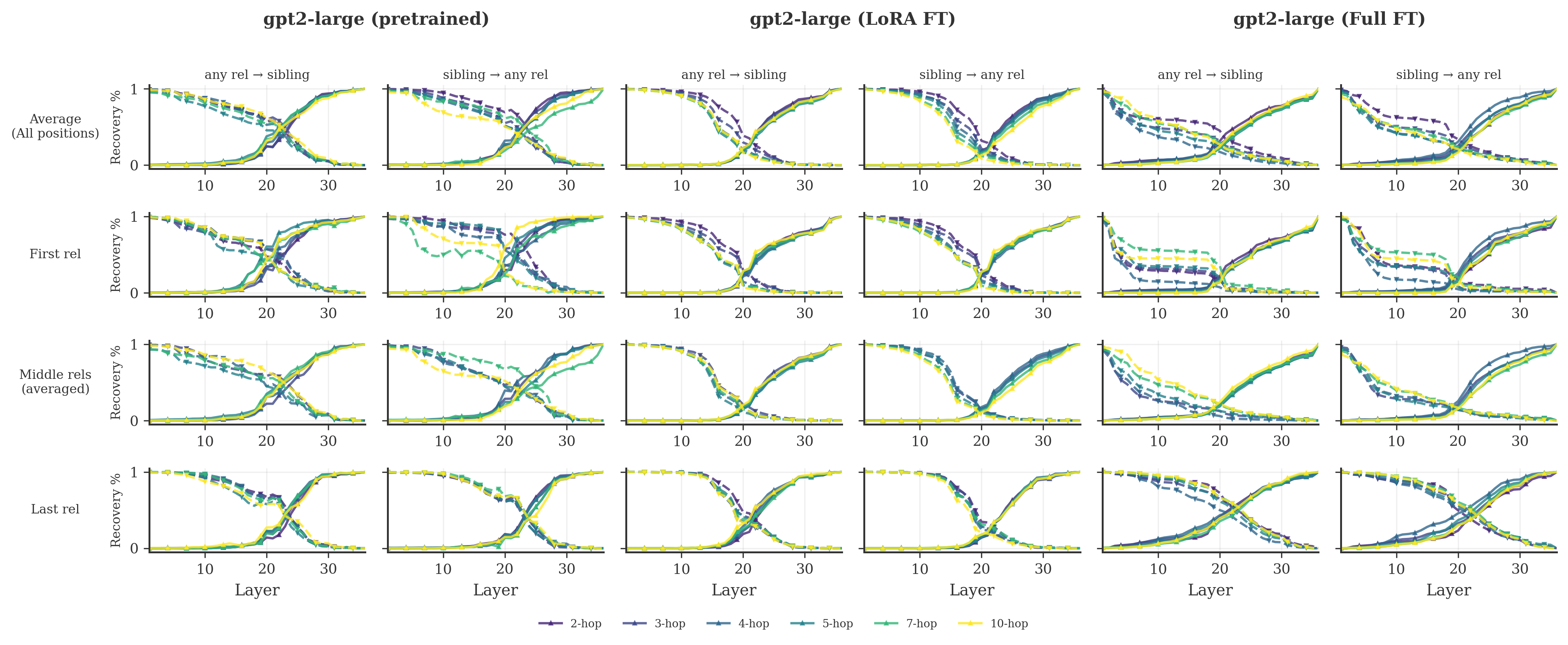}
    \caption{Average recovery score across all relation replacement positions, colored by number of hops.}
    \label{fig:b}
  \end{subfigure}
  \caption{Causal patching analyses for GPT2-large. Average recovery score at the replaced token $t^r$ (dashed line) and the final token $T$ (solid line) by model depth, for stories with all standard CLUTRR relation types, where existing relations are now replaced with "brother/sister". That is, this figure replicates the analyses in \cref{fig:cp-gpt2} without relying on stories that contain brother/sister relations only. General trends are consistent with the main text except that now there is no visible difference between directions of patching. This is expected, as no direction is necessarily easier due to more involved family chains being present either way.}
  \label{fig:cp-finetuned-allrels}
\end{figure}

\FloatBarrier

\subsection{Additional analyses of hidden states}\label{app:repr-metrics}
In this section, we analyse the behaviour of the hidden states directly, replicating part of the analyses of \cite{csordas2025language, hu2025affects} in our setting. Specifically, we monitor the relative contribution of each layer $\Delta_l = h_l - h_{l-1}$ to the residual stream as $\frac{||\Delta_l||_2}{||h_{l-1}||_2}$, and the similarity of the update to the residual stream as $\text{cossim}(\Delta_l, h_{l-1})$. We compute these quantities both for the hidden state at the final token $T$ only, and averaged across all tokens in the prompt. Note that across all plots we cut off the x-axis at the first 25\% of layers and the final layer to ensure readability: as observed in \cite{csordas2025language}, the contribution of the first few and final layers to the residual stream is much larger than that of intermediate layers, which makes smaller relative differences across hop counts indiscernible otherwise.

\begin{figure}[h]
  \centering
    \begin{subfigure}{0.99\textwidth}
    \centering
       \includegraphics[width=\linewidth]{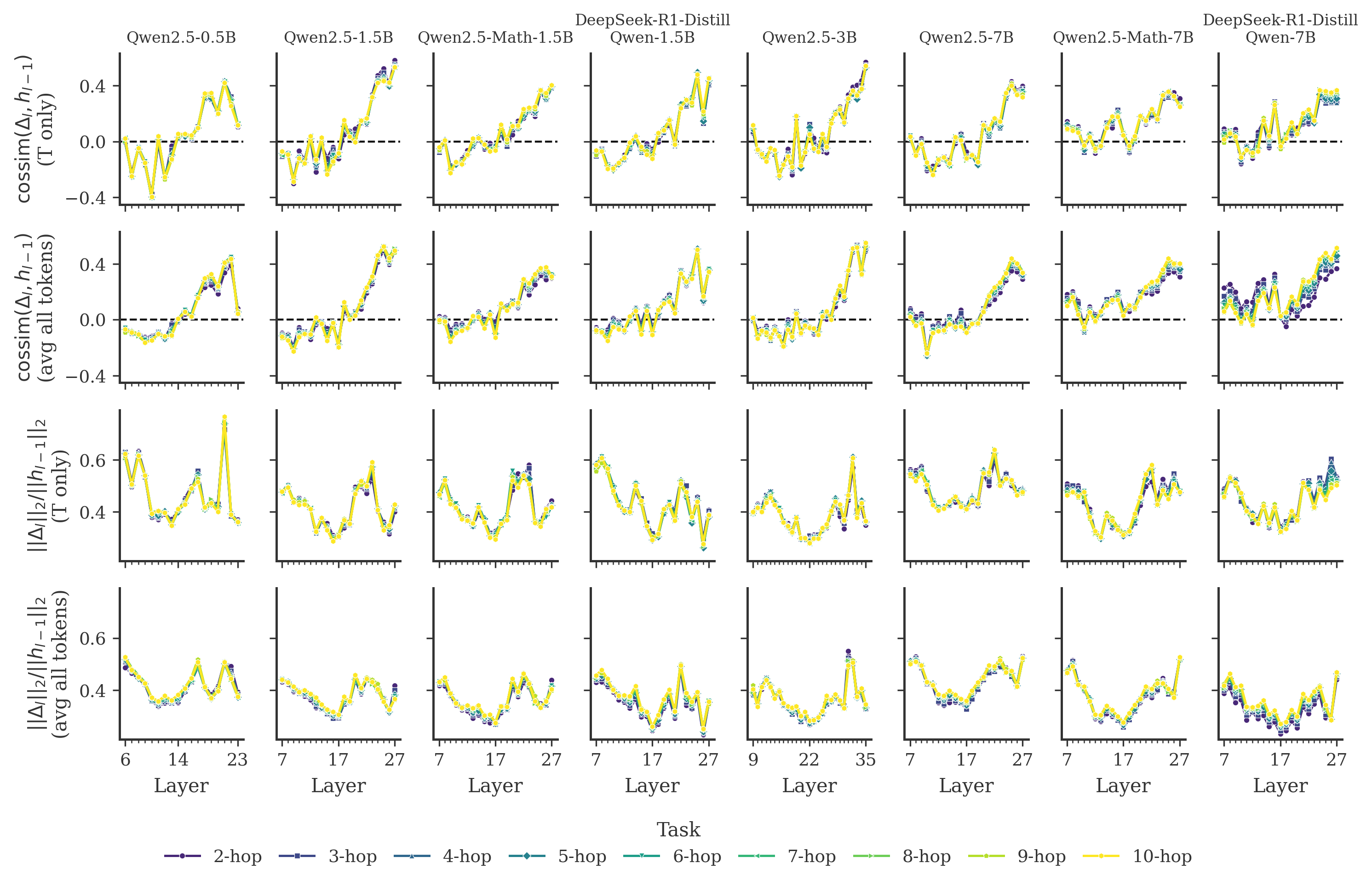}
    \caption{Qwen2.5 family}
    \label{fig:a}
  \end{subfigure}%
  \hfill
      \begin{subfigure}{0.99\textwidth}
    \centering
       \includegraphics[width=\linewidth]{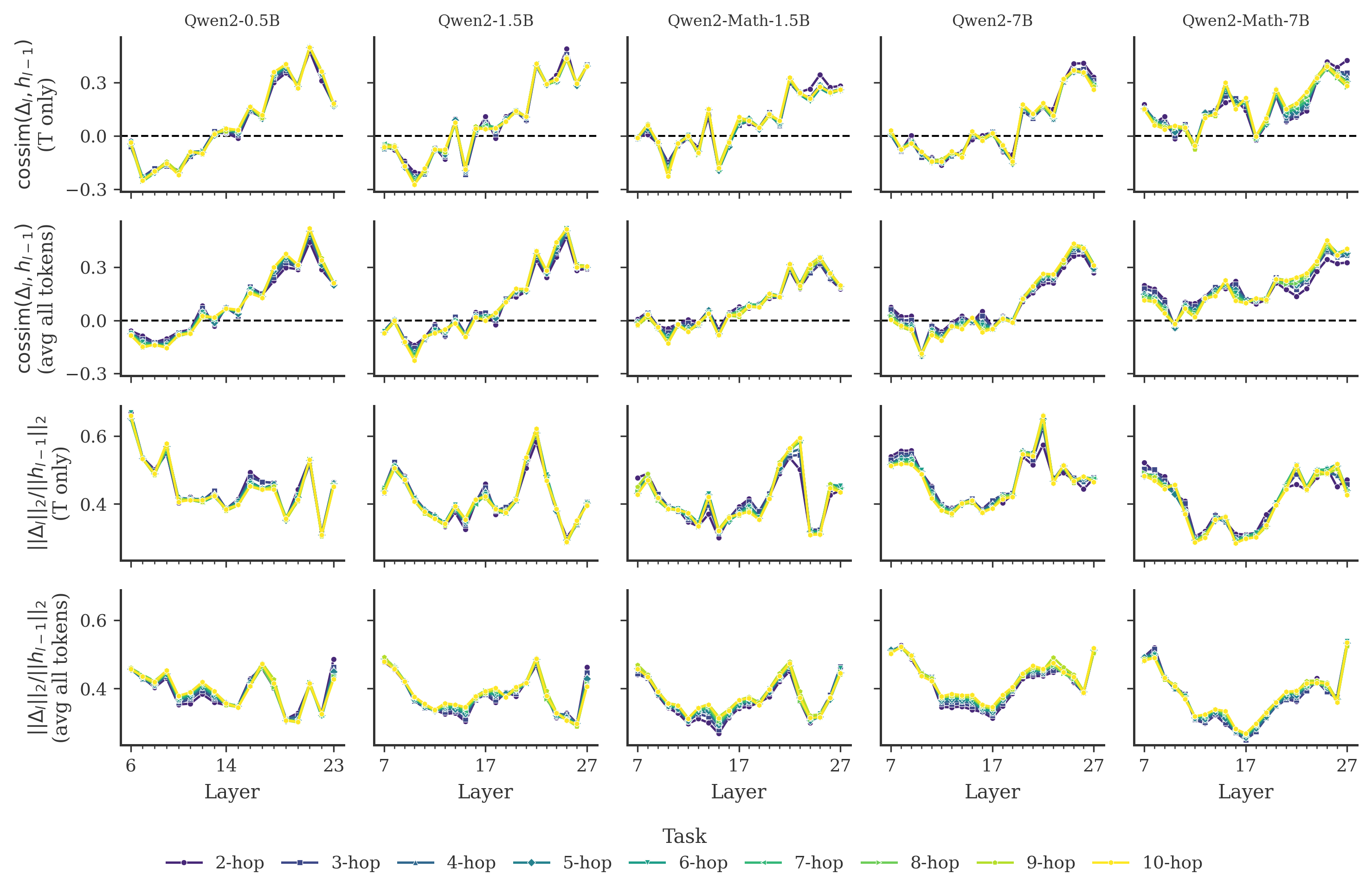}
    \caption{Qwen2 family}
    \label{fig:a}
  \end{subfigure}%
  \caption{Similarity of layer update to residual stream (top 2 rows) and relative contribution of layer update to residual stream (bottom 2 rows) by layer for Qwen models}
  \label{fig:hidden-pretrained}
\end{figure}

\begin{figure}[h]
  \centering
  \begin{subfigure}{0.99\textwidth}
    \centering
       \includegraphics[width=\linewidth]{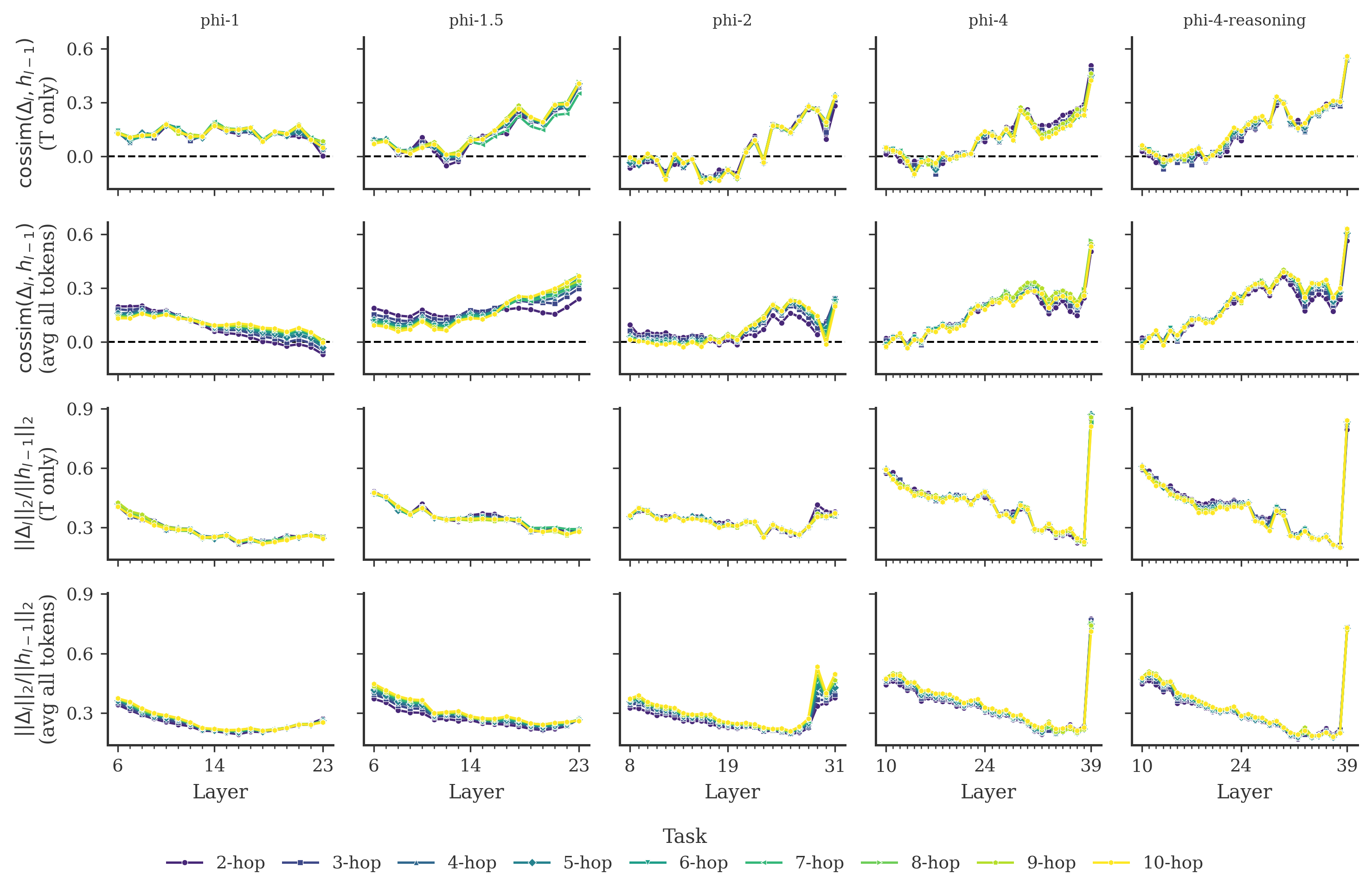}
    \caption{Phi family}
    \label{fig:a}
  \end{subfigure}%
  \hfill
      \begin{subfigure}{0.99\textwidth}
    \centering
       \includegraphics[width=\linewidth]{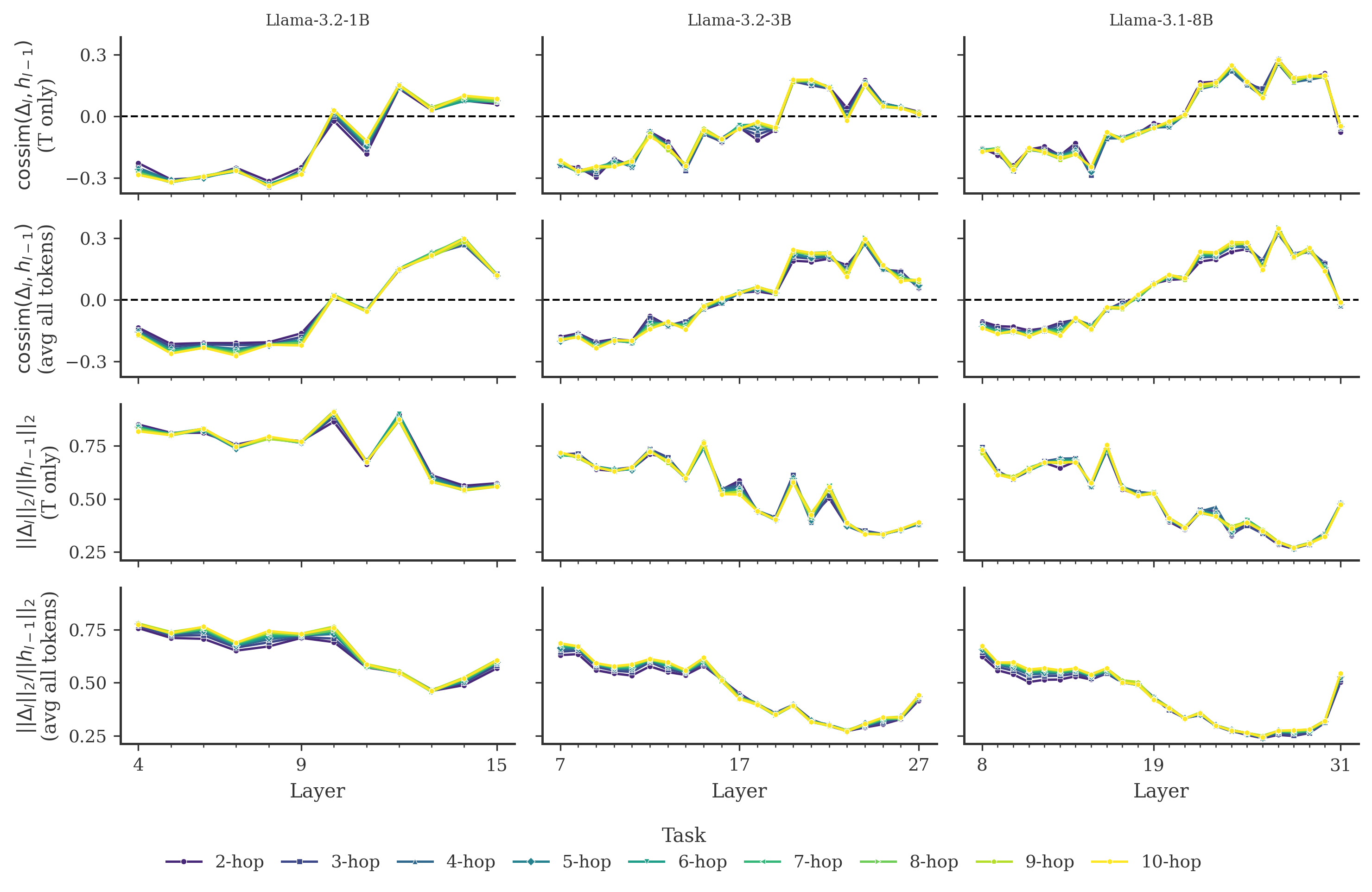}
    \caption{Llama family}
    \label{fig:a}
  \end{subfigure}%
  \hfill
  \caption{Similarity of layer update to residual stream (top 2 rows) and relative contribution of layer update to residual stream (bottom 2 rows) by layers for Phi and Llama models}
  \label{fig:hidden-pretrained-llamaphi}
\end{figure}

\begin{figure}[h]
  \centering
    \begin{subfigure}{0.9\textwidth}
    \centering
       \includegraphics[width=\linewidth]{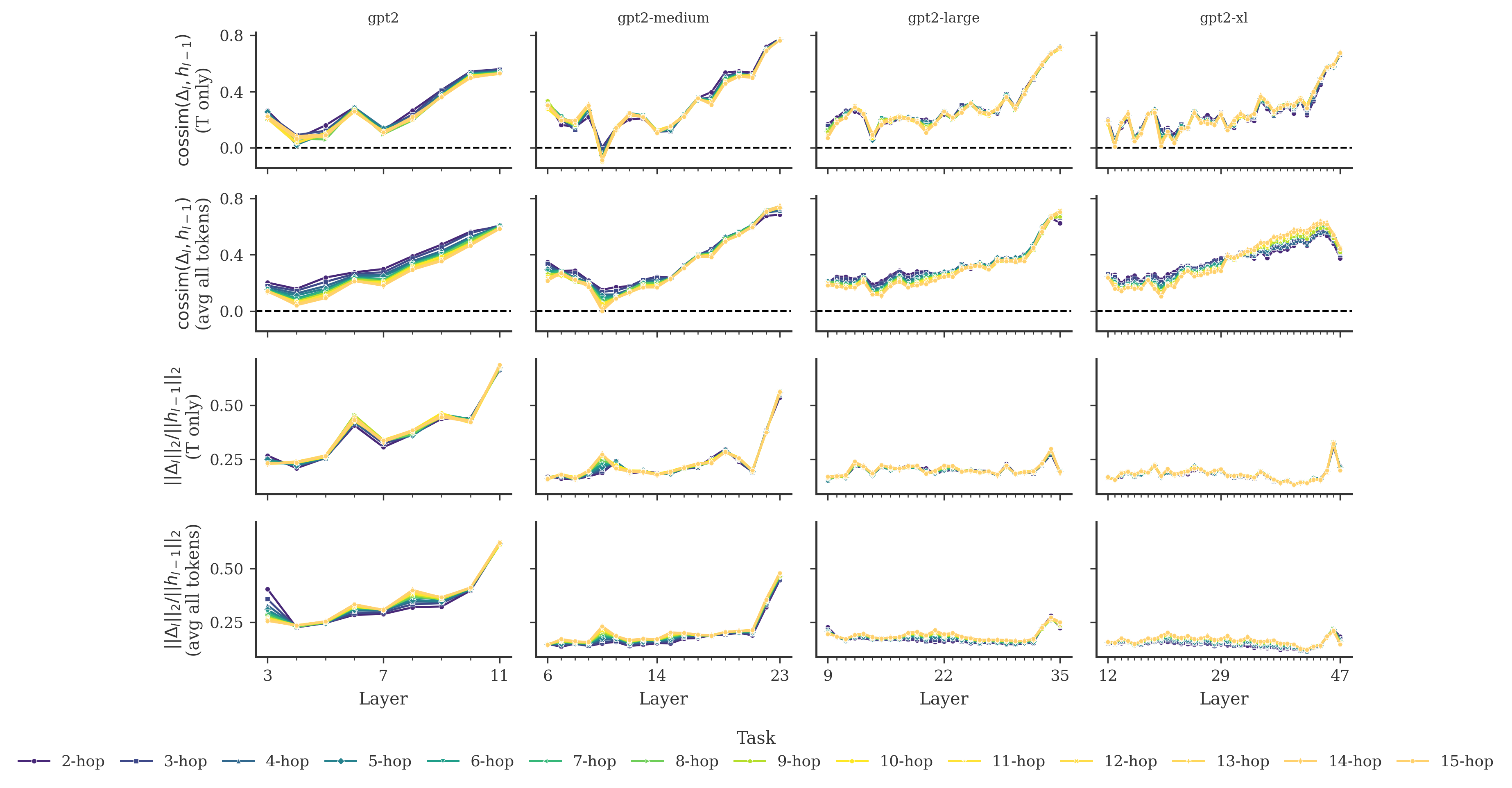}
    \caption{GPT2 Pretrained}
    \label{fig:diff-gpt2-pret}
  \end{subfigure}%
  \hfill
    \centering
      \begin{subfigure}{0.9\textwidth}
    \centering
       \includegraphics[width=\linewidth]{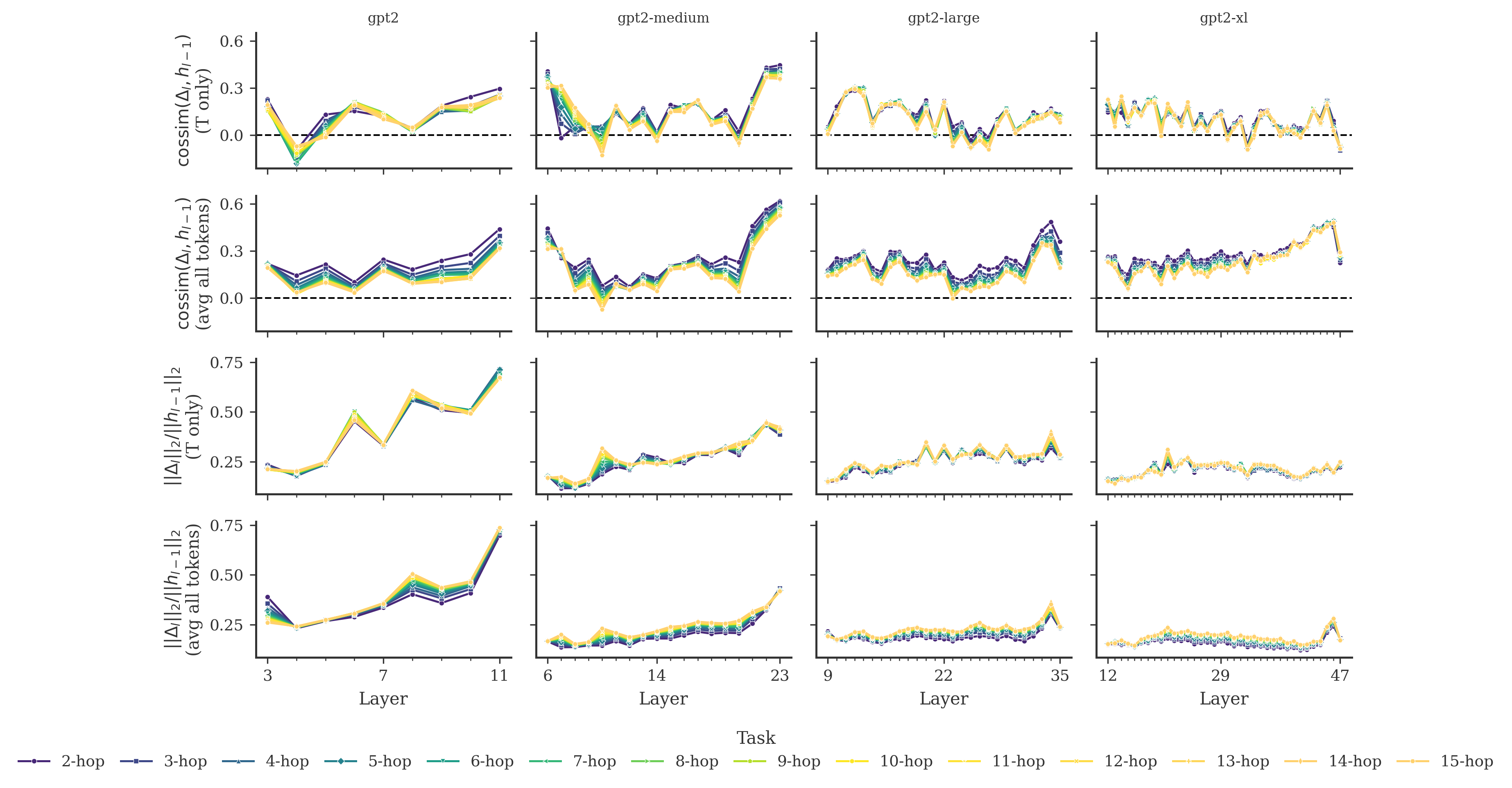}
    \caption{GPT2 LoRA finetuned}
    \label{fig:a}
  \end{subfigure}%
  \hfill
    \centering
      \begin{subfigure}{0.9\textwidth}
    \centering
       \includegraphics[width=\linewidth]{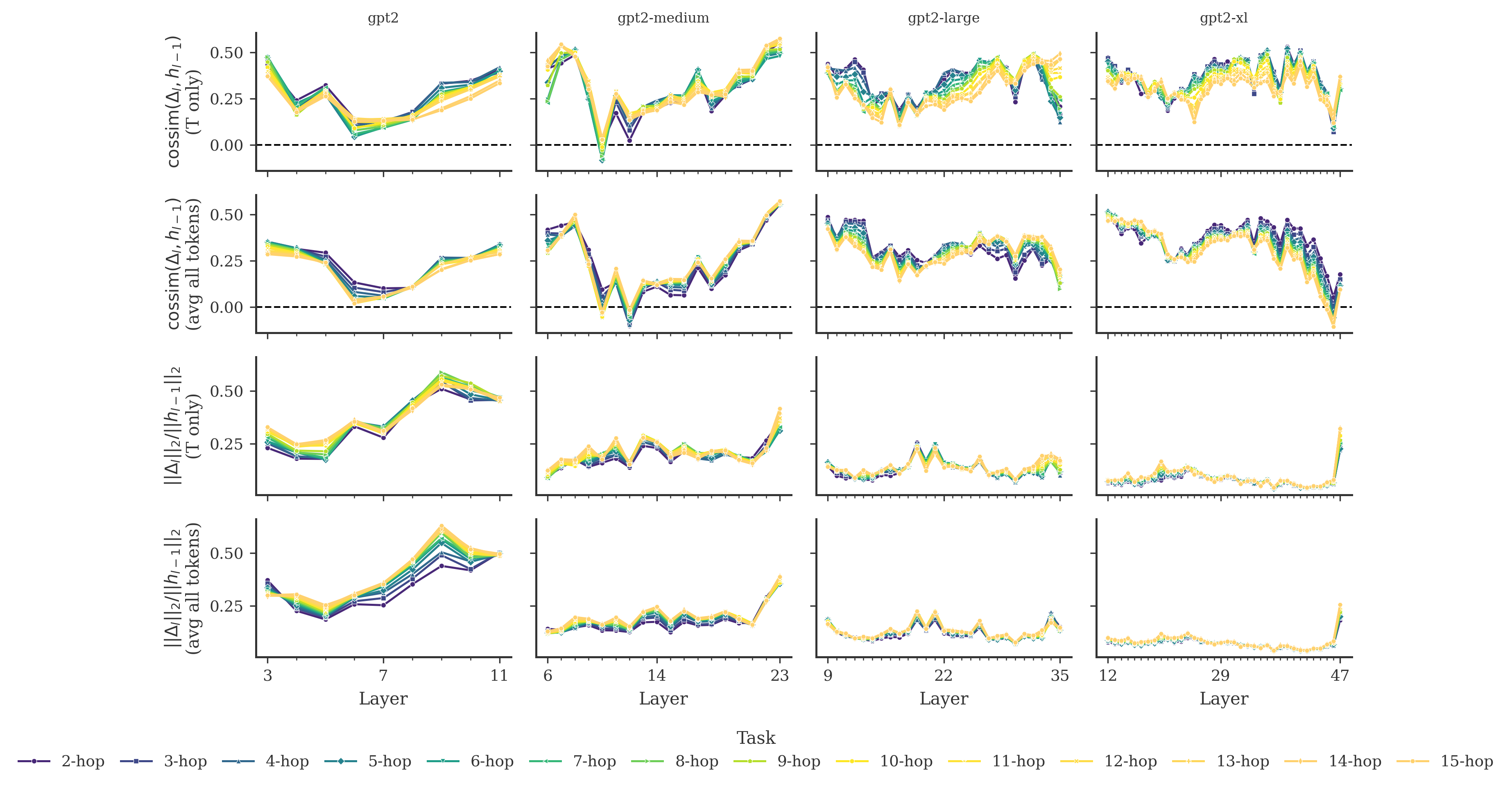}
    \caption{GPT2 fully finetuned}
    \label{fig:a}
  \end{subfigure}%
  \hfill
  \caption{Similarity of layer update to residual stream (top 2 rows) and relative contribution of layer update to residual stream (bottom 2 rows) by layer for GPT2 family: Pretrained, LoRA finetuned and fully finetuned}
  \label{fig:hidden-fine-gpt2}
\end{figure}

\begin{figure}[h]
  \centering
  \hfill
    \begin{subfigure}{0.9\textwidth}
    \centering
       \includegraphics[width=\linewidth]{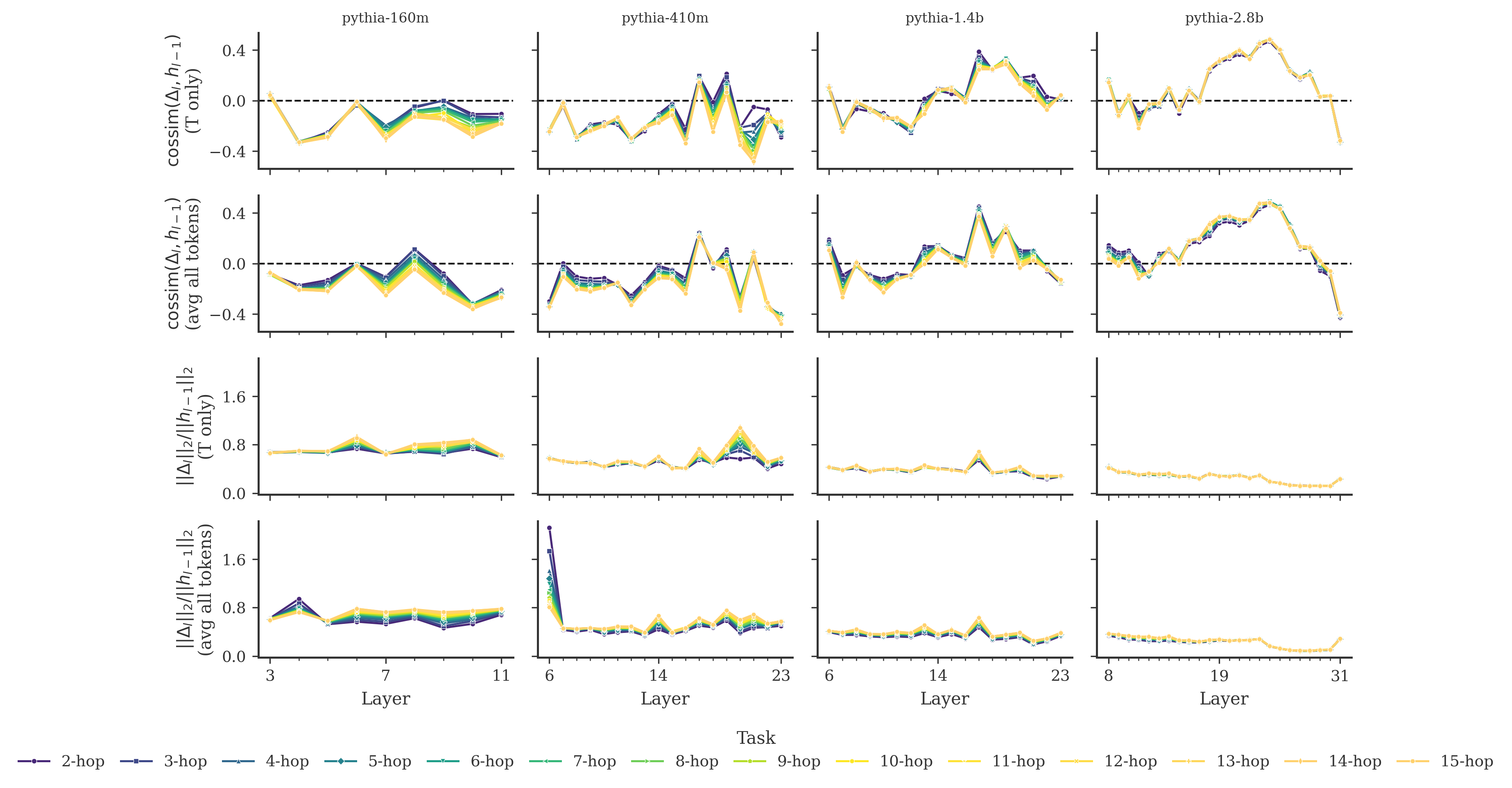}
    \caption{Pythia pretrained}
    \label{fig:diff-pret-pythia}
  \end{subfigure}%
  \hfill
      \begin{subfigure}{0.9\textwidth}
    \centering
       \includegraphics[width=\linewidth]{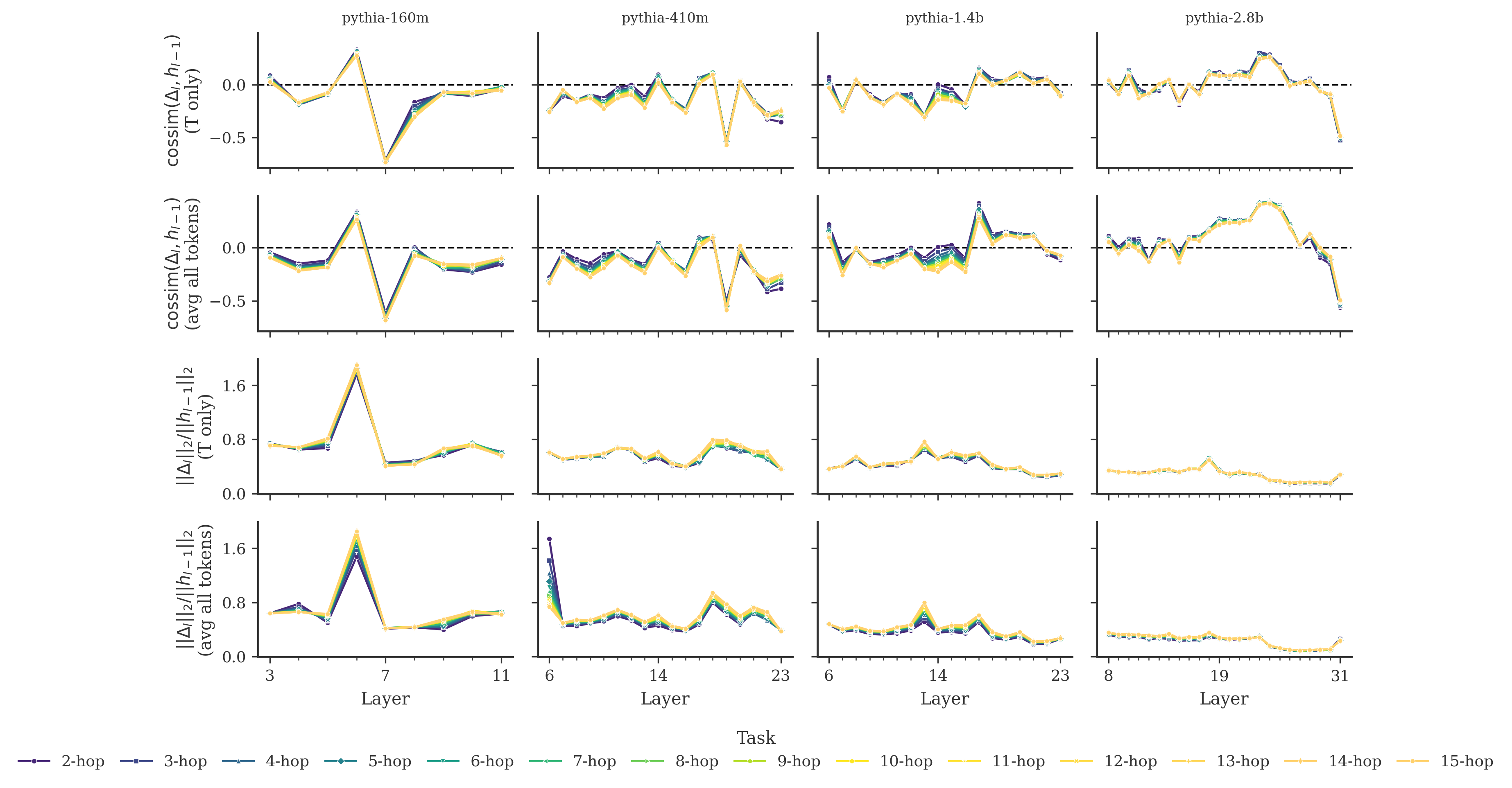}
    \caption{Pythia LoRA finetuned}
    \label{fig:a}
  \end{subfigure}%
  \hfill
      \begin{subfigure}{0.9\textwidth}
    \centering
       \includegraphics[width=\linewidth]{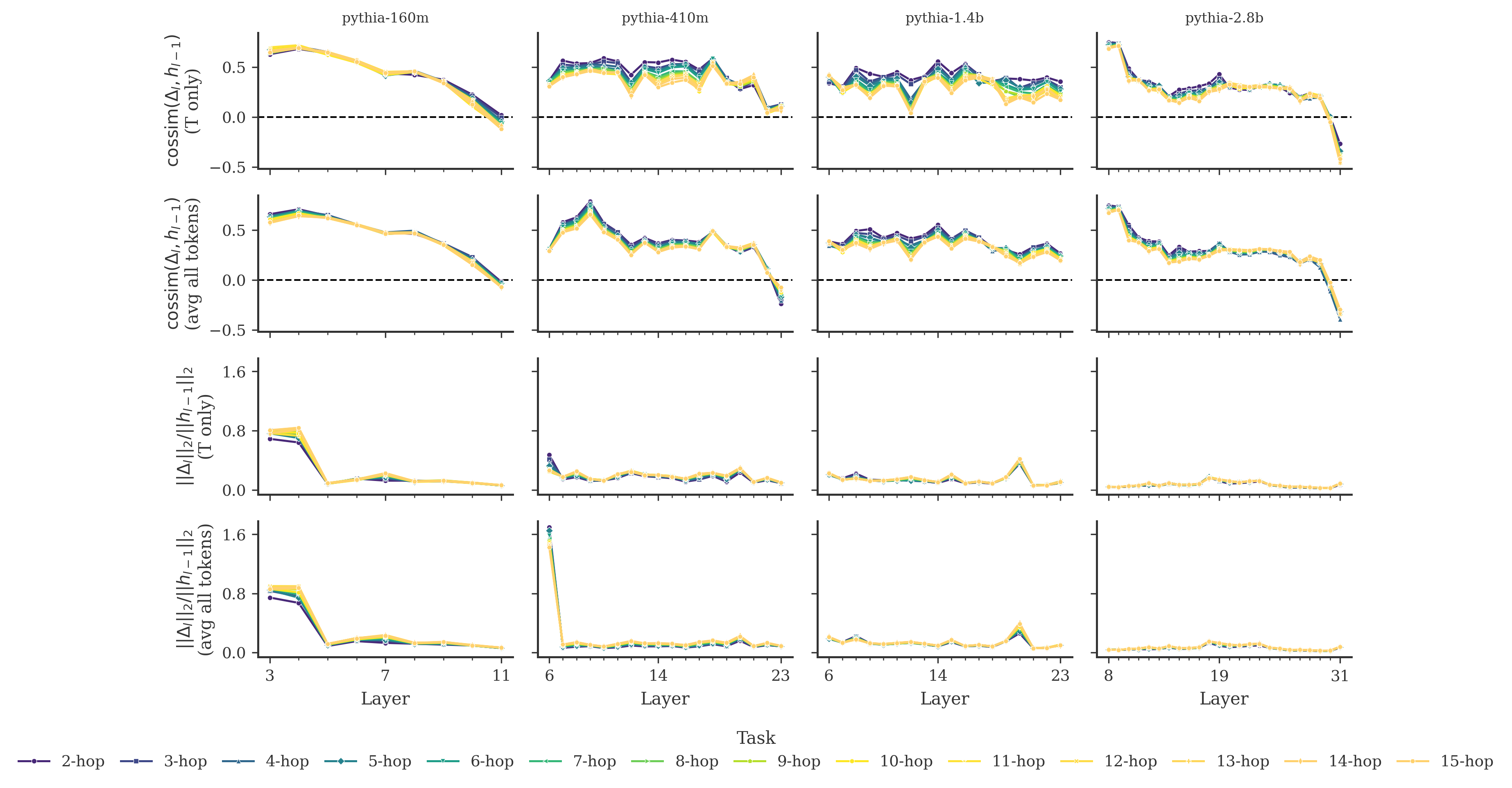}
    \caption{Pythia fully finetuned}
    \label{fig:a}
  \end{subfigure}%
  \hfill
  \caption{Similarity of layer update to residual stream (top 2 rows) and relative contribution of layer update to residual stream (bottom 2 rows) by layer for pythia family: Pretrained, LoRA finetuned and fully finetuned}
  \label{fig:hidden-fine-pythia}
\end{figure}

Across most pretrained models (Qwen2 and Qwen2.5 in \cref{fig:hidden-pretrained}, LLaMA and Phi in \cref{fig:hidden-pretrained-llamaphi}, GPT-2 in \cref{fig:diff-gpt2-pret}, and Pythia in \cref{fig:diff-pret-pythia}), we observe the same general trends as \cite{csordas2025language}: the contribution of later layers to the residual stream decreases in magnitude, and updates become increasingly similar in direction to the residual stream after being near-orthogonal in the middle layers. The main exception is Phi-1, where the cosine similarity trend appears reversed.

Regarding differences across task difficulty, the results are mixed and not strongly consistent across model families, although small differences are visible in all plots. In all families, the average contribution to the residual stream for higher-hop tasks appears slightly larger in some but not all of the middle to later layers when averaged across all tokens, though not at the final token alone (where one might expect most task-relevant computation to occur, given that the query is not known in advance). The novelty of computations, as measured by cosine similarity to the residual stream, tends to be higher (i.e.\ closer to zero) for lower hop counts when differences are present, with the exception of the pretrained GPT-2 and Pythia families, where cosine similarity is sometimes closer to zero for higher hop counts.

The results for the finetuned models, presented in \cref{fig:hidden-fine-gpt2} panels (b) and (c) for GPT-2 and \cref{fig:hidden-fine-pythia} panels (b) and (c) for Pythia, are more clear-cut: across almost all model sizes and finetuning regimes, the cosine similarity between the residual stream and the layer update is markedly lower for higher hop counts, indicating that more novel computation is being performed for harder tasks.

\FloatBarrier
\newpage
\subsection{Perplexity of finetuned models}\label{app:perplexity}
Below, we evaluate the perplexity of finetuned models on the WikiText-2 dataset, confirming that the fully finetuned models do indeed lose their general language modeling ability entirely, while the LoRA finetuned models largely preserve it. We observe that for the smaller Pythia models, language modeling ability degrades under LoRA finetuning as well, whereas this does not occur for the smaller GPT-2 models. This is likely attributable to the difference in LoRA target modules: for Pythia, we include the attention output projection (\texttt{dense}) in addition to the QKV matrices, whereas for GPT-2 we target only the merged QKV projection (\texttt{c\_attn}).
\begin{table}[h!]
\centering
\footnotesize
\setlength{\tabcolsep}{4pt}
\begin{tabular}{lrrrrrrrrrr}
\toprule
& \multicolumn{2}{c}{Pretrained} & \multicolumn{2}{c}{LoRA (5-hop)} & \multicolumn{2}{c}{LoRA (10-hop)} & \multicolumn{2}{c}{FT (5-hop)} & \multicolumn{2}{c}{FT (10-hop)} \\
\cmidrule(lr){2-3} \cmidrule(lr){4-5} \cmidrule(lr){6-7} \cmidrule(lr){8-9} \cmidrule(lr){10-11}
Model & \shortstack{CLUTRR\\(Ans.)} & \shortstack{Wiki\\-2} & \shortstack{CLUTRR\\(Ans.)} & \shortstack{Wiki\\-2} & \shortstack{CLUTRR\\(Ans.)} & \shortstack{Wiki\\-2} & \shortstack{CLUTRR\\(Ans.)} & \shortstack{Wiki\\-2} & \shortstack{CLUTRR\\(Ans.)} & \shortstack{Wiki\\-2}  \\
\midrule
GPT-2 (124M) & 34.1 & 24.4 & 1.4 & 25.5 & 1.2 & 24.9 & 2.4 & $>$1000 & 1.1 & $>$1000 \\
GPT-2 Medium (355M) & 25.5 & 18.0 & 1.5 & 18.5 & 1.2 & 18.4 & 2.7 & $>$1000 & 1.0 & $>$1000 \\
GPT-2 Large (774M) & 20.4 & 15.6 & 1.1 & 15.7 & 1.1 & 15.7 & 1.9 & $>$1000 & 1.0 & $>$1000 \\
GPT-2 XL (1.5B) & 12.0 & 14.2 & 1.1 & 14.4 & 1.0 & 14.5 & 3.2 & $>$1000 & 1.1 & $>$1000 \\
\midrule
Pythia 160M & 32.1 & 23.5 & 2.5 & 414.2 & 1.6 & $>$1000 & 4.5 & $>$1000 & 1.4 & $>$1000 \\
Pythia 410M & 16.3 & 14.6 & 1.2 & 160.7 & 1.1 & 137.3 & 1.8 & $>$1000 & 1.1 & $>$1000 \\
Pythia 1.4B & 9.3 & 10.7 & 1.1 & 22.1 & 1.1 & 20.7 & 1.6 & $>$1000 & 1.5 & $>$1000 \\
Pythia 2.8B & 7.8 & 9.1 & 1.3 & 10.2 & 1.0 & 9.8 & 1.4 & $>$1000 & 1.2 & $>$1000 \\
\bottomrule
\end{tabular}
\caption{Perplexity on CLUTRR final answer and WikiText-2 across model families and training conditions. CLUTRR = answer-only perplexity (averaged across hops 2--10); Wiki-2 = WikiText-2 test set perplexity (sliding window).\label{tab:perplexity}}
\end{table}

\end{document}